\documentclass{article}

\usepackage{arxiv}

\usepackage[utf8]{inputenc} 
\usepackage[T1]{fontenc}    
\usepackage{hyperref}       
\usepackage{url}            
\usepackage{booktabs}       
\usepackage{amsfonts}       
\usepackage{nicefrac}       
\usepackage{microtype}      
\usepackage{lipsum}
\usepackage{graphicx}

\usepackage{colortbl}

\newif\ifdraft
\drafttrue

\newcommand{\manucomment}[1]{\ifdraft{\leavevmode\color{red}{[MF]: 
 {#1}}}\else{\vspace{0ex}}\fi}
\newcommand{\alexcomment}[1]{\ifdraft{\leavevmode\color{purple}{[AM]: 
 {#1}}}\else{\vspace{0ex}}\fi}

\usepackage{algpseudocode}
\usepackage{algorithm}
\usepackage{subcaption}
\newcommand{\codecomment}[1]{\State {\leavevmode\color{gray}{// {#1}}}}
\newcommand{\aftercodecomment}[1]{{\leavevmode\color{gray}{\hspace{1ex} // {#1}}}}
\newcommand{\scikitml}{\textsc{scikit-multilearn}}
\usepackage{dcolumn,booktabs}
\newcolumntype{d}[1]{D{.}{.}{#1}}

\usepackage{array}

\usepackage{amsmath}
 
\usepackage[english]{babel}
 
\usepackage[T1]{fontenc}

\usepackage[right]{lineno}
 \linenumbers
 
\usepackage{multibib}

\usepackage{multicol} 
 
\usepackage{multirow} 
 
\usepackage{pifont}
 
\usepackage{rotating} 
 
\usepackage{soul} 
 \sethlcolor{pink}

\usepackage{tikz}
 
\usepackage{tipa}
 
\usepackage[utf8]{inputenc} 
 
\usepackage{url} 
 
\usepackage{capt-of}

\newcommand{\bqbc}{\textsc{BC+BA}}
\newcommand{\bqmc}{\textsc{MLC+BA}}
\newcommand{\mqbc}{\textsc{BC+MLA}}
\newcommand{\mqmc}{\textsc{MLC+MLA}}
\newcommand{\abse}{$\operatorname{AE}$}
\newcommand{\rabse}{$\operatorname{RAE}$}

\newcommand{\OQ}{OQ}
\newcommand{\MLC}{MLC}
\newcommand{\SLQ}{SLQ}
\newcommand{\MLQ}{MLQ}

\newcommand{\indfn}[1]{\mathbf{1}\hspace{-3px}\left[#1\right]}

\newcommand{\killpunct}[1]{} 

\newcommand{\side}[1]{\begin{sideways}{#1}\end{sideways}}

\newcolumntype{.}{D{.}{.}{-1}}

\newcolumntype{M}[1]{>{\centering\arraybackslash}m{#1}}

\newcolumntype{N}{@{}m{0pt}@{}}

\newcolumntype{C}[1]{>{\centering\let\newline\\\arraybackslash\hspace{0pt}}m{#1}}

\newcolumntype{Y}{>{\centering\arraybackslash}X}
 
\title{Multi-Label Quantification}

\author{
Alejandro Moreo \\
  Istituto di Scienza e Tecnologie dell'Informazione\\
Consiglio Nazionale delle Ricerche\\
Via Giuseppe Moruzzi 1\\
Pisa, Italy, 56124 \\
  \texttt{alejandro.moreo@isti.cnr.it} \\
   \And
Manuel Francisco \\
Department of Computer Science and Artificial
Intelligence\\ 
University of Granada\\
C Periodista Daniel Saucedo Aranda s/n\\
  Granada, Spain 18071 \\
  \texttt{francisco@decsai.ugr.es} \\
  \And
Fabrizio Sebastiani \\
  Istituto di Scienza e Tecnologie dell'Informazione\\
Consiglio Nazionale delle Ricerche\\
Via Giuseppe Moruzzi 1\\
Pisa, Italy, 56124 \\
  \texttt{fabrizio.sebastiani@isti.cnr.it} \\
}

\begin{document}

\maketitle

\begin{abstract}
  Quantification, variously called \emph{supervised prevalence
  estimation} or \emph{learning to quantify}, is the supervised
  learning task of generating predictors of the relative frequencies
  (a.k.a.\ \emph{prevalence values}) of the classes of interest in
  unlabelled data samples. While many quantification methods have been
  proposed in the past for binary problems and, to a lesser extent,
  single-label multiclass problems, the multi-label setting (i.e., the
  scenario in which the classes of interest are not mutually
  exclusive) remains by and large unexplored. A straightforward
  solution to the multi-label quantification problem could simply
  consist of recasting the problem as a set of independent binary
  quantification problems. Such a solution is simple but naïve, since
  the independence assumption upon which it rests is, in most cases,
  not satisfied. In these cases, knowing the relative frequency of one
  class could be of help in determining the prevalence of other
  related classes.  We propose the first truly multi-label
  quantification methods, i.e., methods for inferring estimators of
  class prevalence values that strive to leverage the stochastic
  dependencies among the classes of interest in order to predict their
  relative frequencies more accurately. We show empirical evidence
  that natively multi-label solutions outperform the naïve approaches
  by a large margin. The code to reproduce all our experiments is
  available online.
\end{abstract}




\newpage



\section{Introduction}
\label{sec:intro}

\noindent Many fields such as the social sciences, political science,
market research, or epidemiology (to name a few), are inherently
interested in \emph{aggregate} data, i.e., in how populations of
individuals are distributed according to one or more indicators of
interest. Researchers who operate in these specialty areas are instead
little interested in the individuals \textit{per se}, since in these
fields the individuals are relevant only inasmuch as they are members
of the population of interest; in other words, disciplines such as the
above are interested ``not in the needle, but in the
haystack''~\cite{Hopkins:2010fk}.

Sometimes, researchers active in these fields use supervised learning
to obtain the data they need. For instance, epidemiologists interested
in the distribution of the causes of death across different
geographical regions may sometimes need to \emph{infer} the cause of
death of each person by classifying, via a machine-learned text
classifier, a verbal description of the symptoms that affected a
deceased person~\cite{King:2008fk}. However, these researchers are not
specifically interested in the class (representing a given cause of
death) to which an individual belongs; rather, their final goal is
estimating the \emph{prevalence} (i.e., \emph{relative frequency}, or
\emph{prior probability}) of each class in the unlabelled data.

At a first glance, estimating these prevalence values via supervised
learning looks like a direct application of classification, since one
could simply (i) train a classifier on labelled data, (ii) use this
classifier to issue label predictions for each unlabelled datapoint
(i.e., individual) in the population of interest, (iii) count how many
datapoints have been attributed to each of the classes of interest,
and (iv) normalize the counts by the total number of labelled
datapoints, thus obtaining the estimated relative frequencies of the
classes. However, there is by now abundant evidence
\cite{Bunse:2022dz, Forman:2008kx, Gonzalez:2017it, Moreo:2022bf,
Schumacher:2021ty} that such an approach, known in the literature as
the ``Classify and Count'' (CC) method, yields poor class prevalence
estimates when the distribution of the unlabelled datapoints across
the classes differs substantially from the analogous distribution
observed during training. This latter condition is typically known as
\emph{prior probability
shift}~\cite{Moreno-Torres:2012ay,Storkey:2009lp}; in the
aforementioned disciplines this condition is ubiquitous since, quite
obviously, there is interest in inferring a distribution \emph{only}
when we assume this unknown distribution to be possibly different from
the known distribution that characterizes the training data. The main
reason why CC 
tends to fail
in the presence of prior probability shift is that, in this case, the
IID assumption
on which most classifiers based on supervised learning rest upon, does
not hold.

\emph{Quantification} (variously called \emph{learning to quantify},
\emph{supervised prevalence estimation}, or \emph{class prior
estimation}) is the research field concerned with obtaining accurate
estimators of class prevalence values via machine
learning~\cite{Gonzalez:2017it}.
Given its obvious relationship with classification, quantification has
been extensively studied in the two main settings typical of
classification, i.e., the binary setting (\emph{binary quantification}
-- BQ), in which the \emph{codeframe} (i.e., the set of classes of
interest) contains only two classes~\cite{Card:2018pb, Forman:2008kx,
Bella:2010kx, Esuli:2015gh,
Hassan:2020kq}, 
and the single-label multiclass setting (\emph{single-label
quantification} -- \SLQ), which involves three or more mutually
exclusive classes~\cite{Bunse:2022qq, Firat:2016uq, Gao:2016uq,
Moreo:2022bf, Schumacher:2021ty}. Quantification has also been studied
in other (less popular) settings, such as the ordinal case
(\emph{ordinal quantification} -- \OQ), in which there is a total
ordering among the classes (see, e.g.,~\cite{DaSanMartino:2016jk,
Bunse:2022dz, Castano:2022nz}).

One important setting which 
remains to a large extent unexplored in the quantification literature
is \emph{multi-label quantification} (\MLQ), the scenario in which
every datapoint may belong to zero, one, or several classes at the
same time. Multi-label data arise naturally in many applicative
contexts, including the medical
domain~\cite{Janjua:2022yi}, 
the legal domain~\cite{Csanyi:2022my}, 
industry~\cite{Taqvi:2022oh}, or
cybersecurity~\cite{Mukherjee:2022xw}, among many others, and has been
thoroughly studied under the lens of classification in past literature
\cite{Elisseeff:2001iz, Zhang:2007as, Montanes:2014fu}, especially in
the field of text classification~\cite{McCallum:1999zv, Gao:2004op,
Read:2011zc, Zhang:2021ma}; see~\cite{Tsoumakas:2007nl,
Herrera:2016xt} for an overview of this topic. Despite the ubiquity of
multi-label data,
we 
are aware of only one previous attempt to cope with the \MLQ\ problem
\cite{Levin:2017dq}; in this paper we set out to analyze \MLQ\
systematically.

We start by noting that, since quantification systems are expected to
be robust to prior probability shift, we need to test them against
datasets exhibiting substantial amounts of shift.
Our first contribution is the first experimental protocol
specifically designed for multi-label quantification, a protocol that
guarantees that the data MLQ systems are tested against do comply with
the above desideratum.

We carry on by noting that a trivial solution for MLQ could simply
consist of training one independent binary quantifier for each of the
classes in the codeframe. However, such a solution is arguably a
``naïve'' one, as it assumes the classes to be independent of each
other, and thus does not attempt to leverage the \emph{class-class
correlations}, i.e., the stochastic dependencies that may exist among
different classes. We show empirical evidence that multi-label
quantifiers constructed according to this naïve intuition yield
suboptimal performance, and that this happens independently of the
method used for training the binary quantifiers.


We then move on to studying different possible strategies for tackling
MLQ, and subdivide these strategies in four groups, based on their way
of addressing (if at all)
the multi-label nature of the problem. While the first two groups can
be instantiated by using already available techniques, the other two
cannot, since this would require ``aggregation'' techniques (see
Section~\ref{sec:quant}) that leverage the stochastic relations
between classes, and no such method has been proposed before. We
indeed propose two such methods, called RQ and LPQ. By means of
extensive experiments that we have carried out using 15 publicly
available datasets, we show that, when working in combination with a
classifier that itself leverages the above-mentioned stochastic
relations, LPQ and (especially) RQ outperform all other MLQ
techniques.

The rest of this article is structured as follows. After devoting
Section~\ref{sec:notation} to notation and preliminaries, in
Section~\ref{sec:relwork} we survey related work on quantification and
on coping with multi-label data. In Section~\ref{sec:mlapp} we move on
to propose an evaluation protocol specifically devised for \MLQ. In
Section~\ref{sec:quant} we characterize the four groups of MLQ systems
according to the stages in which the multi-label nature of the problem
is tackled, and propose the first methods (Section~\ref{sec:quant:ml})
that allow doing this also at the ``aggregation'' stage. In
Section~\ref{sec:experiments} we present the experiments we have
carried out and discuss the results, while
Section~\ref{sec:conclusions} wraps up. The code to reproduce all our
experiments is available at
\url{https://github.com/manuel-francisco/quapy-ml/} .

\section{Notation and Definitions}
\label{sec:notation}

\noindent In this paper we use the following notation. By $\mathbf{x}$
we indicate a datapoint drawn from a domain $\mathcal{X}$ of
datapoints, while by $y$ we indicate a class drawn from a finite,
predefined set of classes (also known as a \emph{codeframe})
$\mathcal{Y}=\{y_{1}, ..., y_{n}\}$, with $n$ the number of classes of
interest. Symbol $\sigma$ denotes a \emph{sample}, i.e., a non-empty
set of (labelled or unlabelled) datapoints drawn from
$\mathcal{X}$. By $p_{\sigma}(y)$ we indicate the \emph{true}
prevalence of class $y$ in sample $\sigma$, by $\hat{p}_{\sigma}(y)$
we indicate an \emph{estimate} of this prevalence, and by
$\hat{p}_{\sigma}^{q}(y)$ we indicate the estimate of this prevalence
obtained by means of quantification method $q$. We will denote by
$\mathbf{p}=(p_{1}, \ldots, p_{n})$ a real-valued vector. When
$\mathbf{p}$ is a vector of class prevalence values, then $p_{i}$ is
short for $p_\sigma(y_{i})$.

We first formalize the \SLQ\ problem
(Section~\ref{sec:notation:singlelabel}) and then propose a definition
of the \MLQ\ problem (Section~\ref{sec:notation:multi-label}).


\subsection{Single-Label Codeframes}
\label{sec:notation:singlelabel}

\noindent In single-label problems, each datapoint $\mathbf{x}$
belongs to one and only one class in $\mathcal{Y}$. We denote a
datapoint with its true class label as a pair $(\mathbf{x},y)$,
indicating that $y\in\mathcal{Y}$ is the true label of
$\mathbf{x}\in\mathcal{X}$. We represent a set of $k$ datapoints as
$\{(\mathbf{x}^{(i)},y^{(i)})_{i=1}^k: \mathbf{x}^{(i)}\in\mathcal{X},
y^{(i)}\in\mathcal{Y}\}$. By $L$ we denote a collection of
\underline{l}abelled datapoints, that we typically use as a training
set, while by $U$ we denote a collection of \underline{u}nlabelled
datapoints, that we typically use for testing purposes.

We define a \emph{single-label hard classifier} as a
function $$h:\mathcal{X}\rightarrow \mathcal{Y}$$
\noindent i.e., a predictor of the class attributed to a datapoint.
We will instead take a \emph{single-label soft classifier} to be a
function $$s:\mathcal{X}\rightarrow \Delta^{n-1}$$
\noindent with $\Delta^{n-1}$ the unit ($n$-1)-simplex 
(aka \emph{probability simplex} or \emph{standard simplex}) defined as
$$\Delta^{n-1}=\{(p_{1}, \ldots,p_{n}) \mid p_{i}\in[0,1], \sum_{i=1}^{n} p_{i}=1\}$$ 
\noindent i.e., as the domain of all vectors representing probability
distributions over $\mathcal{Y}$.
%
%
We define a \emph{single-label quantifier} as a function
$$q : 2^\mathcal{X}\rightarrow \Delta^{n-1}$$ \noindent
i.e., a function mapping samples drawn from $\mathcal{X}$ into
probability distributions over $\mathcal{Y}$.

Note that, despite the fact that the codomains of soft classifiers and
quantifiers are the same, in the former case the $i$-th component of
$s(\mathbf{x})$ denotes the posterior probability
$\Pr(y_{i} | \mathbf{x})$, i.e., the probability that $\mathbf{x}$
belongs to class $y_{i}$ as estimated by $s$, while in the latter case
it denotes the class prevalence value $p_\sigma(y_{i})$ as estimated
by $q$.

By $d(\mathbf{p},\hat{\mathbf{p}})$ we denote an evaluation measure
for \SLQ; these measures are typically \emph{divergences}, i.e.,
functions that measure the amount of discrepancy between two
probability distributions. Everything we say for single-label problems
applies to the binary case as well, since the latter is the special
case of the former in which $n=2$, with one class typically acting as
the ``positive class'', and the other as the ``negative class''.


\subsection{Multi-Label Codeframes}
\label{sec:notation:multi-label}

\noindent In multi-label problems each datapoint $\mathbf{x}$ can
belong to zero, one, or more than one class in $\mathcal{Y}$. We
denote a datapoint with its true labels as a pair $(\mathbf{x},Y)$, in
which $Y\subseteq\mathcal{Y}$ is the set of true labels assigned to
$\mathbf{x}\in\mathcal{X}$. A multi-label collection with $k$
datapoints is represented as
$\{(\mathbf{x}^{(i)},Y^{(i)})_{i=1}^k: \mathbf{x}^{(i)}\in\mathcal{X},
Y^{(i)}\subseteq\mathcal{Y}\}$.
We define a \emph{multi-label hard classifier} as a
function $$h:\mathcal{X}\rightarrow 2^\mathcal{Y}$$
\noindent while we define a \emph{multi-label soft classifier} as a
function
$$s:\mathcal{X}\rightarrow [0,1]^{n}$$ Note that, unlike
in the single-label case, the codomain of function $s$ is not a
probability simplex, but the set of all real-valued vectors
$(p_{1}, \ldots, p_{n})$ such that $p_{i}\in[0,1]$, since constraint
$\sum_{i=1}^{n} p_{i} = 1$ is not enforced.

We define a \emph{multi-label quantifier} as a function
$$q : 2^\mathcal{X}\rightarrow [0,1]^{n}$$
\noindent i.e., a function mapping samples from $\mathcal{X}$ into
vectors of $n$ class prevalence values, where the class prevalence
values in a vector do not need to sum up to 1.




\section{Related Work}
\label{sec:relwork}


\subsection{Multi-Label Quantification}
\label{sec:multi-label}

\noindent The task of quantification was first proposed by Forman in
2005~\cite{Forman:2005fk}, and emerges from the observation that in
some applications of classification the real goal is the estimation of
class prevalence values, and the prediction of individual class labels
is nothing but an intermediate step to achieve it.
Forman~\cite{Forman:2005fk} observed that the so-called ``classify and
count'' method (discussed in the introduction) tends to deliver biased
estimators of class prevalence when confronted with situations
characterized by prior probability shift. Since then, many
contributions to this field have been published, describing new
quantification methods (see~\cite{Gonzalez:2017it} for an overview),
measures for evaluating the accuracy of quantifiers
\cite{Sebastiani:2020qf, Sakai:2021lp}, and protocols for the
experimental evaluation of quantification systems~\cite{Esuli:2015gh,
Forman:2005fk, Esuli:2022hy}. However, most of this work has focused
on the binary~\cite{Esuli:2015gh, Esuli:2018rm, Forman:2005fk,
Forman:2008kx, Maletzke:2019qd}, single-label multiclass
\cite{Bunse:2022qq, Gao:2016uq, Firat:2016uq, Moreo:2022bf}, or
ordinal~\cite{Bunse:2022dz, DaSanMartino:2016jk, Sakai:2021lp,
Castano:2022nz} versions of the problem, with essentially no attention
devoted to the multi-label case.


To the best of our knowledge, the only previously proposed method
dealing with multi-label (text) quantification is
\cite{Levin:2017dq}. The method, dubbed PCC-PAV, mainly consists of
performing a refined calibration of the posterior probabilities
returned by a set of binary classifiers, each trained to issue soft
predictions for a specific class in the codeframe, followed by the
application of the (binary version of the) ``probabilistic classify
and count'' (PCC) quantification method (explained in
Section~\ref{sec:quant:sl}). The refinement amounts to constraining
the calibration algorithm (\emph{Pair Adjacent Violators} --
PAV~\cite{Zadrozny:2002eu}) to generate posterior probabilities that
have an expected value equal to the class prevalence values observed
in the training set. This modification results in a complex quadratic
programming problem that the authors try to alleviate by imposing a
``document unification'' heuristics (thus giving rise to the variant
PCC-EPAV). However, aside from its substantial computational cost, the
method presents two important limitations. The first limitation is
that the calibration process encourages the quantifier to stick to the
class prevalence values encountered in the training set (hereafter:
the \emph{training prevalence}), and thus makes it inadequate to
tackle prior probability shift.  Indeed, in~\cite{Levin:2017dq} the
authors test this method by means of an experimental protocol that
offers no guarantee that the system is confronted with substantial
amounts of prior probability shift; this is witnessed by the fact that
the trivial baseline (one that simply returns the training prevalence
for every test sample) achieves reasonably high scores. A second
limitation is that the method uses \emph{independently trained} binary
classifiers; to put it another way, the method does not in any way
address the multi-label nature of the problem, and is billed
``multi-label'' only since it is tested on multi-label datasets. This
method thus squarely falls within the class of ``naïve methods'' that
we have discussed in the introduction, and that we will more formally
define as the simplest group of \MLQ\ methods in the scheme that we
propose in Section~\ref{sec:quant:ml}.


\subsection{Multi-Label Classification}
\label{sec:multi}

\noindent \label{par:rw:mlclf}Apart from PCC-PAV and PCC-EPAV, no
other published method explicitly addresses \MLQ. However, the field
of quantification is closely related to the field of classification,
and many of the concepts and principles adopted in quantification have
been borrowed from the classification literature. We will thus devote
most of this section to review existing approaches for multi-label
\emph{classification}, since some among the methods we propose in
Section~\ref{sec:quant:ml} for \MLQ\ are inspired by them.

In their survey~\cite{Tsoumakas:2007nl}, Tsoumakas and Katakis group
the approaches to \MLC\ in two main families: ``problem
transformation'' approaches (Section~\ref{sec:relwork:probtrans}) and
``algorithm adaptation techniques''
(Section~\ref{sec:relwork:algadapt}). Later on, a third family of
\MLC\ approaches based on ``ensemble methods''
(Section~\ref{sec:relwork:ensembles}) was introduced by Madjarov et
al.~\cite{Madjarov:2012yl}.






\subsubsection{Problem Transformation Approaches}
\label{sec:relwork:probtrans}

\noindent The ``problem transformation'' approach consists of
recasting the original multi-label classification problem as a
single-label one, so that existing techniques for the single-label
case can be applied directly.

The simplest approach relying on this principle is the so-called
``binary relevance'' (BR) approach~\cite{Luaces:2012qi,
Montanes:2014fu}. BR consists of treating each label independently,
thus tackling the multi-label problem as a set of $n$ binary
classification problems. BR is a simplistic approach, since the
stochastic dependencies among the classes are not taken into
account. Despite its simplicity, BR has proven to work well in
previous studies~\cite{Madjarov:2012yl}, and has always been, by far,
the most frequently adopted approach for tackling multi-label
classification problems.

The ``label powerset'' (LP) approach consists instead of transforming
each unique assignment of labels, as found in the training datapoints,
in a new label. For example, if a training datapoint $\mathbf{x}$ has
labels $Y=\{y_{1},y_{5},y_{6}\}$, with
$y_{1}, y_{5}, y_{6} \in \mathcal{Y}$, then the datapoint is
relabelled as $Y'=y_{1:5:6}$, with $y_{1:5:6}$ a new ``synthetic''
class belonging to a single-label codeframe
$\mathcal{Y}'\supseteq \mathcal{Y}$. Once this relabelling has been
performed for all unique label assignments, the problem is treated as
a single-label problem using $\mathcal{Y}'$ as the codeframe. Although
this approach models label dependencies~\cite{Spolaor:2013ir}, the
number of possible classes in the new codeframe $\mathcal{Y}'$ is
exponential in the number of classes in the original codeframe, since
there are actually $2^{n}$ possible assignments of a datapoint to
classes in $\mathcal{Y}$ (hence the name of the approach). Even if not
all label combinations occur in practice in a single dataset, the
total number of different assignments that could be found when large
multi-label codeframes are used could easily make the problem
intractable. Some heuristics based on ensembles (discussed in
Section~\ref{sec:relwork:ensembles}) have been proposed to make the
problem tractable.  A further limitation of this approach is its low
statistical robustness, since, once a training datapoint $\mathbf{x}$
with labels $Y=\{y_{1},y_{5},y_{6}\}$ is given the synthetic label
$y_{1:5:6} \in \mathcal{Y}'$, it ``loses'' the individual labels
$y_{1}$, $y_{5}$, $y_{6}$, which means that these latter classes end
up having fewer training examples than in the standard BR approach. In
other words, having more classes means having, on average, fewer
training examples per class, which may bring about lower accuracy for
the individual (non-synthetic) classes.
Yet another problem of the LP approach is the fact that combinations
of labels never found in the training set cannot be predicted at all.

The ``classifier chains'' (\textsc{CChain}) approach
\cite{Read:2011zc, Dembczynski:2010ei} consists instead of training a
\emph{sequence} of $n$ binary classifiers, one for each class in $n$,
chained so that each classifier receives, as additional inputs, the
predictions made by the previous classifiers in the chain. That is,
the first classifier $h_{1}$ in the chain is trained to predict label
$y_{1}$ using the original vector $\mathbf{x}$ as input, while the
$i$-th classifier in the chain is trained to predict label $y_{i}$
using as inputs the original vector $\mathbf{x}$ concatenated with a
vector of predictions
$(h_{1}(\mathbf{x}), \ldots, h_{i-1}(\mathbf{x}))$ for labels
$y_{1}, \ldots,y_{i-1}$, respectively. While this has turned out to be
one of the best-performing MLC methods~\cite{Madjarov:2012yl},
\textsc{CChain} presents a number of limitations. The first has to do
with the fact that \textsc{CChain} is sensitive to the order of the
classes; in other words, $\mathcal{Y}$ is treated as an \emph{ordered
sequence} (rather than a set) of classes, and reshuffling the sequence
would bring about different results, which is undesirable. Some
approaches to counter this limitation consist of training an ensemble
of
\textsc{CChain}s 
using different label orderings in each of them, and then implementing
some aggregation policy such as majority
voting~\cite{Read:2011zc}. Yet another limitation of \textsc{CChain}
is that the classification phase is (unlike for the simpler BR
strategy) not amenable to parallelization, since the application of
classifier $h_{i}$ to unlabelled data must occur strictly after the
application of classifiers $h_{1}$, ..., $h_{i-1}$.





\subsubsection{Algorithm Adaptation Approaches}
\label{sec:relwork:algadapt}

\noindent ``Algorithm adaptation'' approaches consist of adapting
single-label classifiers to return multi-label predictions.
%
%
Schapire and Singer proposed a multi-label adaptation of
\textsc{AdaBoost}, called \textsc{AdaBoost.MH}~\cite{Schapire:2000nl},
which is based on choosing, for a given iteration of the boosting
process, a unique ``pivot term'' around which all the binary
classifiers for the individual classes
hinge. \textsc{ML-knn}~\cite{Zhang:2007as} is an adaptation of the
well-known (lazy) KNN algorithm, which treats each class
independently, and relies on the maximum \emph{a posteriori} (MAP)
principle to return multi-label predictions, where the prior and
posterior probabilities needed to compute the MAP rule are estimated
on the training set.
Some attempts to adapt SVMs to the multi-label problem include
\textsc{RankSVM}~\cite{Elisseeff:2001iz, Xu:2013gy}, which solves an
empirical risk minimization problem framed as a task of ranking
involving $n$ hyperplanes, and \textsc{ML-TSVM}~\cite{Chen:2016bs,
Ai:2021yp}, based on Twin SVMs (TSVMs)~\cite{Kumar:2009tt}, i.e., SVMs
that, instead of looking for one separation hyperplane, look for a
pair 
of non-parallel separating hyperplanes.
Rastogi et al.~\cite{Rastogi:2022ty} noted that \textsc{ML-TSVM} does
not actually take into account the class-class correlations, and
propose a method called \emph{Multi-Label Minimax Probability Machine}
(\textsc{ML-MPM}), inspired by the ``kernel trick'' used in
SVMs. \textsc{ML-MPM} deals with class-class correlations by imposing
a regularization term on the predicted sets of classes
(``labelsets''), where this term takes into account the label
co-occurrence matrix as found in the training set. Other classical
methods that have been used within algorithm adaptation approaches
include decision trees (DT -- see, e.g.,~\cite{Vens:2008yr}) and
random forests (RF -- see, e.g.,~\cite{Madjarov:2012yl}).

Benites et al.~\cite{Benites:2010ix} proposed ML-ARAM and ML-FAM, two
multi-label extensions of neural networks based on \emph{Adaptive
Resonance Theory} (ART). The methods work by inferring hierarchical
relationships between classes in the codeframe. Although these systems
outperformed \textsc{ML-knn} in the experimental evaluation, it was
later noted that they suffer when facing datasets characterized by
large codeframes and high-dimensional feature
spaces~\cite{Benites:2015nq}, a common setting in text classification
endeavours. A variant of ML-ARAM, called \emph{Hierarchical
ARAM}~\cite{Benites:2015nq} was then proposed to mitigate this
problem; hierarchical ARAM works by clustering the prototypes that
ML-ARAM learns for the input datapoints, generating larger clusters
that can be more efficiently accessed at inference time.

\subsubsection{Ensemble-Based Approaches}
\label{sec:relwork:ensembles}

\noindent Ensemble-based approaches to MLC aim at improving model
performance by combining different base classifiers.
Some ensembles have been proposed as a response to specific problems
encountered in other \MLC\ systems.

A first class of such ensembles are based on a
\emph{divide-and-conquer} approach, according to which the set of
classes is first partitioned (using any clustering method), and the
smaller, cluster-specific multi-label problems are then solved
independently of each other. Different instances implementing this
principle exist in the literature. For example, the authors
of~\cite{Tsoumakas:2008ek} create a tree in which the internal nodes
are associated with sets of classes (called \emph{meta-classes}) and
their children are associated with subsets of the classes of the
parent node. The root represents the set of all classes, while the
leaves represent single classes. For each node, a multi-label
classifier is trained to predict (zero, one, or more) meta-classes,
each associated with one of the children nodes. The final set of
labels returned is the union of all labels in the leaf nodes
reached. \textsc{HOMER} implements the partitioning as balanced
clustering, while \textsc{HOMER-R} relies instead on $k$-means; in
both cases the smaller multi-label problems are tackled via the binary
relevance approach discussed in Section~\ref{sec:relwork:probtrans}.
\textsc{RakEL}~\cite{Tsoumakas:2011vp} instead relies on random
clustering (with the number of clusters a hyperparameter of the model)
for partitioning the classes, and then runs an instance of the label
powerset technique (LP -- discussed in
Section~\ref{sec:relwork:probtrans}) for each cluster. Since the
number of classes in each cluster is smaller than the total number of
classes, the combinatorial problem affecting the LP approach is
mitigated.
In a similar vein, \emph{Label Space Clustering} (LSC)
\cite{Szymanski:2016qj} uses $k$-means for clustering, followed by an
instance of \textsc{ML-knn} local to each cluster.


Other methods rely instead on \emph{label embeddings}, i.e., on
representing each class by means of a low-dimensional dense vector in
a continuous space, so that similar classes (i.e., classes that tend
to co-occur with each other) tend to be close to each other in the
embedding space. The \emph{Cost-Sensitive Label Embedding with
Multidimensional Scaling} (\textsc{CLEMS}) method~\cite{Huang:2017hw}
is one such example, and one that has proven to be among the best
performers in the label embedding arena (see
e.g.,~\cite{Szymanski:2017ru}).

Stacked Generalization (SG)~\cite{Wolpert:1992rq} has often been
employed to carry out multi-label classification. The idea is to train
an ensemble of binary classifiers, each for a different class (somehow
similarly to the BR approach) and use the classification predictions
as additional features to train a \emph{meta}-classifier. Some
variants following this intuition include the \textsc{Fun-TAT} and
\textsc{Fun-KFCV}~\cite{Esuli:2019dp} approaches for cross-lingual
\MLC, in which a language-agnostic meta-classifier receives as input
the predictions returned by language-specific multi-label
classifiers. An advantage of SG with respect to \textsc{CChain} is
that the former can be easily parallelized, since it is not dependent
on the order of presentation of the classes.

\section{An Evaluation Protocol for Testing Multi-Label Quantifiers}
\label{sec:mlapp}

\noindent For the evaluation of quantifiers, researchers often use the
same datasets that are elsewhere used for testing classifiers. On one
hand this looks natural, because both classification and
quantification deal with datapoints that belong to classes in a given
codeframe. On the other hand this looks problematic, since
classification deals with estimating class labels for individual
datapoints while quantification deals with estimating class prevalence
values for \emph{samples} (sets) of such datapoints. Simply estimating
the accuracy of a quantifier on the entire test set of a dataset used
for classification purposes (hereafter: a ``classification dataset'')
would not be enough, since this would be a single prediction only,
which would be akin to testing a classifier on a single datapoint
only. As a result, it is customary to generate a dataset to be used
for quantification purposes (a ``quantification dataset'') from a
classification dataset by extracting from the test set of the latter a
number of samples than will form the test set of the quantification
dataset. Exactly how these samples are extracted is specified by an
\emph{evaluation protocol}. Different evaluation protocols for the
binary case~\cite{Esuli:2015gh, Forman:2005fk}, for the single-label
multiclass case~\cite{Esuli:2022hy}, and for the ordinal
case~\cite{Bunse:2022dz}, have been proposed in the quantification
literature.

For the binary case,
the most widely adopted protocol is the so-called \emph{artificial
prevalence protocol} (APP)~\cite{Forman:2005fk}.  The APP consists of
extracting many samples from a set of test datapoints at controlled
prevalence values. The APP takes four parameters as input: the
unlabelled collection $U$, the sample size $k$, the number of samples
$m$ to draw for each predefined vector of prevalence values, and a
grid of prevalence values $\mathbf{g}$ (e.g.,
$\mathbf{g}=(0.0, 0.1, \ldots, 0.9, 1.0)$). We then generate all the
vectors $\mathbf{p}=(p(\oplus),p(\ominus))$ of $n=2$ prevalence values
consisting of combinations of values from the grid $\mathbf{g}$ that
represent valid distributions (i.e., such that the elements in
$\mathbf{p}$ sum up to 1). For each such prevalence vector, we then
draw $m$ different samples of $k$ elements each. The APP thus
confronts the quantifier with scenarios characterized by class
prevalence values very different from the ones seen during training.
This protocol is, by far, the most popular one in the quantification
literature (see, e.g.,~\cite{Forman:2005fk, Esuli:2018rm,
Maletzke:2019qd, Perez-Gallego:2017wt, Card:2018pb, Reis:2018fk,
Perez-Gallego:2019vl, Moreo:2022bf, Schumacher:2021ty, Vaz:2019eu}).

For the single-label multiclass case (which is the closest to our
concerns) the APP needs to take a slightly different form, since the
number of vectors $\mathbf{p}=(p(y_{1}), ..., p(y_{n}))$ representing
valid distributions for arbitrary $n$ is combinatorially high, for any
reasonable grid of class prevalence values. As a solution, one can
generate a number of random points on the probability simplex, with
the individual class prevalence values not even being constrained to
lie on a predetermined grid; when this number is high enough, it
probabilistically covers the entire spectrum of valid combinations.

However, even this form of the APP is not directly applicable to the
multi-label scenario, because in this latter the class prevalence
values in a valid vector do not necessarily sum up to 1. One could
attempt to simply treat the multi-label problem as a set of
independent binary problems, and then apply the APP independently to
each of the classes. Unfortunately, such a solution is impractical,
for at least three reasons.
\begin{itemize}

\item The first reason is that the number of samples thus generated is
  exponential in $n$, since there are
  $m|\mathbf{g}|^n$ 
  such combinations. Note that $n$ (the number of classes in the
  codeframe) cannot be set at will, and thus, in order to keep the
  number of combinations tractable in cases in which $n$ is large (in
  our experiments we use datasets with up to $n=983$ classes),
  one would be compelled to set $m=1$ and choose a very coarse grid
  $\mathbf{g}$ of values (this would anyway prove insufficient when
  dealing with large codeframes).

\item The second and perhaps most problematic reason is that, in any
  case, many of the combinations are not even realisable.
  That is, there may be prevalence vectors for which no sample could
  be drawn at all. To see why, assume that, among others, we have
  classes $y_{1}$, $y_{2}$, $y_{3}$ in our codeframe, and assume that
  in our test set $U$, every time a datapoint is labelled with $y_{1}$
  it is also labelled with either $y_{2}$ or $y_{3}$ but not
  both. This means that all samples $\sigma$ for which prevalence
  values $p_\sigma(y_{1})\neq (p_\sigma(y_{2}) + p_\sigma(y_{3}))$ are
  requested, cannot be
  generated. 

\item Yet another reason why applying the APP would be, in any case,
  undesirable, is that the classes in most multi-label datasets
  typically follow a power-law distribution, i.e., there are few very
  popular classes and a long tail of many rare, or extremely rare,
  classes. The APP will sometimes impose high prevalence values for
  what in reality are very rare labels, which means that the sampling
  must be carried out \emph{with replacement}, which generates samples
  consisting of many replicas of the same few datapoints, which is
  clearly
  undesirable. 
\end{itemize}
\noindent For all these reasons we have designed a brand new
protocol 
for \MLQ, that we call ML-APP, since it is an adaptation of the APP to
multi-label datasets. The protocol amounts to performing rounds of the
APP, each targeting a specific class, but with the range of prevalence
values explored for each class being limited by the amount of
available positive examples. This allows all samples to be drawn
\emph{without} replacement. In each round, a class $y_{i}$ is actively
sampled at controlled prevalence values while the prevalence values
for the remaining classes are not predetermined. Pseudocode describing
the ML-APP protocol is shown as Algorithm~\ref{alg:mlapp}.

\begin{algorithm}[ht]
  \caption{\label{alg:mlapp}ML-APP protocol for multi-label data.}
  \begin{algorithmic}
    \State \textbf{Input:} $U$, a test collection \State
    \textbf{Input:} $k$, the sample size \State \textbf{Input:} $m$,
    the number of samples to draw for each prevalence \State
    \textbf{Input:} $\mathbf{g}$, the grid of prevalence values \State
    $S = \{\}$ \For{$y_{i}\in\mathcal{Y}$} \codecomment{Split $U$ in
    two sets, with $U_{y_{i}}$ containing all datapoints with label
    $y_{i}$ and $U_{\overline{y}_{i}}$ \newline \mbox{} \hspace{1.0em}
    // containing the others} \State
    {$U_{y_{i}}\leftarrow\{(\mathbf{x},Y)\in U : y_{i}\in Y\}$} \State
    {$U_{\overline{y}_{i}}\leftarrow\{(\mathbf{x},Y)\in U :
    y_{i}\notin Y\}$} \For{$g_{j} \in \mathbf{g}$}
    \codecomment{Compute the number of positives and negatives to
    extract} \State \textsc{Pos}
    $\leftarrow \lceil k \cdot g_{j} \rceil$ \State \textsc{Neg}
    $\leftarrow k - $\textsc{Pos} \codecomment{Generate samples only
    if the number of datapoints in $U_{y_{i}}$ allows}
    \codecomment{the sampling to be performed without replacement}
    \If{ $|U_{y_{i}}| \geq $ \textsc{Pos} } \For{$m$ repetitions}
    \State draw $\sigma_{y_{i}}$ from $U_{y_{i}}$, with
    $|\sigma_{y_{i}}|=$\textsc{Pos}, uniformly at random w/o
    replacement \State draw $\sigma_{\overline{y}_{i}}$ from
    $U_{\overline{y}_{i}}$, with
    $|\sigma_{\overline{y}_{i}}|=$\textsc{Neg}, uniformly at random
    w/o replacement \codecomment{Members of $\sigma_{y_{i}}$ and
    $\sigma_{\overline{y}_{i}}$ are not removed from $U_{y_{i}}$ or
    $U_{\overline{y}_{i}}$} \State
    $\sigma \leftarrow \sigma_{y_{i}} \cup \sigma_{\overline{y}_{i}}$
    \aftercodecomment{Note that $p_{\sigma}(y_{i})=g_{j}$; the
    prevalence for the other \newline \mbox{} \hspace{12.2em} //
    classes is not predetermined} \State
    $S\leftarrow S \cup \{\sigma\}$ \EndFor \EndIf \EndFor \EndFor
    \State \textbf{Return:} $S$
  \end{algorithmic}
\end{algorithm}


The ML-APP covers the entire spectrum of class prevalence values, by
drawing without replacement, for every single class. This means that,
for large enough classes, there will be samples for which the
prevalence of the class exhibits a large prior probability shift with
respect to the training prevalence, while for rare classes the amount
of shift will be limited by the availability of positive
examples. Note that, when actively sampling a class $y_{i}$, any other
class $y_{j}$ will co-occur with it with a probability that depends on
the correlation between $y_{i}$ and $y_{j}$. For cases in which the
class $y_{i}$ being sampled is completely independent of the class
$y_{j}$, the samples generated will display a class prevalence for
$y_{j}$ that is approximately similar to the prevalence of $y_{j}$ in
$U$. In other words, samples generated via the ML-APP preserve the
stochastic correlations between the classes while also exhibiting
widely varying degrees of prior probability shift. Finally, note that
the total number of samples that can be generated via the ML-APP can
vary from dataset to dataset (even if they have the same number of
classes), and depends on the actual number of positive instances for
each class that are contained in the dataset. In any case, the maximum
number of samples that can be generated via the ML-APP is bounded by
$m n |\mathbf{g}|$.



\section{Performing Multi-Label Quantification}
\label{sec:quant}

\noindent In this section we present the multi-label quantification
methods that we will experimentally compare in
Section~\ref{sec:experiments}. Throughout this paper we will focus on
\emph{aggregative} quantification methods, i.e., methods that require
all unlabelled datapoints to be classified (by a hard or a soft
classifier, depending on the method) as an intermediate step, and that
aggregate the individual (hard or soft) predictions in some way to
generate the class prevalence estimates. The reason why we focus on
aggregative methods is that they are by far the most popular
quantification methods in the literature, and that this focus allows
us an easier exposition. We will later show how the most interesting
intuitions for performing MLQ that we discuss in this paper also apply
to the non-aggregative case.

Before presenting truly multi-label quantifiers, though, we will
introduce a number of single-label (aggregative) multiclass
quantification methods from the literature, that will form the basis
for our extensions to the MLQ case.


\subsection{Single-Label Multiclass Quantification Methods}
\label{sec:quant:sl}

\noindent \emph{Classify and Count} (CC), already hinted at in the
introduction, is the na\"ive quantification method, and the one that
is used as a baseline that all genuine quantification methods are
supposed to beat. Given a hard classifier $h$ and a sample $\sigma$,
CC is formally defined as
\begin{equation}\label{eq:CC}
  \begin{aligned}
    \hat{p}_{\sigma}^{\mathrm{CC}}(y_{i}) & = \frac{|\{\mathbf{x}\in
    \sigma|h(\mathbf{x})=y_{i}\}|}{|\sigma|}
  \end{aligned}
\end{equation}
\noindent In other words, the prevalence of a class $y_{i}$ is
estimated by classifying all unlabelled datapoints, counting the
number of datapoints that have been assigned to $y_{i}$, and dividing
the result by the total number of datapoints.

The \emph{Adjusted Classify and Count} (ACC) method
(see~\cite{Forman:2008kx, Vaz:2019eu}) attempts to correct the
estimates returned by CC by relying on the law of total probability,
according to which
%
\begin{align}
  \label{eq:ACC} 
  p(h(\mathbf{x})=y_{i}) = \sum_{y_{j}\in
  \mathcal{Y}}p(h(\mathbf{x})=y_{i}|y_{j})\cdot p(y_{j})
\end{align}
\noindent which can be more conveniently rewritten using matrix
notation as
\begin{align}
  \label{eq:ACC2} 
  \mathbf{p}^{\mathrm{CC}}_{\sigma} = \mathbf{M}_h
  \cdot \mathbf{p}^{\mathrm{ACC}}_{\sigma}
\end{align}
\noindent where $\mathbf{p}^{\mathrm{CC}}_{\sigma}$ is the vector
representing the distribution across $\mathcal{Y}$ of the datapoints
as estimated via CC, and matrix $\mathbf{M}_h$ contains the
misclassification rates of $h$, i.e., $m_{ij}$ is the probability that
$h$ will assign class $y_{i}$ to a datapoint whose true label is
$y_{j}$. Matrix $\mathbf{M}_h$ is unknown, but can be estimated via
$k$-fold cross-validation, or on a validation set. Vector
$\mathbf{p}^{\mathrm{ACC}}_\sigma$ is the true distribution; it is
unknown, and the ACC method consists of estimating it by solving the
system of linear equations of Equation \ref{eq:ACC2} (see
\cite{Bunse:2022oj} for more on the multiclass version of ACC).

While CC and ACC rely on the crisp counts returned by a hard
classifier $h$, it is possible to define variants of them that use
instead the expected counts computed from the posterior probabilities
returned by a calibrated probabilistic classifier $s$
\cite{Bella:2010kx}. This is the core idea behind \emph{Probabilistic
Classify and Count} (PCC) and \emph{Probabilistic Adjusted Classify
and Count} (PACC).
PCC is defined as
\begin{align}
  \begin{split}\label{eq:PCC}
    \hat{p}_{\sigma}^{\mathrm{PCC}}(y_{i}) & = \frac{1}{|\sigma|}\sum_{\mathbf{x}\in \sigma}[s(\mathbf{x})]_{i} \\
    & = \frac{1}{|\sigma|}\sum_{\mathbf{x}\in
    \sigma}\Pr(y_{i}|\mathbf{x})
  \end{split}
\end{align}
\noindent while PACC is defined as
\begin{align}
  \label{eq:PACC2} 
  \mathbf{p}^{\mathrm{PCC}}_{\sigma} = \mathbf{M}_s
  \cdot \mathbf{p}^{\mathrm{PACC}}_{\sigma}
\end{align}
\noindent Equation~\ref{eq:PACC2} is identical to
Equation~\ref{eq:ACC2}, but for the fact that the leftmost part is
replaced with the prevalence values estimated via PCC, and for the
fact that the misclassification rates of the soft classifier $s$
(i.e., the rates computed as expectations using the posterior
probabilities) are used.

Methods CC, ACC, PCC, PACC, are sometimes collectively referred to as
the ``CC variants'', and are all (as it is easy to verify) aggregative
quantification methods.
%
Although more sophisticated quantification systems have been proposed
in the literature, the CC variants have recently been found to be
competitive contenders when properly
optimized~\cite{Moreo:2021sp}. This, along with their simplicity, has
motivated us to focus on the four CC variants as a first step towards
devising multi-label quantifiers.

A further, very popular (aggregative) quantification method is the one
proposed in~\cite{Saerens:2002uq}, which is often called SLD, from the
names of its proposers, and which was called EMQ
in~\cite{Gao:2016uq}. SLD was the best performer in a recent data
challenge centred on quantification~\cite{Esuli:2022hy}, and consists
of training a probabilistic classifier and then using the EM algorithm
(i) to update the posterior probabilities that the classifier returns,
and (ii) to re-estimate the class prevalence values of the test
set. Steps (i) and (ii) are carried out in an iterative, mutually
recursive way, until mutual consistency, defined as the situation in
which
\begin{align}
  \label{eq:calib} 
  \hat{p}_{\sigma}(y_{i}) & \approx \sum_{\mathbf{x}\in \sigma}\Pr(y_{i}|\mathbf{x})
\end{align}  
\noindent is achieved for all $y_{i}\in \mathcal{Y}$.


\subsection{Multi-Label Quantification}
\label{sec:quant:ml}

\noindent In this paper we will describe and compare many different
(aggregative) MLQ methods.  In order to better assess their relative
merits, we subdivide them into four different groups, depending on
whether the correlations between different classes are brought to bear
in the classification phase (i.e., by the classifier on which an
aggregative quantifier rests), or in the aggregation phase (i.e., in
the phase in which the individual predictions are aggregated), or in
both phases, or in neither of the two phases.

%
%

The first and simplest such group is that of \MLQ\ methods that treat
each class as completely independent, and thus solve $n$ independent
binary quantification problems. We call such an approach \bqbc\
(``binary classification followed by binary aggregation''), since in
both the classification phase and the aggregation phase it looks at
the multi-label task as $n$ independent binary tasks, thus
disregarding, in both phases, the correlations among classes when
predicting their relative frequencies. This is akin to the binary
relevance (BR) problem transformation described in
Section~\ref{sec:relwork:probtrans} for classification, and consists
of transforming the multi-label dataset $L$ into a set of binary
datasets $L_{1}, \ldots, L_{n}$ in which
$L_{i}=\{(\mathbf{x},\indfn{y_{i}\in Y}):(\mathbf{x},Y)\in L\}$ is
labelled according to $\mathcal{Y}_{i}=\{\mathbf{0},\mathbf{1}\}$,
since the datapoints are relabelled using the indicator function
$\indfn{z}$ that returns $\mathbf{1}$ (the ``positive class'') if $z$
is true or $\mathbf{0}$ (the ``negative class'') otherwise. \bqbc\
methods then train one quantifier $q_{i}$ for each training set
$L_{i}$. At inference time, the prevalence vector for a given sample
$\sigma$ is computed as
$\mathbf{p}_{\sigma}^{\bqbc}=(p_{\sigma}^{q_{1}}(\mathbf{1}),p_{\sigma}^{q_{2}}(\mathbf{1}),\ldots,p_{\sigma}^{q_{n}}(\mathbf{1}))$.
Although this is technically a multi-label quantification
method, 
\bqbc\ is actually the trivial solution that we expect any truly
multi-label quantifier to improve
upon. 

A second, less trivial group is that of \MLQ\ methods based on the use
of binary aggregative quantifiers working on top of (truly)
multi-label classifiers. Methods in this group consist of $n$
independent binary aggregative quantifiers (e.g., built via one of the
methods described in Section~\ref{sec:quant:sl}) that rely on the
(hard or soft) predictions returned by a classifier natively designed
to tackle the multi-label problem (e.g., built via one of the methods
described in Section~\ref{sec:relwork}). Each binary quantifier takes
into account only the predictions for its associated class,
disregarding the predictions for the other classes. This represents a
straightforward solution to the \MLQ\ problem, as it simply combines
already existing technologies (binary aggregative quantifiers built
via off-the-shelf methods and (truly) multi-label classifiers built
via off-the-shelf methods). In such a setting, the classification
stage is informed by the class-class correlations, but the
quantification methods in charge of producing the class prevalence
estimates for each class do not pay attention to any such correlation,
and are disconnected from each other. Since methods in this group will
consist of a (truly) multi-label classification phase followed by a
binary quantification phase, we will refer to this group of methods as
\bqmc.

We next propose a third group of \MLQ\ systems, i.e., ones consisting
of natively multi-label quantification methods relying on the outputs
of $n$ independent binary classifiers.
%
Methods like these represent a non-trivial novel solution for the
field of quantification, because no natively multi-label
quantification method has been proposed so far in the literature; in
Section~\ref{sec:quant:ml:reg} we propose some such methods. In order
to clearly evaluate the merits of such a multi-label aggregation
phase, as the underlying classifiers we use independent binary
classifiers only. For this reason, we will call this group of methods
\mqbc.

The fourth and last group of methods we consider amounts to combining
any (truly) multi-label classification method with any (truly)
multi-label quantification method among our newly proposed ones, thus
bringing to bear the class dependencies both at the classification
stage and at the aggregation stage. We call this group of methods
\mqmc.

\begin{figure}
  \centering
  \caption{The four groups of multi-label quantification methods. Dotted lines
  connecting class labels with a model (classifier or quantifier)
  indicate that the model learns from (or has access to) the class
  labels of the training datapoints. Solid lines connecting
  classifiers with quantifiers indicate a transfer of outputs from the
  classifier to the quantifier. With a slight deviation from our
  notation, here $h$ denotes any classifier, hard or soft.}
  \label{fig:schema}
  \usetikzlibrary{shapes.geometric, arrows,arrows.meta,matrix}

\tikzstyle{floating} = [text centered]
\tikzstyle{coltitle} = [anchor=west]
\tikzstyle{mintitle} = [anchor=west,font=\scriptsize]
\tikzstyle{label} = [text centered, font=\bfseries]
\tikzstyle{clf} = [rectangle, rounded corners, text centered, draw=black]
\tikzstyle{q} = [circle, text centered, draw=black]

\tikzstyle{dependency} = [thick, -{Latex[round,open]}, dash pattern= on 1pt off 1pt]
\tikzstyle{flow} = [thick, ->]

\begin{tikzpicture}
	\node(y1_top)[label]{$y_1$};
	\node(y2)[label, right of=y1_top]{$y_2$};
    \node(hdots)[floating, right of=y2]{...};
	\node(yn_top)[label, right of=hdots]{$y_n$};

    \node(h1)[clf, below of=y1_top, yshift=-5mm]{$h_1$};
    \node(h2)[clf, right of=h1]{$h_2$};
    \node(hdots)[floating, right of=h2]{...};
    \node(hn)[clf, right of=hdots]{$h_n$};

	\draw [dependency] (y1_top) -- (h1);
	\draw [dependency] (y2) -- (h2);
	\draw [dependency] (yn_top) -- (hn);

	\node(q1)[q, below of=h1, yshift=-5mm]{$q_1$};
	\node(q2)[q, right of=q1]{$q_2$};
	\node(hdots)[floating, right of=q2]{...};
	\node(qn)[q, right of=hdots]{$q_n$};

	\draw [flow] (h1) -- (q1);
	\draw [flow] (h2) -- (q2);
	\draw [flow] (hn) -- (qn);

	\node(y1_bot)[label, below of=q1, yshift=-5mm]{$y_1$};
	\node(y2)[label, right of=y1_bot]{$y_2$};
    \node(hdots)[floating, right of=y2]{...};
	\node(yn)[label, right of=hdots]{$y_n$};

	\draw [dependency] (y1_bot) -- (q1);
	\draw [dependency] (y2) -- (q2);
	\draw [dependency] (yn) -- (qn);

	\node(y1_top2)[label, right of=yn_top,xshift=2cm]{$y_1$};
	\node(y2)[label, right of=y1_top2]{$y_2$};
    \node(hdots)[floating, right of=y2]{...};
	\node(yn)[label, right of=hdots]{$y_n$};

    \node(h)[clf, below right of=y2, yshift=-8mm]{$h$};

	\draw [dependency] (y1_top2) -- (h);
	\draw [dependency] (y2) -- (h);
	\draw [dependency] (yn) -- (h);

	\node(q2)[q, below left of=h, yshift=-8mm]{$q_2$};
	\node(q1)[q, left of=q2]{$q_1$};
	\node(hdots)[floating, right of=q2]{...};
	\node(qn)[q, right of=hdots]{$q_n$};

	\draw [flow] (h) -- (q1);
	\draw [flow] (h) -- (q2);
	\draw [flow] (h) -- (qn);

	\node(y1)[label, below of=q1, yshift=-5mm]{$y_1$};
	\node(y2)[label, right of=y1]{$y_2$};
    \node(hdots)[floating, right of=y2]{...};
	\node(yn)[label, right of=hdots]{$y_n$};

	\draw [dependency] (y1) -- (q1);
	\draw [dependency] (y2) -- (q2);
	\draw [dependency] (yn) -- (qn);

	\node(y1_top3)[label, below of=y1_bot,yshift=-5mm]{$y_1$};
	\node(y2)[label, right of=y1_top3]{$y_2$};
    \node(hdots)[floating, right of=y2]{...};
	\node(yn_top)[label, right of=hdots]{$y_n$};

    \node(h1)[clf, below of=y1_top3, yshift=-5mm]{$h_1$};
    \node(h2)[clf, right of=h1]{$h_2$};
    \node(hdots)[floating, right of=h2]{...};
    \node(hn)[clf, right of=hdots]{$h_n$};

	\draw [dependency] (y1_top3) -- (h1);
	\draw [dependency] (y2) -- (h2);
	\draw [dependency] (yn_top) -- (hn);

	\node(q)[q, below right of=h2, yshift=-5mm]{$q$};

	\draw [flow] (h1) -- (q);
	\draw [flow] (h2) -- (q);
	\draw [flow] (hn) -- (q);

	\node(y2)[label, below left of=q, yshift=-5mm]{$y_2$};
	\node(y1)[label, left of=y2]{$y_1$};
    \node(hdots)[floating, right of=y2]{...};
	\node(yn)[label, right of=hdots]{$y_n$};

	\draw [dependency] (y1) -- (q);
	\draw [dependency] (y2) -- (q);
	\draw [dependency] (yn) -- (q);

	\node(y1_top4)[label, right of=yn_top,xshift=2cm]{$y_1$};
	\node(y2)[label, right of=y1_top4]{$y_2$};
    \node(hdots)[floating, right of=y2]{...};
	\node(yn)[label, right of=hdots]{$y_n$};

    \node(h)[clf, below right of=y2, yshift=-5mm]{$h$};

	\draw [dependency] (y1_top4) -- (h);
	\draw [dependency] (y2) -- (h);
	\draw [dependency] (yn) -- (h);

	\node(q)[q, below of=h, yshift=-5mm]{$q$};

	\draw [flow] (h) -- (q);

	\node(y2)[label, below left of=q, yshift=-5mm]{$y_2$};
	\node(y1)[label, left of=y2]{$y_1$};
    \node(hdots)[floating, right of=y2]{...};
	\node(yn)[label, right of=hdots]{$y_n$};

	\draw [dependency] (y1) -- (q);
	\draw [dependency] (y2) -- (q);
	\draw [dependency] (yn) -- (q);

	\node(bc)[coltitle,above right of=y1_top,xshift=8mm,yshift=5mm]{Binary Classification};
	\node(mc)[coltitle, right of=bc, xshift=50mm]{Multi-Label Classification};
	\node(bq)[coltitle,rotate=90, above left of=bc,anchor=east,xshift=-10mm,yshift=25mm]{Binary Aggregation};
	\node(mq)[coltitle, rotate=90, anchor=east, left of=bq, xshift=-48mm]{Multi-Label Aggregation};

	\node(bqbc)[mintitle,above of=y1_top,xshift=15mm,yshift=-5mm]{\bqbc};
	\node(bqmc)[mintitle,above of=y1_top2,xshift=15mm,yshift=-5mm]{\bqmc};
	\node(mqbc)[mintitle,above of=y1_top3,xshift=15mm,yshift=-5mm]{\mqbc};
	\node(mqmc)[mintitle,above of=y1_top4,xshift=15mm,yshift=-5mm]{\mqmc};
\end{tikzpicture}
\end{figure}
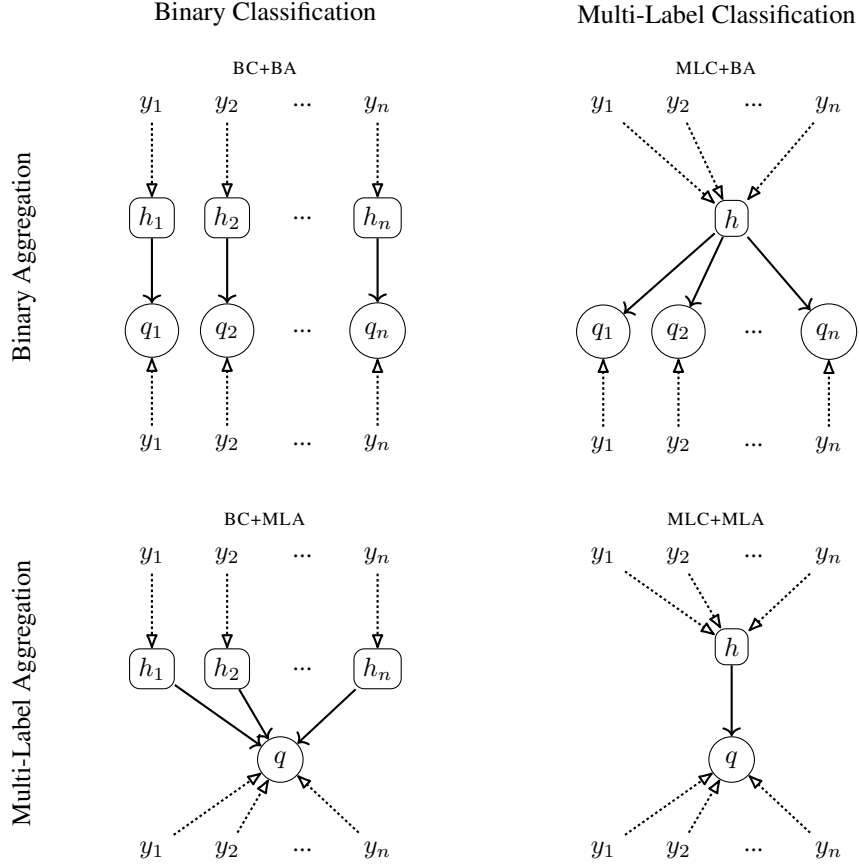

Figure~\ref{fig:schema} illustrates in diagrammatic form the four
types of multi-label quantification methods we study in this paper. In
order to generate members of these four classes, we already have
off-the-shelf components for implementing the binary classification,
multi-label classification, and binary aggregation phases, but we have
no known method from the literature to implement multi-label
aggregation; Sections~\ref{sec:quant:ml:reg} and \ref{sec:quant:ml:lp}
are devoted to proposing two novel such methods.
%


\subsubsection{Bringing to Bear Class-Class Correlations at the
Aggregation Stage via Regression}
\label{sec:quant:ml:reg}

\noindent Let us assume we have a multi-label quantifier $q$ of type
\bqbc\ or \bqmc. Our idea is to detect how quantifier $q$ fails in
capturing the correlations between classes,
and correct $q$ accordingly.
This is somehow similar to the type of correction implemented in ACC
and PACC.  However, we will formalize this intuition as a general
regression problem, thus not necessarily assuming this correction to
be linear (as ACC and PACC instead do).

More concretely, we split our training set $L$ into two parts, $L_{Q}$
(that we use for training our quantifier $q$) and $L_{R}$ (that we use
for training a regressor $r$, i.e., a function
$r : \mathbb{R}^{n} \rightarrow
\mathbb{R}^{n}$).\footnote{\protect\label{foot:iterativestrat}Note
that, for reasons discussed in~\cite{Sechidis:2011tu,
Szymanski:2017yc}, multi-label datasets cannot be split in a
stratified way using standard algorithms for single-label
stratification. For splitting the training set $L$, we thus use the
\emph{iterative stratification} method implemented in
\scikitml\footnote{\url{http://scikit.ml/stratification.html}} and
described in~\cite{Szymanski:2017yc}.} We then use the ML-APP protocol
described in Section~\ref{sec:mlapp} to extract, from set $L_{R}$, a
set
$\mathcal{R}=\{\sigma_{i} \sim \operatorname{ML-APP}(L_{R}, k, m,
\mathbf{g})\}$ of $l$ samples, where $k$ (sample size), $m$ (number of
samples to draw for each prevalence value on the grid), and
$\mathbf{g}$ (grid of prevalence values) are the parameters of the
ML-APP protocol.
%
%
%
Having done this, we first train our quantifier $q$ on $L_{Q}$; since
$q$ is a multi-label quantifier, it is a function that, given a sample
$\sigma$, returns a vector $\hat{\mathbf{p}}^q_{\sigma}$ of $n$ class
prevalence values, not necessarily summing up to
1. 
We then apply $q$ to all the samples in $\mathcal{R}$; as a result,
for each sample $\sigma_{i}\in\mathcal{R}$ we have a pair
$(\hat{\mathbf{p}}^q_{\sigma_{i}}, \mathbf{p}_{\sigma_{i}})$, where
$\hat{\mathbf{p}}^q_{\sigma_{i}}$ is the vector of the $n$ prevalence
values estimated by $q$, and $\mathbf{p}_{\sigma_{i}}$ is the vector
of the $n$ true prevalence values. We use this set of $l$ pairs as the
training set for training a multi-output regressor
$r : \mathbb{R}^{n} \rightarrow \mathbb{R}^{n}$ that takes as input a
vector of $n$ ``uncorrected'' prevalence values and returns a vector
of $n$ ``corrected'' prevalence values; for training the regressor we
use any off-the-shelf multi-output regression algorithm. Note that the
regressor indeed captures the correlations between classes, since it
receives as input, for each sample, the class prevalence estimates for
all the $n$ classes altogether.\footnote{The fact that the regressor
captures the correlations between classes does not depend on it being
a multi-output regressor; even a set of $n$ single-output regressors
would obtain the same effect, which is only due to the fact that the
regressor takes as input all the $n$ class prevalence estimates at the
same time.
%
}


At inference time, given an (unlabelled) sample $\sigma$, we first
obtain a preliminary estimate of the class prevalence values
$\hat{\mathbf{p}}^q_\sigma$ via $q$, and then apply the correction
learned by $r$, thus computing
$\hat{\mathbf{p}}^r_\sigma=r(\hat{\mathbf{p}}^q_\sigma)$. We then
normalize, via clipping,\footnote{Clipping a value $v$ to the interval
$[a,b]$ amounts to returning $v$ if $v\in[a,b]$, $a$ if $v<a$, or $b$
if $v>b$. This is needed since, in principle, the regressor might
sometimes return values that fall outside the [0,1] interval.} every
prevalence value in $\hat{\mathbf{p}}^r_\sigma$ so that it falls in
the $[0,1]$ interval, and return the estimate.

The method (which we here call RQ, for ``\underline{r}egression-based
\underline{q}uantification'') is described succinctly as
Algorithm~\ref{alg:reg:train} (training phase) and
Algorithm~\ref{alg:reg:test} (inference
phase). 

\begin{minipage}{\textwidth}
  \begin{minipage}{.48\textwidth}
    \vspace{-1ex} \footnotesize
    \begin{algorithm}[H]
      \caption{\MLQ\ correction via regression:
      Training}\label{alg:reg:train}
      \begin{algorithmic}
        \State \textbf{Input:} $L$, a training collection \State
        \textbf{Input:} $q$, a multi-label quantifier \State
        \textbf{Input:} $r$, a multi-output regressor \State
        \textbf{Input:} $k$, $m$, $\mathbf{g}$, parameters of the
        ML-APP \State
        $L_Q, L_{R} = \operatorname{iterative-stratification}(L)$
        \State train $q$ on $L_Q$ \State
        $\mathcal{R}=\{(\hat{\mathbf{p}}^q_{\sigma_{i}},
        \mathbf{p}_{\sigma_{i}})_{i=1}^l : \sigma_{i} \sim
        \operatorname{ML-APP}(L_{R}, k, m, \mathbf{g})\}$ \State train
        $r$ on $\mathcal{R}$ \State \textbf{Return:} $q, r$
      \end{algorithmic}
    \end{algorithm}
    \vspace{3ex}
  \end{minipage}
  \hfill
  \begin{minipage}{.48\textwidth}
    \vspace{-1ex} \footnotesize
    \begin{algorithm}[H]
      \caption{\MLQ\ correction via regression: Inference
      \vspace{.5ex} \mbox{}}\label{alg:reg:test}
      \begin{algorithmic}
        \vfill \State \textbf{Input:} $\sigma$, an unlabelled sample
        \State \textbf{Input:} $q$, a trained multi-label quantifier
        \State \textbf{Input:} $r$, a trained multi-output regressor
        \Statex \State
        $\hat{\mathbf{p}}^q_\sigma \leftarrow q(\sigma)$ \State
        $\hat{\mathbf{p}}^r_\sigma \leftarrow
        \operatorname{clip}(r(\hat{\mathbf{p}}^q_\sigma), [0,1])$
        \Statex
        \State \textbf{Return:} $\hat{\mathbf{p}}^r_\sigma$ \\
        \mbox{}
      \end{algorithmic}
    \end{algorithm}
    \vspace{4ex}
  \end{minipage}
\end{minipage}

\medskip


As noted above, the regressor exploits the class-class correlations
during the aggregation phase. This means that, according to the
subdivision of MLQ methods illustrated in Table~\ref{fig:schema}, the
addition of a regression layer on top of an existing quantifier $q$
has the effect of transforming a \bqbc\ method into a \mqbc\ method,
or of transforming a \bqmc\ method into a \mqmc\ method.

\subsubsection{Bringing to Bear Class-Class Correlations at the
Aggregation Stage via Label Powersets}
\label{sec:quant:ml:lp}

\noindent We investigate an alternative way of modelling class-class
correlations at the quantification level, this time by gaining
inspiration from label powersets (LPs -- see
Section~\ref{sec:relwork:probtrans}) and the heuristics for making
their application tractable (Section~\ref{sec:relwork:ensembles}).

LP is a problem transformation technique devised for transforming any
multi-label classification problem into a single-label one by
replacing the original codeframe with another one that encodes subsets
of this codeframe into ``synthetic'' classes (see
Section~\ref{sec:relwork:probtrans} for details).
This problem transformation is directly applicable to the case of
quantification as well. Of course, the combinatorial explosion of the
number of synthetic classes has to be controlled somehow but,
fortunately enough, the same heuristics investigated for \MLC\ can
come to the rescue here.

Our method (which we here call LPQ, for ``\underline{l}abel
\underline{p}owerset -based \underline{q}uantification'') consists of generating, by
means of any existing clustering algorithm, a set $\mathcal{C}$ of
(non-overlapping) clusters consisting of few classes each, before
applying the LP strategy, so that the number of possible synthetic
classes remains under reasonable bounds.
For example, if our codeframe has $n=100$ classes,
extracting 25 clusters of 4 classes each results in the maximum
possible number of synthetic classes being $25 \cdot 2^4=400$, which
is much smaller than the original $2^{100}$.
We perform this clustering by treating classes in $\mathcal{Y}$ as
instances and training datapoints as features, so that a class is
represented by a binary vector of datapoints, where 1 indicates that
the datapoint belongs to the class and 0 that it does not.  The
clustering algorithm is thus expected to put classes displaying
similar assignment patterns (i.e., that tend to label the same
documents) in the same cluster.



Once we have performed the clustering, given the set of classes
$\mathcal{Y}_c \subseteq \mathcal{Y}$ contained in each cluster
$c\in\mathcal{C}$, we need to take the single-label codeframe
$\mathcal{Y}'_c$ determined from the
$2^{\mathcal{Y}}\rightarrow \mathcal{Y}'$ multi-label-to-single-label
mapping (a mapping that, e.g., would attribute to the set of classes
$\{y_{1},y_{5},y_{6}\}\subseteq \mathcal{Y}_{c}$ the synthetic class
$y_{1:5:6} \in \mathcal{Y}'_{c}$) and train a single-label quantifier
on it; this needs to be repeated for each
cluster. 
At inference time, in order to provide class prevalence estimates for
the classes in $\mathcal{Y}_{c}$ from the predictions made for the
classes in $\mathcal{Y}'_{c}$ by the above-mentioned quantifier, we
have to ``reverse'' the multi-label-to-single-label mapping, so that
the estimated prevalence value of $y_{i}\in\mathcal{Y}_{c}$ is the sum
of the estimated prevalence values of all labels
$y'\in\mathcal{Y}'_{c}$ that involve $y_{i}$; performing this for each
cluster $c\in\mathcal{C}$ returns prevalence estimates for all classes
$y_{i}\in\mathcal{Y}$.

More formally, let us define a matrix $\mathbf{A}$ that records the
label assignment in cluster $c$, so that $a_{ij}=1$ if the set of
classes represented by the synthetic class $y'_{i}\in\mathcal{Y}'_c$
contains class $y_{j}\in\mathcal{Y}_c$, and $a_{ij}=0$ if this is not
the case. Note that $\mathbf{A}$ has as many rows as there are classes
in $\mathcal{Y}'_c$ and as many columns as there are classes in
$\mathcal{Y}_c$. Once our single-label quantifier $q$ produces an
output $\hat{\mathbf{p}}^q_\sigma$, we only need to compute the
product $(\hat{\mathbf{p}}^q_\sigma)^\top \mathbf{A}$ to obtain the
vector of prevalence estimates for the classes in
$\mathcal{Y}_{c}$. Performing all this for each cluster
$c\in\mathcal{C}$ returns prevalence estimates for all classes
$y_{i}\in\mathcal{Y}$. The example shown in Figure~\ref{fig:example}
may clarify things.
\begin{figure}[ht]
  \caption{An example considering a cluster made of three classes only
  (left), and the computations carried out for reconstructing the
  prevalence values for the original multi-label codeframe
  (right).\newline\mbox{}}
  \label{fig:example}
  \begin{subfigure}{0.45\textwidth}
    \centering \resizebox{\textwidth}{!}{%
    \begin{tabular}{|c|c|c|c|c|l|r|}
      \cline{2-4} \cline{6-7}
      \multicolumn{1}{c|}{} & \multicolumn{3}{c|}{$\mathcal{Y}_c$} & & \multicolumn{1}{c|}{\multirow{2}{*}{$\mathcal{Y}'_c$}} & \multicolumn{1}{c|}{\multirow{2}{*}{$\mathbf{\hat{p}}^q_\sigma$}}\\\cline{2-4}
      \multicolumn{1}{c|}{} & $y_1$ & $y_{2}$ & $y_{3}$ & & & \\\cline{1-4} \cline{6-7}
      \multirow{8}{*}{$\mathbf{A}$:} 
                            & 0 & 0 & 0 & & $y'_\emptyset$ & $\hat{p}^q_\sigma(y'_\emptyset)=0.15$\\
                            & 1 & 0 & 0 & & $y'_{1}$ & $\hat{p}^q_\sigma(y'_{1})=0.10$ \\
                            & 0 & 1 & 0 & & $y'_{2}$ & $\hat{p}^q_\sigma(y'_{2})=0.26$ \\
                            & 1 & 1 & 0 & & $y'_{1:2}$ & $\hat{p}^q_\sigma(y'_{1:2})=0.19$ \\
                            & 0 & 0 & 1 & & $y'_3$ & $\hat{p}^q_\sigma(y'_3)=0.05$ \\
                            & 1 & 0 & 1 & & $y'_{1:3}$ & $\hat{p}^q_\sigma(y'_{1:3})=0.13$ \\
                            & 0 & 1 & 1 & & $y'_{2:3}$ & $\hat{p}^q_\sigma(y'_{2:3})=0.11$ \\
                            & 1 & 1 & 1 & & $y'_{1:2:3}$ & $\hat{p}^q_\sigma(y'_{1:2:3})=0.01$ \\
      \cline{1-4} \cline{6-7}
    \end{tabular}
    }%
  \end{subfigure}%
  \begin{subfigure}{0.5\textwidth}
    \centering
    \begin{tabular}{ll}
    
      \multicolumn{2}{l}{$(\hat{p}^q_\sigma(y_{1}),\hat{p}^q_\sigma(y_{2}),\hat{p}^q_\sigma(y_3))=(\mathbf{\hat{p}}^q_\sigma)^\top \mathbf{A}$} \\ 
      \\
      $\hat{p}^q_\sigma(y_{1})$ & = $\hat{p}^q_\sigma(y'_{1})+\hat{p}^q_\sigma(y'_{1:2})+\hat{p}^q_\sigma(y'_{1:3})+\hat{p}^q_\sigma(y'_{1:2:3})$ \\ & = $0.30$ \\
      $\hat{p}^q_\sigma(y_{2})$ & = $\hat{p}^q_\sigma(y'_{2})+\hat{p}^q_\sigma(y'_{1:2})+\hat{p}^q_\sigma(y'_{2:3})+\hat{p}^q_\sigma(y'_{1:2:3})$ \\ &= $0.57$ \\
      $\hat{p}^q_\sigma(y_3)$ & = $\hat{p}^q_\sigma(y'_{3})+\hat{p}^q_\sigma(y'_{1:3})+\hat{p}^q_\sigma(y'_{2:3})+\hat{p}^q_\sigma(y'_{1:2:3})$ \\ &= $0.43$
    \end{tabular}
%
  \end{subfigure}%
\end{figure}

In principle, the disadvantage of this method is that it cannot learn
the correlations between classes that belong to different clusters.
However, the method is based on the intuition that classes that are
indeed correlated tend to end up in the same cluster, and that the
inability to model correlations between classes that belong to
different clusters will be more than compensated by the reduction in
the number of combinations that one needs to take into account.

In the experiments of Section~\ref{sec:exp:mlq} we explore different
configurations of this approach, in which we combine different
clustering strategies.



\section{Experiments}
\label{sec:experiments}

\noindent In this section we turn to describing the experiments we
have carried out in order to evaluate the performance of the different
methods for \MLQ\ that we have presented in the previous sections.  In
Section~\ref{sec:exp:eval} we discuss the evaluation measure we adopt,
while in Section~\ref{sec:exp:datasets} we describe the datasets on
which we perform our experiments. In Section~\ref{sec:exp:main} we
report experiments aiming at comparing the four groups of methods
discussed in Section~\ref{sec:quant:ml} and illustrated in
Figure~\ref{fig:schema}. In
Section~\ref{sec:exp:mlc}~and~\ref{sec:exp:mlq} we then move on to
exploring further instances of methods belonging to those four groups.


\subsection{Evaluation Measures}
\label{sec:exp:eval}

\noindent
Any evaluation measure for binary quantification can be easily turned
into an evaluation measure for multi-label quantification, since
evaluating a multi-label quantifier can be done by evaluating how well
the prevalence value $p(y_{i})$ of each class $y_{i}\in|\mathcal{Y}|$
is approximated by the prediction $\hat{p}(y_{i})$. As a result, it is
natural to take a standard measure $d(\mathbf{p}, \hat{\mathbf{p}})$
for the evaluation of binary quantification, and turn it into a
measure
\begin{align}
  \label{eq:d}
  \operatorname{D}(\mathbf{p}, \hat{\mathbf{p}})=\frac{1}{n}\sum_{i=1}^{n}d((p_{i},(1-p_{i})), (\hat{p}_{i},(1-\hat{p}_{i})))
\end{align}
\noindent for the evaluation of multi-label quantification. (This is
exactly what we do in multi-label \emph{classification}, in which we
take $F_{1}$, a standard measure for the evaluation of binary
classification, and turn it into macroaveraged $F_{1}$, which is the
standard measure for the evaluation of multi-label classification.)

The study of evaluation measures for binary (and single-label
multiclass) quantification performed in~\cite{Sebastiani:2020qf}
concludes that the most satisfactory such measures are \emph{absolute
error} and \emph{relative absolute error}; these are the two measures
that we are going to use in this paper.
In the binary case, absolute error is defined as
\begin{align}
  \begin{split}
    \label{eq:ae}
    \operatorname{ae}(\mathbf{p},\hat{\mathbf{p}}) & =  \frac{|p_{1}-\hat{p}_{1}| + |p_{2}-\hat{p}_{2}|}{2} \\
    & =  \frac{|p_{1}-\hat{p}_{1}| + |(1-p_{1})-(1-\hat{p}_{1})|}{2} \\
    & = |p_{1}-\hat{p}_{1}|
  \end{split}
\end{align}
\noindent which yields the multi-label version
\begin{equation}
  \label{eq:AE}
  \operatorname{AE}(\mathbf{p}, \hat{\mathbf{p}})=\frac{1}{n}\sum_{i=1}^{n}|p_{i}-\hat{p}_{i}|
\end{equation}
\noindent In the binary case, relative absolute error is instead
defined as
\begin{align}
  \begin{split}
    \label{eq:rae}
    \operatorname{rae}(\mathbf{p},\hat{\mathbf{p}})  & =  \frac{1}{2}\left(\frac{|p_{1}-\hat{p}_{1}|}{p_{1}} + \frac{|p_{2}-\hat{p}_{2}|}{p_{2}}\right)\\
    & = \frac{1}{2}\left(\frac{|p_{1}-\hat{p}_{1}|}{p_{1}} +
      \frac{|(1-p_{1})-(1-\hat{p}_{1})|}{(1-p_{1})}\right)
  \end{split}
\end{align}
\noindent which yields the multi-label version
\begin{equation}
  \label{eq:RAE}
  \operatorname{RAE}(\mathbf{p}, \hat{\mathbf{p}})=\frac{1}{2n}\sum_{i=1}^{n}\left(\frac{|p_{i}-\hat{p}_{i}|}{p_{i}} + \frac{|(1-p_{i})-(1-\hat{p}_{i})|}{(1-p_{i})}\right)
\end{equation}
\noindent Since \rabse\ is undefined when $p_{i}=0$ or $p_{i}=1$, we
smooth the probability distributions $\mathbf{p}$ and
$\hat{\mathbf{p}}$ via additive smoothing; in the binary case, this
maps a distribution $\mathbf{p}=(p_{i},(1-p_{i}))$ into
\begin{equation}
  \label{eq:smooth}
  \operatorname{s}(\mathbf{p}) =
  \left(\frac{\epsilon+p_{i}}{2\epsilon+1},\frac{\epsilon+(1-p_{i})}{2\epsilon+1}\right)
\end{equation}
\noindent with $\epsilon$ the smoothing factor, that we set, following
\cite{Forman:2008kx}, to $\epsilon=(2|\sigma|)^{-1}$.

Note that we do \emph{not} use, as a measure, \emph{concordance
ratio}, i.e.,
\begin{equation}
  \label{eq:cr}
  \operatorname{CR}(\mathbf{p}, \hat{\mathbf{p}})=\frac{1}{n}\sum_{y=1}^{n}\frac{\min\{p_{i}, \hat{p}_{i}\}}{\max \{ p_{i}, \hat{p}_{i}\}}
\end{equation}
\noindent despite the fact that it is the measure used in
\cite{Levin:2017dq}, the only paper in the literature that addresses
multi-label quantification. The reason why we do not use it is the
fact that, as later shown in~\cite{Sebastiani:2020qf}, the
mathematical properties of CR do not make it (similarly to other
measures used in the quantification literature in the past, such as
the Kullback-Leibler Divergence) a satisfactory measure for
quantification; see~\cite[pp.\ 272--273]{Sebastiani:2020qf} for
details.

In the experiments we describe in Section~\ref{sec:experiments}, the
trends we observe and the conclusions we draw for \abse\ hold for
\rabse\ as well. In Section~\ref{sec:experiments} we will thus report
our results in terms of \abse\ only, deferring the results in terms of
\rabse\ to Appendix \ref{sec:app:rae}.

\subsection{Datasets}
\label{sec:exp:datasets}

\noindent For our experiments we use 15 popular \MLC\ datasets,
including 3 datasets specific to text classification
(Reuters-21578,\footnote{\url{http://www.daviddlewis.com/resources/testcollections/reuters21578/}}
Ohsumed~\cite{Hersh:1994qm}, and
RCV1-v2\footnote{\url{http://www.ai.mit.edu/projects/jmlr/papers/volume5/lewis04a/lyrl2004_rcv1v2_{R}EADME.htm}}),
plus all the datasets linked from the \textsc{scikit-multilearn}
package~\cite{Szymanski:2017ru} with the exception of the RCV1-v2
subsets (we omit them since we already include the much larger
collection from which they were extracted). We refer to the original
sources for detailed descriptions of these datasets.\footnote{See also
\url{http://mlkd.csd.auth.gr/multilabel.html\#Datasets} and
\url{http://mulan.sourceforge.net/datasets-mlc.html}}

For the three textual datasets, we pre-process the text by applying
lowercasing, stop word removal, and punctuation removal, as
implemented in
\textsc{scikit-learn},\footnote{\url{https://scikit-learn.org/stable/index.html}}
and by masking numbers with a special token. We retain all terms
appearing at least 5 times in the training set, and convert the
resulting set of words into (sparse) tfidf-weighted vectors using
\textsc{scikit-learn}'s default
vectorizer.\footnote{\url{https://scikit-learn.org/stable/modules/generated/sklearn.feature_extraction.text.TfidfVectorizer.html}}

\begin{figure}
  \centering
  \captionof{table}{Description of the datasets. Columns \#Classes, \#Train,
  and \#Test indicate the number of classes, training datapoints, and
  test datapoints, respectively. Label cardinality (Card) reports the
  mean number of labels per datapoint. Label density (Dens) is the
  result of dividing the label cardinality by the total number of
  labels. Label diversity (Div) is the number of unique labelsets that
  are present in the dataset. Normalised label diversity (NormDiv)
  reports the ratio between label diversity and the total number of
  labels. The proportion of unique label combinations (PUniq) is the
  total number of labelsets that are unique in the dataset, divided by
  the number of examples. PMax reports the ratio of datapoints with
  the most frequent labelset divided by the total number of
  datapoints.}
  \label{tab:datasets}
  \resizebox{\textwidth}{!}{%
  \begin{tabular}{lrrrrrrrrrr}
    \toprule
    Dataset &  \#Classes &  \#Train &  \#Test &  \#Features &   Card &  Dens &   Div &  NormDiv &  PUniq &  PMax \\
    \midrule
    \texttt{Emotions} &         6 &     391 &    202 &         72 &  1.868 & 0.311 &    27 &    4.500 &  0.010 & 0.207 \\
    \texttt{Scene} &         6 &    1211 &   1196 &        294 &  1.074 & 0.179 &    15 &    2.500 &  0.002 & 0.334 \\
    \texttt{Yeast} &        14 &    1500 &    917 &        103 &  4.237 & 0.303 &   198 &   14.143 &  0.051 & 0.158 \\
    \texttt{Birds} &        17 &     322 &    323 &        260 &  0.991 & 0.058 &   124 &    7.294 &  0.205 & 0.932 \\
    \texttt{Genbase} &        18 &     463 &    199 &       1186 &  1.219 & 0.068 &    23 &    1.278 &  0.006 & 0.369 \\
    \texttt{Medical} &        18 &     333 &    645 &       1449 &  1.135 & 0.063 &    50 &    2.778 &  0.033 & 0.495 \\
    \texttt{Tmc2007\_{5}00} &        22 &   21519 &   7077 &        500 &  2.220 & 0.101 &  1172 &   53.273 &  0.019 & 0.115 \\
    \texttt{Ohsumed} &        23 &   24061 &  10328 &      18238 &  1.657 & 0.072 &  1901 &   82.652 &  0.041 & 0.120 \\
    \texttt{Enron} &        45 &    1123 &    579 &       1001 &  3.357 & 0.075 &   734 &   16.311 &  0.491 & 0.147 \\
    \texttt{Reuters21578} &        72 &    9603 &   3299 &       8250 &  1.029 & 0.014 &   447 &    6.208 &  0.028 & 0.409 \\
    \texttt{RCV1-v2} &        98 &   23149 & 781265 &      24816 &  3.199 & 0.033 & 14820 &  151.224 &  0.345 & 2.323 \\
    \texttt{Mediamill} &       100 &   30993 &  12914 &        120 &  4.374 & 0.044 &  6548 &   65.480 &  0.132 & 0.076 \\
    \texttt{Bibtex} &       159 &    4880 &   2515 &       1836 &  2.402 & 0.015 &  2856 &   17.962 &  0.451 & 0.097 \\
    \texttt{Corel5k} &       292 &    4500 &    500 &        499 &  3.480 & 0.012 &  3113 &   10.661 &  0.543 & 0.012 \\
    \texttt{Delicious} &       983 &   12920 &   3185 &        500 & 19.020 & 0.019 & 15806 &   16.079 &  1.211 & 0.001 \\
    \bottomrule
  \end{tabular}  
  }

\vspace{5ex}

  \centering
  \includegraphics[width=\textwidth]{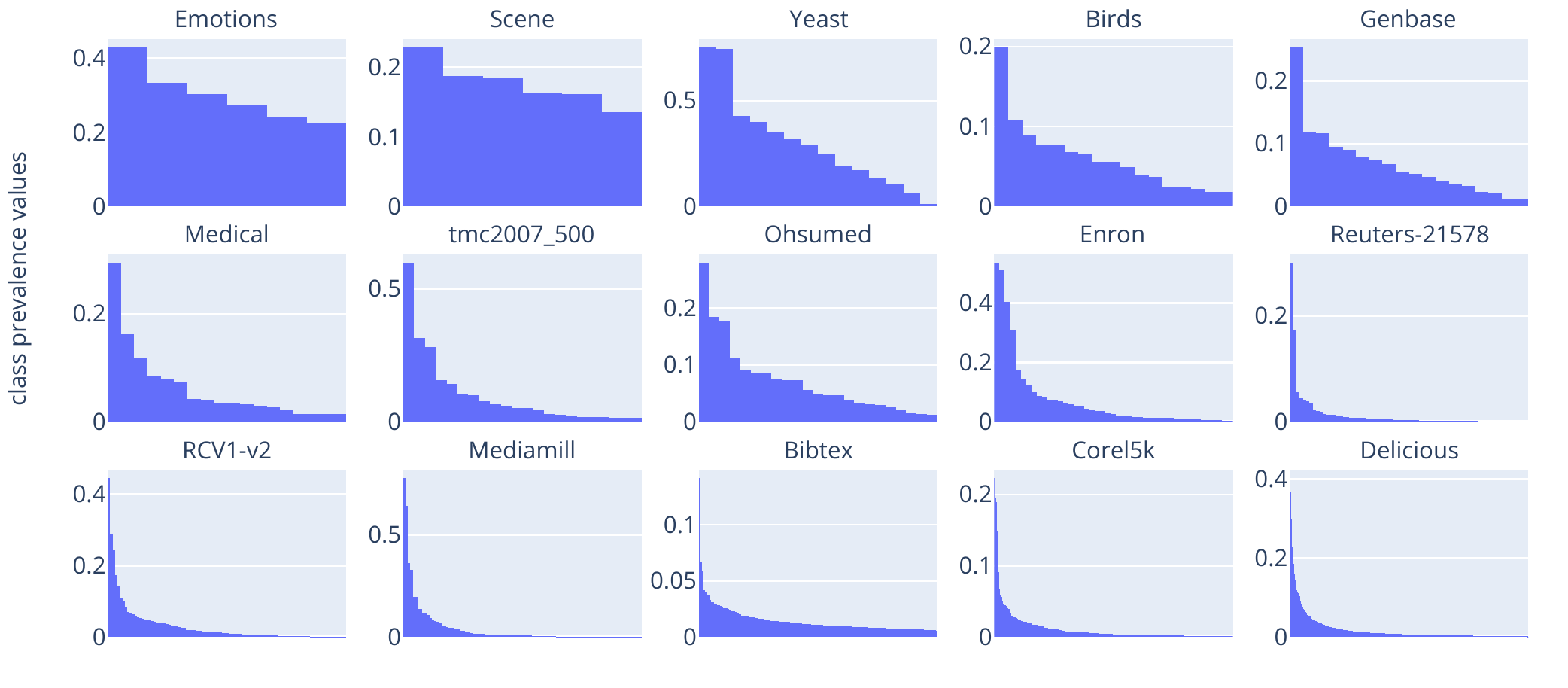}
  \captionof{figure}{Histograms of class prevalence values, one per dataset, sorted from highly populated datasets to lowly populated ones; values on the $x$ axis indicate intervals $[\alpha_{k},\beta_{k}]$ of class prevalence values, while values on the $y$ axis indicate the fraction of classes in the dataset that have prevalence values in the $[\alpha_{k},\beta_{k}]$ interval.
  }
  \label{fig:prevhist}
\vspace{-3ex}
\end{figure}

For all datasets, we remove very rare classes (i.e., those with fewer
than 5 training examples) from consideration, since they pose a
problem when it comes to generating validation (i.e., held-out data)
sets. Indeed, since we optimize the hyperparameters for all the
methods we use (as explained below), we need validation sets, and it
is sometimes impossible to have positive examples for these classes in
both the training and validation sets (let us remember that pure
stratification in multi-label datasets is not always achievable, as
argued in~\cite{Sechidis:2011tu, Szymanski:2017yc}).
Note that all this only concerns the training set,
and has nothing to do with the test set, which can include (and indeed
includes, for most datasets) extremely rare classes, since removing
classes that are rare in the test set would lead to an unrealistic
experimentation.  Note also that removing classes that are rare in the
training set is ``fair'', i.e., equally affects all methods that we
experimentally compare, since all of them involve hyperparameter
optimization.
Finally, note that, whenever a method requires generating
\emph{additional} (and maybe nested) validation sets, it is inevitably
exposed to the problems mentioned above, and can thus be at a
disadvantage with respect to other methods that do not require
additional validation data.
Table~\ref{tab:datasets} shows a complete description of the datasets
we use (after deleting rare classes), along with some useful
statistics proposed in~\cite{Read:2010rk, Zhang:2014zn}, while Figure~\ref{fig:prevhist} shows the distribution of prevalence values for each dataset. Note that, in most datasets, this distribution obeys a power law.

We set the parameters of the ML-APP for generating test samples (see
Section~\ref{sec:mlapp}) as follows. We fix the sample size to $k=100$
in all cases. We set the grid of prevalence values to
$\mathbf{g}=\{0.00,0.01,\ldots,0.99,1.00\}$ in all cases but for
dataset \verb+Delicious+, since in this latter the number of
combinations thus generated would be intractable, given that this is
dataset with no fewer than 983 classes; for \verb+Delicious+ we use
the coarser-grained grid
$\mathbf{g}=\{0.00,0.05,\ldots,0.95,1.00\}$. We set $m$ (the number of
samples to be drawn for each prevalence value) independently for each
dataset, to the smallest number that yields more than 10,000 test
samples ($m$ ranges from 1 in \verb+Delicious+ to 40 in \verb+Birds+).

We break down the results into three groups, each corresponding to a
different amount of shift. The rationale behind this choice is to
allow for a more meaningful analysis of the quantifiers' performance,
since the APP (and, by extension, the ML-APP) has often been the
subject of criticism for generating samples exhibiting degrees of
shift that are judged unrealistic and unlikely to occur in real
cases~\cite{Esuli:2015gh,Hassan:2021af}.
We instead believe that general-purpose quantification methods should
be tested in widely varying situations, from low-shift to high-shift
ones, and we thus prefer to test all such scenarios, but split the
corresponding results into groups characterized by of more or less
homogeneous amounts of shift.

More specifically, for each test sample generated via the ML-APP, we
compute its prior probability shift with respect to the training set
in terms of $\operatorname{AE}$ between the vectors of training and
test class prevalence values. We then bring together all the resulting
shift values and split the range of such values in three equally-sized
intervals (that we dub \emph{low shift}, \emph{mid shift}, and
\emph{high shift}). The accuracy values we report are thus not
averages across the same number of experiments, since the ML-APP often
tends to generate more samples in the low-shift region than samples in
the mid-shift region and (above all) in the high-shift region. The
number of samples, as well as the distribution of shift values,
depends on each dataset. 

\begin{figure}[h!]
  \centering \includegraphics[trim={0 0 0
  1cm},clip,width=\textwidth]{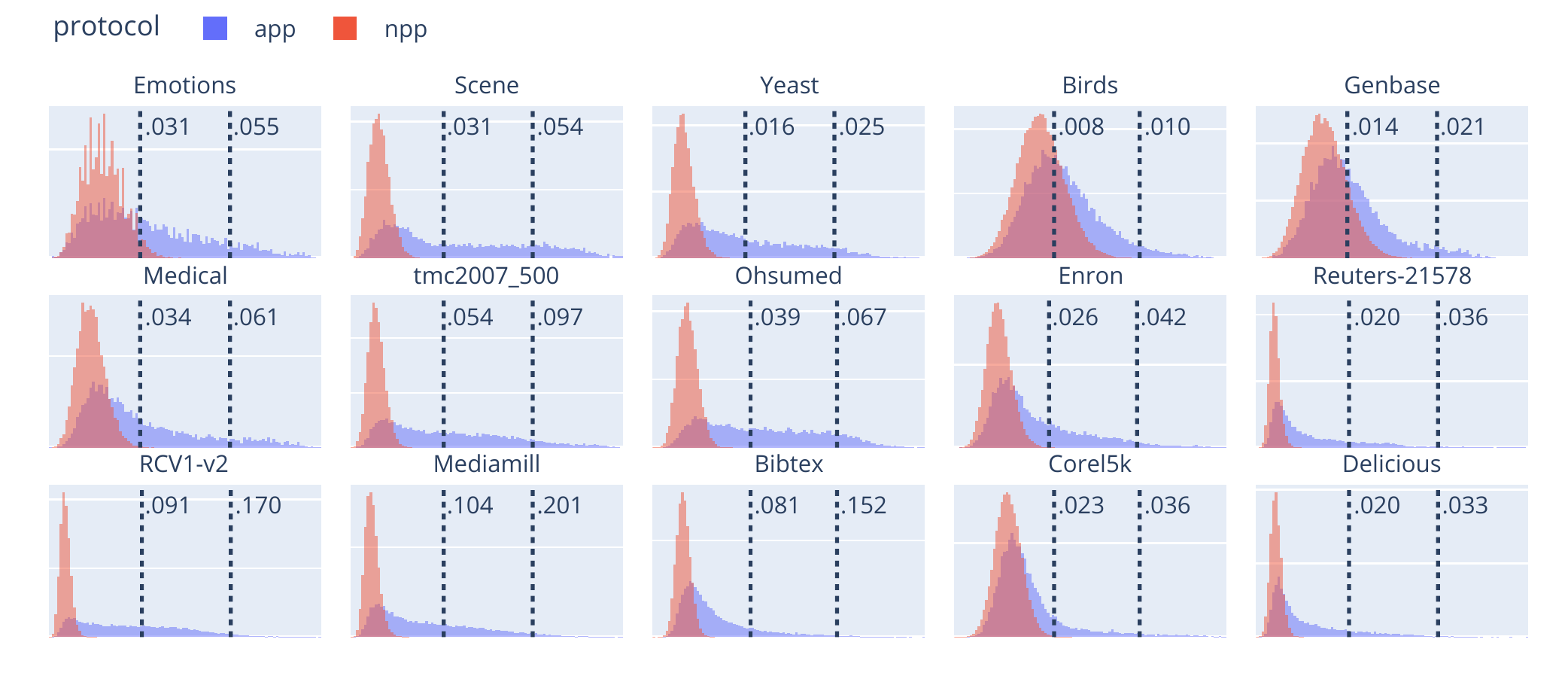}
  \caption{Shifts generated via the proposed
  ML-APP (blue) and via uniform sampling (blue), as computed in terms of $\operatorname{AE}$ between the training set and the test samples.}
  \label{fig:drift:appvsnpp}
\end{figure}

Figure~\ref{fig:drift:appvsnpp} shows the
distributions of shift values that the ML-APP generates (blue) along with the distributions of shift values that we would obtain via uniform sampling (red). Note that the ML-APP succeeds in generating larger amounts of shift, and that most of the samples generated via uniform sampling would fall within what we call the ``low shift'' region.

\subsection{Testing Instances of the Four Types of Multi-Label
Quantification Methods}
\label{sec:exp:main}

\noindent The goal of this section is to provide an answer to the
question: ``Which among the four groups of multi-label quantification
methods tends to perform best?''

To this aim, we choose one representative instance from each group,
and carry out the experiments using all the datasets. We perform this
choice by combining the following components:
\begin{itemize}

\item As the \textbf{binary classification method}, we choose logistic
  regression (LR), and use the implementation of it available from
  \textsc{scikit-learn}.\footnote{\url{https://scikit-learn.org/stable/modules/generated/sklearn.linear_model.LogisticRegression.html}}
  We consider LR a good choice, given that it is a probabilistic
  classifier that already provides fairly well calibrated posterior
  probabilities (which is of fundamental importance in PCC, PACC, and
  SLD), and given that, as indicated by previously reported
  results~\cite{Moreo:2021bs}, it tends to perform well. A set of LR
  classifiers are used when testing the binary relevance (BR) method
  described in Section~\ref{sec:relwork:algadapt}.
 
\item As the \textbf{multi-label classification method}, we adopt
  \emph{stacked generalization}~\cite{Wolpert:1992rq} (SG -- see
  Section~\ref{sec:relwork:ensembles}). We use our own implementation
  (since the implementation of stacked generalization available from
  \textsc{scikit-learn} only caters for the single-label
  case)\footnote{\url{https://scikit-learn.org/stable/modules/generated/sklearn.ensemble.StackingClassifier.html}},
  that relies on 5-fold cross-validation to generate the intermediate
  representations (in the form of posterior probabilities) given as
  input to the meta-learner, concatenated with the original input
  features. The base members of the ensemble consist of binary
  logistic regression classifiers as implemented in
  \textsc{scikit-learn}.
 
\item As the \textbf{binary aggregation method} Q, we experiment with
  all the methods covered in Section~\ref{sec:quant:sl}, i.e., CC,
  PCC, ACC, PACC, SLD. For all these methods we use the
  implementations made available in the \textsc{QuaPy} open-source
  library~\cite{Moreo:2021bs}.\footnote{\url{https://github.com/HLT-ISTI/QuaPy}}
 
\item As the \textbf{multi-label aggregation method}, we use the
  regressor-based strategy for quantification (that we dub RQ)
  described in Section~\ref{sec:quant:ml:reg}. We implement this
  method as part of the \textsc{QuaPy} framework. For training the base quantifier $q$ we experiment again 
  with all the methods covered in Section~\ref{sec:quant:sl}, 
  i.e., CC, PCC, ACC, PACC, SLD, while as the internal 
  regressor which receives its input from the base quantifier $q$ we use linear
  support vector regression (SVR), for which
  we use the \textsc{scikit-learn}
  implementation.\footnote{\url{https://scikit-learn.org/stable/modules/generated/sklearn.svm.LinearSVR.html}}
  As the held-out validation set $L_{R}$ needed for training the
  regressor we use a set consisting of 40\% of the training
  datapoints, 
  chosen via iterative stratification~\cite{Szymanski:2017yc,
  Sechidis:2011tu} as implemented in
  \textsc{scikit-multilearn}.\footnote{\url{http://scikit.ml/stratification.html}}
  We call this aggregation method SVR-RQ. 
  
\end{itemize}

\noindent The methods we use in this experiment thus amount to the
combinations illustrated in
Table~\ref{tab:representative}. 
\begin{table}[h!]
  \centering
  \caption{Methods we use as instances of the four types of methods
  illustrated in Figure~\ref{fig:schema}.}
  \label{tab:representative}
  \begin{tabular}{ccc}
    \toprule
    \textbf{Type} & \textbf{Classification} & \textbf{Aggregation} \\ \midrule
    \bqbc & LR & Q$\in$\{CC,PCC,ACC,PACC,SLD\} \\
    \bqmc & SG & Q$\in$\{CC,PCC,ACC,PACC,SLD\} \\
    \mqbc & LR & Q$\in$\{CC,PCC,ACC,PACC,SLD\} + \textsc{SVR-RQ} \\
    \mqmc & SG & Q$\in$\{CC,PCC,ACC,PACC,SLD\} + \textsc{SVR-RQ} \\\bottomrule
  \end{tabular}
\end{table}

Following~\cite{Moreo:2021sp}, we perform model selection by using, as
the loss function to minimize, a quantification-oriented error measure
(and not a classification-oriented one), and by adopting the same
protocol used for the evaluation of our quantifiers. That is, model
selection is carried out by first splitting the training set $L$ into
two disjoint sets, i.e., (a) a proper training set
$L_{\operatorname{tr}}$ and (b) a held-out validation set
$L_{\operatorname{va}}$ consisting of 40\% of the labelled
datapoints. For splitting the training set, we again rely on the
iterative stratification routine of \textsc{scikit-multilearn} (see
Footnote~\protect\ref{foot:iterativestrat}). We use
$L_{\operatorname{tr}}$ to train the quantifiers with different
combinations of hyperparameters, while from $L_{\operatorname{va}}$ we
extract, via the ML-APP, validation samples on which we assess, via
$\operatorname{AE}$ (the same measure we use in the evaluation phase),
the quality of the hyperparameter combinations.  We explore the
hyperparameters via grid-search optimization, and use the best
configuration to retrain the quantifier on the entire training set $L$
after model selection. During the model selection phase, for the
ML-APP we use the same parameters $k$ and $\mathbf{g}$ that we use in
the test phase, but we reduce the number of repetitions $m$ to 5 in
the datasets with fewer than 90 classes, and to 1 in the other
datasets, in order to keep the computational burden under reasonable
bounds. 

The hyperparameters we explore for LR include $C$, the inverse of the
regularization strength, in the range
$\{10^{-1},10^{0},10^{1},10^{2},10^{3}\}$, and \emph{ClassWeight},
which takes values in Balanced (which reweights the importance
of the examples so as to equate the overall contribution of each
class) or None (which gives the same weight to all datapoints,
irrespectively of the prevalence of the class they belong to). In
cases in which the class-specific classifiers are independent of each
other (i.e., for methods belonging to the \bqbc\ and \mqbc\ types) we
optimize the hyperparameters independently for each class. For SG, we
only optimize the hyperparameters of the meta-classifier, leaving the
hyperparameters of the base members to their default values. In
particular, we explore the parameters $C$ and \emph{ClassWeight}
as before, plus the hyperparameter \emph{Normalize}, which takes
values in \emph{True} (which has the effect of standardizing the
inputs of the meta-classifier so that every dimension has zero mean
and unit variance) and \emph{False} (which does not standardize the
inputs). For RQ we only explore the regularization hyperparameter $C$
in the range $\{10^{-1},10^{0},10^{1},10^{2},10^{3}\}$.
Note that the base quantifiers (i.e., CC, PCC, ACC, PACC, SLD) have no
specific internal hyperparameters to be tuned.

The results we have obtained for the different choices of the base
quantifier are reported in Table~\ref{tab:cc_general} for CC,
Table~\ref{tab:pcc_general} for PCC, Table~\ref{tab:acc_general} for
ACC, Table~\ref{tab:pacc_general} for PACC, and
Table~\ref{tab:sld_general} for SLD. The results clearly show (see
especially the last two rows of each table) that there is an ordering
\bqbc\ $\prec$ \bqmc\ $\prec$ \mqbc\ $\prec$ \mqmc, in which $\prec$
means ``performs worse than'', which holds, independently of the base
quantifier of choice, in almost all cases. The same experiments also
indicate that \emph{there is a substantial improvement in performance that
derives from simply replacing the binary classifiers with one
multi-label classifier} (moving from \bqbc\ to \bqmc\ or from \mqbc\ to
\mqmc), i.e., from bringing to bear the class-class correlations at
the classification
stage, 
and that \emph{there is an equally substantial improvement}
\emph{when binary aggregation is replaced by multi-label aggregation}
(switching from \bqbc\ to \mqbc\ or from \bqmc\ to \mqmc), i.e., when
the class-class correlations are exploited at the aggregation
stage. What also emerges from these results is that, consistently with
the above observations, \emph{the best-performing group of methods is \mqmc},
i.e., methods that explicitly take class dependencies into account
\emph{both} at the classification stage and at the aggregation stage.

Note that methods that learn from the stochastic correlations among
the classes perform much better than methods that do not, even in the
low shift regime. Overall, the best-performing method on average is
\mqmc\ when equipped with PCC as the base quantifier.

The reader might wonder why we do not use as a baseline the system
presented in the only paper in the literature that tackles multi-label
quantification, i.e.,~\cite{Levin:2017dq}. There are several reasons
for this: (a) the authors do not make the code available; (b) the
method is, as already discussed in Section~\ref{sec:multi-label},
computationally expensive, and as a result the authors test it on a
single dataset whose codeframe consists of 16 classes only; using this
method on our 15 datasets, whose codeframes count up to 983 classes,
and 125 classes on average, would be prohibitive; (c) the method is
essentially a calibration strategy for binary classification, which
means that it falls in the group of ``naive'' \bqbc\ methods since it
does not tackle at all, as already mentioned in
Section~\ref{sec:multi-label}, the multi-label nature of the MLQ
problem.


\begin{table}[tb]
  \centering
  \caption{Values of \abse\ obtained in our experiments for different
  amounts of shift using CC as the base quantifier. The number of test
  samples generated for each dataset exceeds 10,000, though there is a
  variable number of samples allocated in each region of
  shift. \textbf{Boldface} indicates the best method for a given
  dataset and shift region. Superscripts $\dag$ and $\ddag$ denote the
  methods (if any) whose scores are not statistically significantly
  different from the best one according to a Wilcoxon signed-rank test
  at different confidence levels: symbol $\dag$ indicates 0.001 $<$
  $p$-value $<$ 0.05 while symbol $\ddag$ indicates 0.05 $\le$
  $p$-value. For ease of readability, for each pair \{dataset, shift\}
  we colour-code cells via intense green for the best result, intense
  red for the worst result, and an interpolated tone for the scores
  in-between.}
  \label{tab:cc_general}
  
                \resizebox{\textwidth}{!}{%
                        \begin{tabular}{|l|cccc|cccc|cccc|} \cline{2-13}
\multicolumn{1}{c|}{} & \multicolumn{4}{c|}{low shift} & \multicolumn{4}{c|}{mid shift} & \multicolumn{4}{c|}{high shift}\\\cline{2-13}

\multicolumn{1}{c|}{} & \side{\bqbc} & \side{\bqmc} & \side{\mqbc} & \side{\mqmc\phantom{0}} & \side{\bqbc} & \side{\bqmc} & \side{\mqbc} & \side{\mqmc\phantom{0}} & \side{\bqbc} & \side{\bqmc} & \side{\mqbc} & \side{\mqmc\phantom{0}}\\\hline
\texttt{Emotions} & .0749 \cellcolor{red!35} & .0626 \cellcolor{red!3} & \textbf{.0478} \cellcolor{green!35} & .0644$^{\phantom{\ddag}}$ \cellcolor{red!8} & .0838 \cellcolor{red!32} & .0776 \cellcolor{red!15} & \textbf{.0598}$^{\phantom{\ddag}}$ \cellcolor{green!35} & .0848 \cellcolor{red!35} & .0967 \cellcolor{red!23} & .0899 \cellcolor{red!11} & \textbf{.0616}$^{\phantom{\ddag}}$ \cellcolor{green!35} & .1039$^{\phantom{\ddag}}$ \cellcolor{red!35} \\
\texttt{Scene} & .0754 \cellcolor{red!35} & .0558 \cellcolor{red!1} & .0458 \cellcolor{green!16} & \textbf{.0349}$^{\phantom{\ddag}}$ \cellcolor{green!35} & .0908 \cellcolor{red!35} & .0759 \cellcolor{red!8} & .0553$^{\phantom{\ddag}}$ \cellcolor{green!27} & \textbf{.0508} \cellcolor{green!35} & .1110 \cellcolor{red!35} & .0966 \cellcolor{red!14} & \textbf{.0609}$^{\phantom{\ddag}}$ \cellcolor{green!35} & .0668$^{\phantom{\ddag}}$ \cellcolor{green!26} \\
\texttt{Yeast} & .1481 \cellcolor{red!35} & .1119 \cellcolor{red!8} & \textbf{.0511} \cellcolor{green!35} & .0516$^{\dag\phantom{\dag}}$ \cellcolor{green!34} & .1644 \cellcolor{red!35} & .1397 \cellcolor{red!12} & \textbf{.0879}$^{\phantom{\ddag}}$ \cellcolor{green!35} & .0958 \cellcolor{green!27} & .1919 \cellcolor{red!35} & .1754 \cellcolor{red!17} & \textbf{.1272}$^{\phantom{\ddag}}$ \cellcolor{green!35} & .1403$^{\phantom{\ddag}}$ \cellcolor{green!20} \\
\texttt{Birds} & .0229 \cellcolor{red!15} & .0243 \cellcolor{red!35} & \textbf{.0191} \cellcolor{green!35} & .0202$^{\phantom{\ddag}}$ \cellcolor{green!19} & .0258 \cellcolor{green!25} & .0276 \cellcolor{red!35} & \textbf{.0255}$^{\phantom{\ddag}}$ \cellcolor{green!35} & .0264 \cellcolor{green!6} & \textbf{.0293} \cellcolor{green!35} & .0331 \cellcolor{green!1} & .0371$^{\phantom{\ddag}}$ \cellcolor{red!35} & .0357$^{\phantom{\ddag}}$ \cellcolor{red!22} \\
\texttt{Genbase} & \textbf{.0003} \cellcolor{green!35} & .0006 \cellcolor{green!19} & .0016 \cellcolor{red!35} & .0007$^{\phantom{\ddag}}$ \cellcolor{green!13} & \textbf{.0003} \cellcolor{green!35} & .0006 \cellcolor{green!22} & .0018$^{\phantom{\ddag}}$ \cellcolor{red!35} & .0007 \cellcolor{green!18} & \textbf{.0003} \cellcolor{green!35} & .0005 \cellcolor{green!22} & .0017$^{\phantom{\ddag}}$ \cellcolor{red!35} & .0006$^{\phantom{\ddag}}$ \cellcolor{green!17} \\
\texttt{Medical} & .0182 \cellcolor{red!35} & \textbf{.0121} \cellcolor{green!35} & .0175 \cellcolor{red!26} & .0121$^{\phantom{\ddag}}$ \cellcolor{green!34} & .0183 \cellcolor{red!13} & \textbf{.0130} \cellcolor{green!35} & .0206$^{\phantom{\ddag}}$ \cellcolor{red!35} & .0130 \cellcolor{green!34} & .0174 \cellcolor{green!9} & \textbf{.0141} \cellcolor{green!35} & .0230$^{\phantom{\ddag}}$ \cellcolor{red!35} & .0141$^{\phantom{\ddag}}$ \cellcolor{green!34} \\
\texttt{tmc2007\_500} & .0700 \cellcolor{red!35} & .0333 \cellcolor{green!17} & .0222 \cellcolor{green!33} & \textbf{.0214}$^{\phantom{\ddag}}$ \cellcolor{green!35} & .0758 \cellcolor{red!35} & .0464 \cellcolor{green!7} & .0313$^{\phantom{\ddag}}$ \cellcolor{green!30} & \textbf{.0278} \cellcolor{green!35} & .0651 \cellcolor{red!35} & .0468 \cellcolor{green!3} & .0333$^{\phantom{\ddag}}$ \cellcolor{green!32} & \textbf{.0320}$^{\phantom{\ddag}}$ \cellcolor{green!35} \\
\texttt{Ohsumed} & .0338 \cellcolor{red!35} & .0184 \cellcolor{green!31} & .0198 \cellcolor{green!25} & \textbf{.0176}$^{\phantom{\ddag}}$ \cellcolor{green!35} & .0395 \cellcolor{red!35} & .0252 \cellcolor{green!28} & .0246$^{\phantom{\ddag}}$ \cellcolor{green!30} & \textbf{.0236} \cellcolor{green!35} & .0457 \cellcolor{red!35} & .0303 \cellcolor{green!22} & \textbf{.0270}$^{\phantom{\ddag}}$ \cellcolor{green!35} & .0272$^{\ddag}$ \cellcolor{green!34} \\
\texttt{Enron} & .0239 \cellcolor{red!35} & .0207 \cellcolor{red!1} & \textbf{.0172} \cellcolor{green!35} & .0198$^{\phantom{\ddag}}$ \cellcolor{green!7} & .0274 \cellcolor{red!35} & .0247 \cellcolor{green!6} & \textbf{.0228}$^{\phantom{\ddag}}$ \cellcolor{green!35} & .0245 \cellcolor{green!9} & .0280 \cellcolor{red!35} & \textbf{.0253} \cellcolor{green!35} & .0254$^{\ddag}$ \cellcolor{green!31} & .0265$^{\phantom{\ddag}}$ \cellcolor{green!2} \\
\texttt{Reuters-21578} & .0067 \cellcolor{red!35} & \textbf{.0035} \cellcolor{green!35} & .0055 \cellcolor{red!7} & .0036$^{\phantom{\ddag}}$ \cellcolor{green!31} & .0120 \cellcolor{red!35} & \textbf{.0056} \cellcolor{green!35} & .0081$^{\phantom{\ddag}}$ \cellcolor{green!7} & .0058 \cellcolor{green!32} & .0291 \cellcolor{red!35} & \textbf{.0071} \cellcolor{green!35} & .0113$^{\phantom{\ddag}}$ \cellcolor{green!21} & .0075$^{\phantom{\ddag}}$ \cellcolor{green!33} \\
\texttt{RCV1-v2} & .0198 \cellcolor{red!35} & \textbf{.0081} \cellcolor{green!35} & .0101 \cellcolor{green!22} & .0083$^{\phantom{\ddag}}$ \cellcolor{green!33} & .0251 \cellcolor{red!35} & \textbf{.0118} \cellcolor{green!35} & .0162$^{\phantom{\ddag}}$ \cellcolor{green!11} & .0122 \cellcolor{green!32} & .0360 \cellcolor{red!35} & \textbf{.0176} \cellcolor{green!35} & .0240$^{\phantom{\ddag}}$ \cellcolor{green!10} & .0192$^{\phantom{\ddag}}$ \cellcolor{green!28} \\
\texttt{Mediamill} & .1390 \cellcolor{red!35} & .0236 \cellcolor{green!30} & .0159 \cellcolor{green!34} & \textbf{.0155}$^{\phantom{\ddag}}$ \cellcolor{green!35} & .1488 \cellcolor{red!35} & .0341 \cellcolor{green!29} & .0252$^{\phantom{\ddag}}$ \cellcolor{green!34} & \textbf{.0247} \cellcolor{green!35} & .1690 \cellcolor{red!35} & .0476 \cellcolor{green!26} & .0322$^{\phantom{\ddag}}$ \cellcolor{green!34} & \textbf{.0310}$^{\phantom{\ddag}}$ \cellcolor{green!35} \\
\texttt{Bibtex} & .0137 \cellcolor{red!35} & .0097 \cellcolor{green!29} & \textbf{.0093} \cellcolor{green!35} & .0096$^{\phantom{\ddag}}$ \cellcolor{green!30} & .0150 \cellcolor{red!35} & \textbf{.0111} \cellcolor{green!35} & .0117$^{\phantom{\ddag}}$ \cellcolor{green!23} & .0113 \cellcolor{green!32} & .0177 \cellcolor{red!35} & \textbf{.0120} \cellcolor{green!35} & .0131$^{\phantom{\ddag}}$ \cellcolor{green!22} & .0124$^{\phantom{\ddag}}$ \cellcolor{green!29} \\
\texttt{Corel5k} & .0357 \cellcolor{red!35} & .0093 \cellcolor{green!32} & \textbf{.0082} \cellcolor{green!35} & .0082$^{\phantom{\ddag}}$ \cellcolor{green!34} & .0363 \cellcolor{red!35} & .0097 \cellcolor{green!32} & \textbf{.0087}$^{\phantom{\ddag}}$ \cellcolor{green!35} & .0092 \cellcolor{green!33} & .0371 \cellcolor{red!35} & .0100 \cellcolor{green!33} & .0099$^{\phantom{\ddag}}$ \cellcolor{green!33} & \textbf{.0095}$^{\phantom{\ddag}}$ \cellcolor{green!35} \\
\texttt{Delicious} & .1037 \cellcolor{red!35} & .0116 \cellcolor{green!33} & .0096 \cellcolor{green!34} & \textbf{.0094}$^{\phantom{\ddag}}$ \cellcolor{green!35} & .1036 \cellcolor{red!35} & .0134 \cellcolor{green!33} & .0113$^{\ddag}$ \cellcolor{green!34} & \textbf{.0112} \cellcolor{green!35} & .0904 \cellcolor{red!35} & .0128 \cellcolor{green!33} & \textbf{.0109}$^{\phantom{\ddag}}$ \cellcolor{green!35} & .0110$^{\ddag}$ \cellcolor{green!34} \\
\hline
\multicolumn{1}{|c|}{Average} & .0492 \cellcolor{red!35} & .0237 \cellcolor{green!21} & .0180 \cellcolor{green!34} & \textbf{.0177}$^{\phantom{\ddag}}$ \cellcolor{green!35} & .0555 \cellcolor{red!35} & .0358 \cellcolor{green!16} & \textbf{.0288}$^{\phantom{\ddag}}$ \cellcolor{green!35} & .0295 \cellcolor{green!33} & .0757 \cellcolor{red!35} & .0576 \cellcolor{green!4} & \textbf{.0433}$^{\phantom{\ddag}}$ \cellcolor{green!35} & .0483$^{\phantom{\ddag}}$ \cellcolor{green!24} \\
\multicolumn{1}{|c|}{Rank Average} &  3.7 \cellcolor{red!35} &  2.5 \cellcolor{green!10} &  2.0 \cellcolor{green!27} & \textbf{1.8} \cellcolor{green!35} &  3.5 \cellcolor{red!35} &  2.4 \cellcolor{green!14} &  2.1 \cellcolor{green!26} & \textbf{1.9} \cellcolor{green!35} &  3.5 \cellcolor{red!35} & \textbf{2.1} \cellcolor{green!35} &  2.3 \cellcolor{green!28} &  2.1 \cellcolor{green!35} \\

                \hline
                \end{tabular}%
                }
            
\end{table}

\begin{table}[tb]
  \centering
  \caption{Values of \abse\ obtained in our experiments for different
  amounts of shift using PCC as the base quantifier. Notational
  conventions are as in Table~\ref{tab:cc_general}.}
  \label{tab:pcc_general}
  
                \resizebox{\textwidth}{!}{%
                        \begin{tabular}{|l|cccc|cccc|cccc|} \cline{2-13}
\multicolumn{1}{c|}{} & \multicolumn{4}{c|}{low shift} & \multicolumn{4}{c|}{mid shift} & \multicolumn{4}{c|}{high shift}\\\cline{2-13}

\multicolumn{1}{c|}{} & \side{\bqbc} & \side{\bqmc} & \side{\mqbc} & \side{\mqmc\phantom{0}} & \side{\bqbc} & \side{\bqmc} & \side{\mqbc} & \side{\mqmc\phantom{0}} & \side{\bqbc} & \side{\bqmc} & \side{\mqbc} & \side{\mqmc\phantom{0}}\\\hline
\texttt{Emotions} & \textbf{.0418} \cellcolor{green!35} & .0420 \cellcolor{green!34} & .0586 \cellcolor{red!35} & .0474 \cellcolor{green!11} & .0685 \cellcolor{red!19} & .0654$^{\phantom{\ddag}}$ \cellcolor{red!5} & .0720 \cellcolor{red!35} & \textbf{.0564}$^{\phantom{\ddag}}$ \cellcolor{green!35} & .0923 \cellcolor{red!28} & .0857 \cellcolor{red!12} & .0951 \cellcolor{red!35} & \textbf{.0664} \cellcolor{green!35} \\
\texttt{Scene} & .0842 \cellcolor{red!35} & .0362 \cellcolor{green!29} & .0384 \cellcolor{green!26} & \textbf{.0323} \cellcolor{green!35} & .1140 \cellcolor{red!35} & .0827$^{\phantom{\ddag}}$ \cellcolor{red!3} & .0459 \cellcolor{green!33} & \textbf{.0445}$^{\phantom{\ddag}}$ \cellcolor{green!35} & .1427 \cellcolor{red!35} & .1221 \cellcolor{red!19} & \textbf{.0499} \cellcolor{green!35} & .0589 \cellcolor{green!28} \\
\texttt{Yeast} & .1756 \cellcolor{red!35} & \textbf{.0444} \cellcolor{green!35} & .0480 \cellcolor{green!33} & .0472 \cellcolor{green!33} & .1900 \cellcolor{red!35} & .0992$^{\phantom{\ddag}}$ \cellcolor{green!25} & .0913 \cellcolor{green!30} & \textbf{.0845}$^{\phantom{\ddag}}$ \cellcolor{green!35} & .2206 \cellcolor{red!35} & .1549 \cellcolor{green!11} & .1361 \cellcolor{green!25} & \textbf{.1226} \cellcolor{green!35} \\
\texttt{Birds} & .0286 \cellcolor{red!35} & .0208 \cellcolor{green!17} & \textbf{.0181} \cellcolor{green!35} & .0213 \cellcolor{green!13} & .0328 \cellcolor{red!35} & .0246$^{\phantom{\ddag}}$ \cellcolor{green!32} & \textbf{.0243} \cellcolor{green!35} & .0288$^{\phantom{\ddag}}$ \cellcolor{red!2} & .0440 \cellcolor{red!35} & \textbf{.0321} \cellcolor{green!35} & .0351 \cellcolor{green!17} & .0404 \cellcolor{red!14} \\
\texttt{Genbase} & .0011 \cellcolor{green!12} & \textbf{.0005} \cellcolor{green!35} & .0022 \cellcolor{red!35} & .0007 \cellcolor{green!29} & .0011 \cellcolor{green!15} & \textbf{.0005}$^{\phantom{\ddag}}$ \cellcolor{green!35} & .0025 \cellcolor{red!35} & .0007$^{\phantom{\ddag}}$ \cellcolor{green!30} & .0010 \cellcolor{green!13} & \textbf{.0005} \cellcolor{green!35} & .0023 \cellcolor{red!35} & .0006 \cellcolor{green!30} \\
\texttt{Medical} & .0127 \cellcolor{green!28} & .0138 \cellcolor{green!17} & .0191 \cellcolor{red!35} & \textbf{.0120} \cellcolor{green!35} & .0146 \cellcolor{green!30} & .0156$^{\phantom{\ddag}}$ \cellcolor{green!25} & .0279 \cellcolor{red!35} & \textbf{.0136}$^{\phantom{\ddag}}$ \cellcolor{green!35} & .0169 \cellcolor{green!31} & .0183 \cellcolor{green!26} & .0351 \cellcolor{red!35} & \textbf{.0160} \cellcolor{green!35} \\
\texttt{tmc2007\_500} & .1108 \cellcolor{red!35} & .0193 \cellcolor{green!34} & .0213 \cellcolor{green!32} & \textbf{.0186} \cellcolor{green!35} & .1154 \cellcolor{red!35} & .0331$^{\phantom{\ddag}}$ \cellcolor{green!27} & .0292 \cellcolor{green!30} & \textbf{.0231}$^{\phantom{\ddag}}$ \cellcolor{green!35} & .1008 \cellcolor{red!35} & .0399 \cellcolor{green!21} & .0321 \cellcolor{green!28} & \textbf{.0255} \cellcolor{green!35} \\
\texttt{Ohsumed} & .1004 \cellcolor{red!35} & .0177 \cellcolor{green!34} & .0183 \cellcolor{green!33} & \textbf{.0168} \cellcolor{green!35} & .1087 \cellcolor{red!35} & .0262$^{\phantom{\ddag}}$ \cellcolor{green!30} & \textbf{.0209} \cellcolor{green!35} & .0238$^{\phantom{\ddag}}$ \cellcolor{green!32} & .1177 \cellcolor{red!35} & .0321 \cellcolor{green!27} & \textbf{.0215} \cellcolor{green!35} & .0284 \cellcolor{green!30} \\
\texttt{Enron} & .0347 \cellcolor{red!35} & \textbf{.0161} \cellcolor{green!35} & .0169 \cellcolor{green!31} & .0185 \cellcolor{green!25} & .0397 \cellcolor{red!35} & .0235$^{\phantom{\ddag}}$ \cellcolor{green!31} & \textbf{.0227} \cellcolor{green!35} & .0228$^{\dag}$ \cellcolor{green!34} & .0439 \cellcolor{red!35} & .0287 \cellcolor{green!20} & .0253 \cellcolor{green!32} & \textbf{.0246} \cellcolor{green!35} \\
\texttt{Reuters-21578} & .0167 \cellcolor{red!35} & \textbf{.0036} \cellcolor{green!35} & .0049 \cellcolor{green!27} & .0037 \cellcolor{green!34} & .0243 \cellcolor{red!35} & \textbf{.0059}$^{\phantom{\ddag}}$ \cellcolor{green!35} & .0070 \cellcolor{green!30} & .0060$^{\phantom{\ddag}}$ \cellcolor{green!34} & .0370 \cellcolor{red!35} & \textbf{.0075} \cellcolor{green!35} & .0088 \cellcolor{green!31} & .0078 \cellcolor{green!34} \\
\texttt{RCV1-v2} & .0456 \cellcolor{red!35} & \textbf{.0084} \cellcolor{green!35} & .0093 \cellcolor{green!33} & .0084 \cellcolor{green!34} & .0533 \cellcolor{red!35} & .0129$^{\dag}$ \cellcolor{green!34} & .0146 \cellcolor{green!31} & \textbf{.0128}$^{\phantom{\ddag}}$ \cellcolor{green!35} & .0654 \cellcolor{red!35} & \textbf{.0198} \cellcolor{green!35} & .0215 \cellcolor{green!32} & .0201 \cellcolor{green!34} \\
\texttt{Mediamill} & .1697 \cellcolor{red!35} & .0154 \cellcolor{green!34} & .0157 \cellcolor{green!34} & \textbf{.0148} \cellcolor{green!35} & .1736 \cellcolor{red!35} & .0285$^{\phantom{\ddag}}$ \cellcolor{green!32} & .0251 \cellcolor{green!34} & \textbf{.0231}$^{\phantom{\ddag}}$ \cellcolor{green!35} & .1806 \cellcolor{red!35} & .0401 \cellcolor{green!29} & .0322 \cellcolor{green!33} & \textbf{.0285} \cellcolor{green!35} \\
\texttt{Bibtex} & .0354 \cellcolor{red!35} & \textbf{.0091} \cellcolor{green!35} & .0092 \cellcolor{green!34} & .0093 \cellcolor{green!34} & .0374 \cellcolor{red!35} & .0126$^{\phantom{\ddag}}$ \cellcolor{green!32} & \textbf{.0116} \cellcolor{green!35} & .0120$^{\phantom{\ddag}}$ \cellcolor{green!34} & .0423 \cellcolor{red!35} & .0152 \cellcolor{green!29} & \textbf{.0129} \cellcolor{green!35} & .0142 \cellcolor{green!31} \\
\texttt{Corel5k} & .0582 \cellcolor{red!35} & .0077 \cellcolor{green!33} & .0075 \cellcolor{green!33} & \textbf{.0066} \cellcolor{green!35} & .0585 \cellcolor{red!35} & .0083$^{\phantom{\ddag}}$ \cellcolor{green!33} & .0082 \cellcolor{green!33} & \textbf{.0074}$^{\phantom{\ddag}}$ \cellcolor{green!35} & .0594 \cellcolor{red!35} & .0089 \cellcolor{green!34} & .0087 \cellcolor{green!34} & \textbf{.0085} \cellcolor{green!35} \\
\texttt{Delicious} & .1420 \cellcolor{red!35} & .0088 \cellcolor{green!34} & .0093 \cellcolor{green!34} & \textbf{.0084} \cellcolor{green!35} & .1417 \cellcolor{red!35} & .0117$^{\phantom{\ddag}}$ \cellcolor{green!34} & .0109 \cellcolor{green!34} & \textbf{.0098}$^{\phantom{\ddag}}$ \cellcolor{green!35} & .1238 \cellcolor{red!35} & .0119 \cellcolor{green!33} & .0104 \cellcolor{green!34} & \textbf{.0092} \cellcolor{green!35} \\
\hline
\multicolumn{1}{|c|}{Average} & .0677 \cellcolor{red!35} & \textbf{.0158} \cellcolor{green!35} & .0177 \cellcolor{green!32} & .0160 \cellcolor{green!34} & .0761 \cellcolor{red!35} & .0312$^{\phantom{\ddag}}$ \cellcolor{green!27} & .0288 \cellcolor{green!31} & \textbf{.0261}$^{\phantom{\ddag}}$ \cellcolor{green!35} & .1012 \cellcolor{red!35} & .0602 \cellcolor{green!12} & .0452 \cellcolor{green!30} & \textbf{.0414} \cellcolor{green!35} \\
\multicolumn{1}{|c|}{Rank Average} &  3.6 \cellcolor{red!35} & \textbf{1.7} \cellcolor{green!35} &  2.9 \cellcolor{red!7} &  1.8 \cellcolor{green!32} &  3.7 \cellcolor{red!35} &  2.5 \cellcolor{green!2} &  2.3 \cellcolor{green!10} & \textbf{1.5} \cellcolor{green!35} &  3.7 \cellcolor{red!35} &  2.4 \cellcolor{green!7} &  2.3 \cellcolor{green!9} & \textbf{1.5} \cellcolor{green!35} \\

                \hline
                \end{tabular}%
                }
            
\end{table}

\begin{table}[tb]
  \centering
  \caption{Values of \abse\ obtained in our experiments for different
  amounts of shift using ACC as the base quantifier. Notational
  conventions are as in Table~\ref{tab:cc_general}.}
  \label{tab:acc_general}
  
                \resizebox{\textwidth}{!}{%
                        \begin{tabular}{|l|cccc|cccc|cccc|} \cline{2-13}
\multicolumn{1}{c|}{} & \multicolumn{4}{c|}{low shift} & \multicolumn{4}{c|}{mid shift} & \multicolumn{4}{c|}{high shift}\\\cline{2-13}

\multicolumn{1}{c|}{} & \side{\bqbc} & \side{\bqmc} & \side{\mqbc} & \side{\mqmc\phantom{0}} & \side{\bqbc} & \side{\bqmc} & \side{\mqbc} & \side{\mqmc\phantom{0}} & \side{\bqbc} & \side{\bqmc} & \side{\mqbc} & \side{\mqmc\phantom{0}}\\\hline
\texttt{Emotions} & .1060 \cellcolor{red!16} & .1279 \cellcolor{red!35} & .0626 \cellcolor{green!19} & \textbf{.0440} \cellcolor{green!35} & .1367 \cellcolor{red!25} & .1487 \cellcolor{red!35} & .0732$^{\phantom{\ddag}}$ \cellcolor{green!21} & \textbf{.0551}$^{\phantom{\ddag}}$ \cellcolor{green!35} & .1608 \cellcolor{red!35} & .1602$^{\phantom{\ddag}}$ \cellcolor{red!34} & .1125 \cellcolor{green!4} & \textbf{.0750}$^{\phantom{\ddag}}$ \cellcolor{green!35} \\
\texttt{Scene} & .0537 \cellcolor{red!35} & .0436 \cellcolor{green!5} & .0456 \cellcolor{red!2} & \textbf{.0361} \cellcolor{green!35} & .0626 \cellcolor{red!35} & .0601 \cellcolor{red!19} & \textbf{.0512}$^{\phantom{\ddag}}$ \cellcolor{green!35} & .0513$^{\ddag}$ \cellcolor{green!34} & .0615 \cellcolor{green!23} & .0603$^{\ddag}$ \cellcolor{green!30} & \textbf{.0596} \cellcolor{green!35} & .0711$^{\phantom{\ddag}}$ \cellcolor{red!35} \\
\texttt{Yeast} & .1937 \cellcolor{red!35} & .1765 \cellcolor{red!26} & \textbf{.0506} \cellcolor{green!35} & .0531 \cellcolor{green!33} & .2164 \cellcolor{red!33} & .2191 \cellcolor{red!35} & \textbf{.0896}$^{\phantom{\ddag}}$ \cellcolor{green!35} & .0957$^{\phantom{\ddag}}$ \cellcolor{green!31} & .2301 \cellcolor{red!19} & .2620$^{\phantom{\ddag}}$ \cellcolor{red!35} & \textbf{.1204} \cellcolor{green!35} & .1480$^{\phantom{\ddag}}$ \cellcolor{green!21} \\
\texttt{Birds} & .1542 \cellcolor{red!35} & .1353 \cellcolor{red!24} & \textbf{.0233} \cellcolor{green!35} & .0278 \cellcolor{green!32} & .1526 \cellcolor{red!35} & .1384 \cellcolor{red!26} & \textbf{.0286}$^{\phantom{\ddag}}$ \cellcolor{green!35} & .0307$^{\phantom{\ddag}}$ \cellcolor{green!33} & .1547 \cellcolor{red!35} & .1472$^{\phantom{\ddag}}$ \cellcolor{red!30} & \textbf{.0359} \cellcolor{green!35} & .0391$^{\phantom{\ddag}}$ \cellcolor{green!33} \\
\texttt{Genbase} & .0014 \cellcolor{green!16} & .0007 \cellcolor{green!34} & .0036 \cellcolor{red!35} & \textbf{.0006} \cellcolor{green!35} & .0014 \cellcolor{green!19} & \textbf{.0007} \cellcolor{green!35} & .0039$^{\phantom{\ddag}}$ \cellcolor{red!35} & .0007$^{\ddag}$ \cellcolor{green!34} & .0013 \cellcolor{green!20} & .0006$^{\ddag}$ \cellcolor{green!34} & .0037 \cellcolor{red!35} & \textbf{.0006}$^{\phantom{\ddag}}$ \cellcolor{green!35} \\
\texttt{Medical} & .0212 \cellcolor{green!18} & .0307 \cellcolor{red!35} & .0187 \cellcolor{green!32} & \textbf{.0183} \cellcolor{green!35} & \textbf{.0189} \cellcolor{green!35} & .0329 \cellcolor{red!35} & .0259$^{\phantom{\ddag}}$ \cellcolor{green!0} & .0248$^{\phantom{\ddag}}$ \cellcolor{green!5} & \textbf{.0153} \cellcolor{green!35} & .0329$^{\phantom{\ddag}}$ \cellcolor{red!35} & .0311 \cellcolor{red!27} & .0294$^{\phantom{\ddag}}$ \cellcolor{red!21} \\
\texttt{tmc2007\_500} & .0366 \cellcolor{red!35} & .0365 \cellcolor{red!34} & \textbf{.0220} \cellcolor{green!35} & .0222 \cellcolor{green!33} & .0612 \cellcolor{red!35} & .0581 \cellcolor{red!28} & .0302$^{\phantom{\ddag}}$ \cellcolor{green!29} & \textbf{.0278}$^{\phantom{\ddag}}$ \cellcolor{green!35} & .0647 \cellcolor{red!35} & .0591$^{\phantom{\ddag}}$ \cellcolor{red!23} & .0339 \cellcolor{green!29} & \textbf{.0315}$^{\phantom{\ddag}}$ \cellcolor{green!35} \\
\texttt{Ohsumed} & .0253 \cellcolor{red!35} & .0241 \cellcolor{red!23} & .0198 \cellcolor{green!18} & \textbf{.0180} \cellcolor{green!35} & .0320 \cellcolor{red!35} & .0292 \cellcolor{red!14} & .0228$^{\dag}$ \cellcolor{green!33} & \textbf{.0226}$^{\phantom{\ddag}}$ \cellcolor{green!35} & .0407 \cellcolor{red!35} & .0337$^{\phantom{\ddag}}$ \cellcolor{red!5} & \textbf{.0239} \cellcolor{green!35} & .0263$^{\phantom{\ddag}}$ \cellcolor{green!24} \\
\texttt{Enron} & .1810 \cellcolor{red!35} & .0935 \cellcolor{green!2} & \textbf{.0187} \cellcolor{green!35} & .0207 \cellcolor{green!34} & .1882 \cellcolor{red!35} & .0982 \cellcolor{green!3} & \textbf{.0243}$^{\phantom{\ddag}}$ \cellcolor{green!35} & .0272$^{\phantom{\ddag}}$ \cellcolor{green!33} & .1853 \cellcolor{red!35} & .1043$^{\phantom{\ddag}}$ \cellcolor{green!0} & \textbf{.0269} \cellcolor{green!35} & .0339$^{\phantom{\ddag}}$ \cellcolor{green!31} \\
\texttt{Reuters-21578} & .0307 \cellcolor{red!35} & .0074 \cellcolor{green!29} & \textbf{.0055} \cellcolor{green!35} & .0065 \cellcolor{green!32} & .0336 \cellcolor{red!35} & .0121 \cellcolor{green!23} & \textbf{.0078}$^{\phantom{\ddag}}$ \cellcolor{green!35} & .0096$^{\phantom{\ddag}}$ \cellcolor{green!30} & .0421 \cellcolor{red!35} & .0254$^{\phantom{\ddag}}$ \cellcolor{green!1} & \textbf{.0104} \cellcolor{green!35} & .0123$^{\phantom{\ddag}}$ \cellcolor{green!30} \\
\texttt{RCV1-v2} & .0124 \cellcolor{green!19} & .0217 \cellcolor{red!35} & \textbf{.0099} \cellcolor{green!35} & .0106 \cellcolor{green!30} & .0189 \cellcolor{green!13} & .0259 \cellcolor{red!35} & \textbf{.0158}$^{\phantom{\ddag}}$ \cellcolor{green!35} & .0167$^{\phantom{\ddag}}$ \cellcolor{green!28} & .0287 \cellcolor{green!2} & .0335$^{\phantom{\ddag}}$ \cellcolor{red!35} & \textbf{.0246} \cellcolor{green!35} & .0269$^{\phantom{\ddag}}$ \cellcolor{green!16} \\
\texttt{Mediamill} & .0539 \cellcolor{red!35} & .0425 \cellcolor{red!13} & .0164 \cellcolor{green!34} & \textbf{.0163} \cellcolor{green!35} & .0976 \cellcolor{red!35} & .0539 \cellcolor{green!6} & \textbf{.0246}$^{\phantom{\ddag}}$ \cellcolor{green!35} & .0274$^{\phantom{\ddag}}$ \cellcolor{green!32} & .1467 \cellcolor{red!35} & .0647$^{\phantom{\ddag}}$ \cellcolor{green!14} & \textbf{.0316} \cellcolor{green!35} & .0374$^{\phantom{\ddag}}$ \cellcolor{green!31} \\
\texttt{Bibtex} & .0692 \cellcolor{red!22} & .0816 \cellcolor{red!35} & \textbf{.0102} \cellcolor{green!35} & .0107 \cellcolor{green!34} & .0734 \cellcolor{red!23} & .0858 \cellcolor{red!35} & \textbf{.0125}$^{\phantom{\ddag}}$ \cellcolor{green!35} & .0152$^{\phantom{\ddag}}$ \cellcolor{green!32} & .0861 \cellcolor{red!31} & .0903$^{\phantom{\ddag}}$ \cellcolor{red!35} & \textbf{.0146} \cellcolor{green!35} & .0196$^{\phantom{\ddag}}$ \cellcolor{green!30} \\
\texttt{Corel5k} & .1515 \cellcolor{red!35} & .0173 \cellcolor{green!29} & .0081 \cellcolor{green!34} & \textbf{.0062} \cellcolor{green!35} & .1537 \cellcolor{red!35} & .0158 \cellcolor{green!31} & .0089$^{\phantom{\ddag}}$ \cellcolor{green!34} & \textbf{.0076}$^{\phantom{\ddag}}$ \cellcolor{green!35} & .1509 \cellcolor{red!35} & .0149$^{\phantom{\ddag}}$ \cellcolor{green!32} & .0099 \cellcolor{green!34} & \textbf{.0094}$^{\phantom{\ddag}}$ \cellcolor{green!35} \\
\texttt{Delicious} & .0846 \cellcolor{red!35} & .0528 \cellcolor{red!5} & .0097 \cellcolor{green!34} & \textbf{.0093} \cellcolor{green!35} & .1000 \cellcolor{red!35} & .0531 \cellcolor{green!1} & .0111$^{\phantom{\ddag}}$ \cellcolor{green!34} & \textbf{.0110}$^{\phantom{\ddag}}$ \cellcolor{green!35} & .1016 \cellcolor{red!35} & .0492$^{\phantom{\ddag}}$ \cellcolor{green!5} & \textbf{.0106} \cellcolor{green!35} & .0109$^{\dag}$ \cellcolor{green!34} \\
\hline
\multicolumn{1}{|c|}{Average} & .0750 \cellcolor{red!35} & .0558 \cellcolor{red!11} & .0193 \cellcolor{green!33} & \textbf{.0182} \cellcolor{green!35} & .0841 \cellcolor{red!35} & .0716 \cellcolor{red!18} & .0302$^{\phantom{\ddag}}$ \cellcolor{green!34} & \textbf{.0296}$^{\phantom{\ddag}}$ \cellcolor{green!35} & .0901 \cellcolor{red!35} & .0808$^{\phantom{\ddag}}$ \cellcolor{red!19} & \textbf{.0482} \cellcolor{green!35} & .0495$^{\phantom{\ddag}}$ \cellcolor{green!32} \\
\multicolumn{1}{|c|}{Rank Average} &  3.7 \cellcolor{red!35} &  3.1 \cellcolor{red!18} &  1.7 \cellcolor{green!26} & \textbf{1.5} \cellcolor{green!35} &  3.5 \cellcolor{red!35} &  3.2 \cellcolor{red!24} & \textbf{1.7} \cellcolor{green!35} &  1.7 \cellcolor{green!35} &  3.5 \cellcolor{red!35} &  3.1 \cellcolor{red!22} & \textbf{1.5} \cellcolor{green!35} &  1.9 \cellcolor{green!22} \\

                \hline
                \end{tabular}%
                }
            
\end{table}

\begin{table}[tb]
  \centering
  \caption{Values of \abse\ obtained in our experiments for different
  amounts of shift using PACC as the base quantifier. Notational
  conventions are as in Table~\ref{tab:cc_general}.}
  \label{tab:pacc_general}
  
                \resizebox{\textwidth}{!}{%
                        \begin{tabular}{|l|cccc|cccc|cccc|} \cline{2-13}
\multicolumn{1}{c|}{} & \multicolumn{4}{c|}{low shift} & \multicolumn{4}{c|}{mid shift} & \multicolumn{4}{c|}{high shift}\\\cline{2-13}

\multicolumn{1}{c|}{} & \side{\bqbc} & \side{\bqmc} & \side{\mqbc} & \side{\mqmc\phantom{0}} & \side{\bqbc} & \side{\bqmc} & \side{\mqbc} & \side{\mqmc\phantom{0}} & \side{\bqbc} & \side{\bqmc} & \side{\mqbc} & \side{\mqmc\phantom{0}}\\\hline
\texttt{Emotions} & .1326 \cellcolor{red!35} & .0715 \cellcolor{green!17} & \textbf{.0509}$^{\phantom{\ddag}}$ \cellcolor{green!35} & .0516 \cellcolor{green!34} & .1451 \cellcolor{red!35} & .1049 \cellcolor{red!0} & .0659$^{\phantom{\ddag}}$ \cellcolor{green!32} & \textbf{.0633} \cellcolor{green!35} & .1578 \cellcolor{red!35} & .1485$^{\phantom{\ddag}}$ \cellcolor{red!25} & .1101$^{\phantom{\ddag}}$ \cellcolor{green!12} & \textbf{.0872}$^{\phantom{\ddag}}$ \cellcolor{green!35} \\
\texttt{Scene} & .0508 \cellcolor{red!35} & .0379 \cellcolor{green!10} & .0391$^{\phantom{\ddag}}$ \cellcolor{green!6} & \textbf{.0310} \cellcolor{green!35} & .0697 \cellcolor{red!35} & .0498 \cellcolor{green!17} & \textbf{.0433}$^{\phantom{\ddag}}$ \cellcolor{green!35} & .0441 \cellcolor{green!32} & .0757 \cellcolor{red!35} & .0499$^{\ddag}$ \cellcolor{green!34} & \textbf{.0496}$^{\phantom{\ddag}}$ \cellcolor{green!35} & .0645$^{\phantom{\ddag}}$ \cellcolor{red!4} \\
\texttt{Yeast} & .1654 \cellcolor{red!35} & .1436 \cellcolor{red!21} & \textbf{.0489}$^{\phantom{\ddag}}$ \cellcolor{green!35} & .0494 \cellcolor{green!34} & .2122 \cellcolor{red!35} & .1803 \cellcolor{red!17} & .0878$^{\phantom{\ddag}}$ \cellcolor{green!32} & \textbf{.0837} \cellcolor{green!35} & .2474 \cellcolor{red!35} & .2053$^{\phantom{\ddag}}$ \cellcolor{red!10} & .1316$^{\phantom{\ddag}}$ \cellcolor{green!31} & \textbf{.1264}$^{\phantom{\ddag}}$ \cellcolor{green!35} \\
\texttt{Birds} & .1253 \cellcolor{red!32} & .1290 \cellcolor{red!35} & \textbf{.0191}$^{\phantom{\ddag}}$ \cellcolor{green!35} & .0227 \cellcolor{green!32} & .1243 \cellcolor{red!32} & .1286 \cellcolor{red!35} & \textbf{.0244}$^{\phantom{\ddag}}$ \cellcolor{green!35} & .0282 \cellcolor{green!32} & .1250 \cellcolor{red!35} & .1173$^{\phantom{\ddag}}$ \cellcolor{red!28} & \textbf{.0377}$^{\phantom{\ddag}}$ \cellcolor{green!35} & .0434$^{\phantom{\ddag}}$ \cellcolor{green!30} \\
\texttt{Genbase} & .0018 \cellcolor{green!17} & \textbf{.0010} \cellcolor{green!35} & .0041$^{\phantom{\ddag}}$ \cellcolor{red!35} & .0016 \cellcolor{green!22} & .0018 \cellcolor{green!19} & \textbf{.0010} \cellcolor{green!35} & .0043$^{\phantom{\ddag}}$ \cellcolor{red!35} & .0015 \cellcolor{green!25} & .0018 \cellcolor{green!16} & \textbf{.0010}$^{\phantom{\ddag}}$ \cellcolor{green!35} & .0041$^{\phantom{\ddag}}$ \cellcolor{red!35} & .0015$^{\phantom{\ddag}}$ \cellcolor{green!24} \\
\texttt{Medical} & .0395 \cellcolor{red!35} & .0272 \cellcolor{green!1} & .0169$^{\phantom{\ddag}}$ \cellcolor{green!32} & \textbf{.0161} \cellcolor{green!35} & .0477 \cellcolor{red!35} & .0276 \cellcolor{green!16} & \textbf{.0204}$^{\phantom{\ddag}}$ \cellcolor{green!35} & .0222 \cellcolor{green!30} & .0484 \cellcolor{red!35} & \textbf{.0267}$^{\phantom{\ddag}}$ \cellcolor{green!35} & .0286$^{\dag}$ \cellcolor{green!29} & .0286$^{\phantom{\ddag}}$ \cellcolor{green!29} \\
\texttt{tmc2007\_500} & .0351 \cellcolor{red!35} & .0285 \cellcolor{red!6} & .0213$^{\phantom{\ddag}}$ \cellcolor{green!25} & \textbf{.0192} \cellcolor{green!35} & .0617 \cellcolor{red!35} & .0454 \cellcolor{red!4} & .0281$^{\phantom{\ddag}}$ \cellcolor{green!26} & \textbf{.0236} \cellcolor{green!35} & .0619 \cellcolor{red!35} & .0466$^{\phantom{\ddag}}$ \cellcolor{red!3} & .0320$^{\phantom{\ddag}}$ \cellcolor{green!25} & \textbf{.0275}$^{\phantom{\ddag}}$ \cellcolor{green!35} \\
\texttt{Ohsumed} & .0239 \cellcolor{red!35} & .0221 \cellcolor{red!14} & .0189$^{\phantom{\ddag}}$ \cellcolor{green!22} & \textbf{.0179} \cellcolor{green!35} & .0345 \cellcolor{red!35} & .0285 \cellcolor{red!4} & \textbf{.0205}$^{\phantom{\ddag}}$ \cellcolor{green!35} & .0234 \cellcolor{green!20} & .0452 \cellcolor{red!35} & .0332$^{\phantom{\ddag}}$ \cellcolor{green!0} & \textbf{.0216}$^{\phantom{\ddag}}$ \cellcolor{green!35} & .0278$^{\phantom{\ddag}}$ \cellcolor{green!16} \\
\texttt{Enron} & .1433 \cellcolor{red!35} & .1429 \cellcolor{red!34} & \textbf{.0198}$^{\phantom{\ddag}}$ \cellcolor{green!35} & .0201 \cellcolor{green!34} & .1618 \cellcolor{red!35} & .1539 \cellcolor{red!31} & .0243$^{\dag}$ \cellcolor{green!34} & \textbf{.0239} \cellcolor{green!35} & .1504 \cellcolor{red!25} & .1693$^{\phantom{\ddag}}$ \cellcolor{red!35} & .0275$^{\phantom{\ddag}}$ \cellcolor{green!33} & \textbf{.0253}$^{\phantom{\ddag}}$ \cellcolor{green!35} \\
\texttt{Reuters-21578} & .0086 \cellcolor{green!29} & .0465 \cellcolor{red!35} & \textbf{.0053}$^{\phantom{\ddag}}$ \cellcolor{green!35} & .0054 \cellcolor{green!34} & .0311 \cellcolor{red!2} & .0519 \cellcolor{red!35} & \textbf{.0076}$^{\phantom{\ddag}}$ \cellcolor{green!35} & .0087 \cellcolor{green!33} & .0643 \cellcolor{red!35} & .0619$^{\phantom{\ddag}}$ \cellcolor{red!31} & \textbf{.0113}$^{\phantom{\ddag}}$ \cellcolor{green!35} & .0117$^{\phantom{\ddag}}$ \cellcolor{green!34} \\
\texttt{RCV1-v2} & .0173 \cellcolor{red!35} & .0130 \cellcolor{green!2} & .0098$^{\phantom{\ddag}}$ \cellcolor{green!30} & \textbf{.0093} \cellcolor{green!35} & .0334 \cellcolor{red!35} & .0228 \cellcolor{green!4} & .0159$^{\phantom{\ddag}}$ \cellcolor{green!30} & \textbf{.0148} \cellcolor{green!35} & .0505 \cellcolor{red!35} & .0342$^{\phantom{\ddag}}$ \cellcolor{green!6} & .0243$^{\phantom{\ddag}}$ \cellcolor{green!32} & \textbf{.0232}$^{\phantom{\ddag}}$ \cellcolor{green!35} \\
\texttt{Mediamill} & .0482 \cellcolor{red!35} & .0466 \cellcolor{red!31} & .0158$^{\dag}$ \cellcolor{green!34} & \textbf{.0158} \cellcolor{green!35} & .0941 \cellcolor{red!35} & .0779 \cellcolor{red!18} & .0237$^{\phantom{\ddag}}$ \cellcolor{green!34} & \textbf{.0235} \cellcolor{green!35} & .1387 \cellcolor{red!35} & .1076$^{\phantom{\ddag}}$ \cellcolor{red!14} & \textbf{.0299}$^{\phantom{\ddag}}$ \cellcolor{green!35} & .0300$^{\ddag}$ \cellcolor{green!34} \\
\texttt{Bibtex} & .0320 \cellcolor{red!35} & .0218 \cellcolor{red!2} & \textbf{.0099}$^{\phantom{\ddag}}$ \cellcolor{green!35} & .0100 \cellcolor{green!34} & .0459 \cellcolor{red!35} & .0387 \cellcolor{red!20} & \textbf{.0121}$^{\phantom{\ddag}}$ \cellcolor{green!35} & .0133 \cellcolor{green!32} & .0684 \cellcolor{red!35} & .0662$^{\phantom{\ddag}}$ \cellcolor{red!32} & \textbf{.0140}$^{\phantom{\ddag}}$ \cellcolor{green!35} & .0161$^{\phantom{\ddag}}$ \cellcolor{green!32} \\
\texttt{Corel5k} & .1519 \cellcolor{red!35} & .1014 \cellcolor{red!9} & \textbf{.0107}$^{\phantom{\ddag}}$ \cellcolor{green!35} & .0111 \cellcolor{green!34} & .1559 \cellcolor{red!35} & .1087 \cellcolor{red!12} & .0112$^{\ddag}$ \cellcolor{green!34} & \textbf{.0112} \cellcolor{green!35} & .1711 \cellcolor{red!35} & .1221$^{\phantom{\ddag}}$ \cellcolor{red!13} & .0120$^{\phantom{\ddag}}$ \cellcolor{green!34} & \textbf{.0116}$^{\phantom{\ddag}}$ \cellcolor{green!35} \\
\texttt{Delicious} & .0594 \cellcolor{red!35} & .0480 \cellcolor{red!18} & .0094$^{\phantom{\ddag}}$ \cellcolor{green!34} & \textbf{.0092} \cellcolor{green!35} & .0837 \cellcolor{red!35} & .0828 \cellcolor{red!34} & .0107$^{\phantom{\ddag}}$ \cellcolor{green!34} & \textbf{.0105} \cellcolor{green!35} & .0871 \cellcolor{red!31} & .0914$^{\phantom{\ddag}}$ \cellcolor{red!35} & .0101$^{\dag}$ \cellcolor{green!34} & \textbf{.0099}$^{\phantom{\ddag}}$ \cellcolor{green!35} \\
\hline
\multicolumn{1}{|c|}{Average} & .0646 \cellcolor{red!35} & .0561 \cellcolor{red!22} & .0181$^{\phantom{\ddag}}$ \cellcolor{green!34} & \textbf{.0175} \cellcolor{green!35} & .0823 \cellcolor{red!35} & .0701 \cellcolor{red!19} & .0279$^{\phantom{\ddag}}$ \cellcolor{green!34} & \textbf{.0276} \cellcolor{green!35} & .0958 \cellcolor{red!35} & .0782$^{\phantom{\ddag}}$ \cellcolor{red!10} & \textbf{.0463}$^{\phantom{\ddag}}$ \cellcolor{green!35} & .0467$^{\dag}$ \cellcolor{green!34} \\
\multicolumn{1}{|c|}{Rank Average} &  3.8 \cellcolor{red!35} &  2.9 \cellcolor{red!8} &  1.7 \cellcolor{green!28} & \textbf{1.5} \cellcolor{green!35} &  3.8 \cellcolor{red!35} &  3.0 \cellcolor{red!11} &  1.7 \cellcolor{green!26} & \textbf{1.5} \cellcolor{green!35} &  3.8 \cellcolor{red!35} &  2.8 \cellcolor{red!2} &  1.7 \cellcolor{green!32} & \textbf{1.7} \cellcolor{green!35} \\

                \hline
                \end{tabular}%
                }
            
\end{table}

\begin{table}[tb]
  \centering
  \caption{Values of \abse\ obtained in our experiments for different
  amounts of shift using SLD as the base quantifier. Notational
  conventions are as in Table~\ref{tab:cc_general}.}
  \label{tab:sld_general}
  
                \resizebox{\textwidth}{!}{%
                    \begin{tabular}{|l|cccc|cccc|cccc|} \cline{2-13}
\multicolumn{1}{c|}{} & \multicolumn{4}{c|}{low shift} & \multicolumn{4}{c|}{mid shift} & \multicolumn{4}{c|}{high shift}\\\cline{2-13}

\multicolumn{1}{c|}{} & \side{\bqbc} & \side{\bqmc} & \side{\mqbc} & \side{\mqmc\phantom{0}} & \side{\bqbc} & \side{\bqmc} & \side{\mqbc} & \side{\mqmc\phantom{0}} & \side{\bqbc} & \side{\bqmc} & \side{\mqbc} & \side{\mqmc\phantom{0}}\\\hline
\texttt{Emotions} & .2169 \cellcolor{red!35} & .0549 \cellcolor{green!33} & .0710$^{\phantom{\ddag}}$ \cellcolor{green!26} & \textbf{.0509} \cellcolor{green!35} & .2189 \cellcolor{red!35} & .0719$^{\phantom{\ddag}}$ \cellcolor{green!31} & .0791 \cellcolor{green!28} & \textbf{.0652} \cellcolor{green!35} & .2088 \cellcolor{red!35} & .0890$^{\phantom{\ddag}}$ \cellcolor{green!26} & .0822$^{\phantom{\ddag}}$ \cellcolor{green!29} & \textbf{.0717}$^{\phantom{\ddag}}$ \cellcolor{green!35} \\
\texttt{Scene} & .0407 \cellcolor{red!16} & .0433 \cellcolor{red!35} & \textbf{.0337}$^{\phantom{\ddag}}$ \cellcolor{green!35} & .0424 \cellcolor{red!28} & \textbf{.0467} \cellcolor{green!35} & .0753$^{\phantom{\ddag}}$ \cellcolor{red!35} & .0497 \cellcolor{green!27} & .0709 \cellcolor{red!24} & \textbf{.0487} \cellcolor{green!35} & .1012$^{\phantom{\ddag}}$ \cellcolor{red!35} & .0628$^{\phantom{\ddag}}$ \cellcolor{green!16} & .0881$^{\phantom{\ddag}}$ \cellcolor{red!17} \\
\texttt{Yeast} & .2557 \cellcolor{red!35} & .0948 \cellcolor{green!19} & .0511$^{\phantom{\ddag}}$ \cellcolor{green!34} & \textbf{.0500} \cellcolor{green!35} & .2607 \cellcolor{red!35} & .1192$^{\phantom{\ddag}}$ \cellcolor{green!20} & .0889 \cellcolor{green!32} & \textbf{.0827} \cellcolor{green!35} & .2939 \cellcolor{red!35} & .1438$^{\phantom{\ddag}}$ \cellcolor{green!24} & .1362$^{\phantom{\ddag}}$ \cellcolor{green!27} & \textbf{.1171}$^{\phantom{\ddag}}$ \cellcolor{green!35} \\
\texttt{Birds} & .0759 \cellcolor{red!35} & .0284 \cellcolor{green!24} & \textbf{.0196}$^{\phantom{\ddag}}$ \cellcolor{green!35} & .0281 \cellcolor{green!24} & .0819 \cellcolor{red!35} & .0312$^{\phantom{\ddag}}$ \cellcolor{green!27} & \textbf{.0255} \cellcolor{green!35} & .0312 \cellcolor{green!27} & .1089 \cellcolor{red!35} & .0355$^{\dag}$ \cellcolor{green!34} & .0358$^{\dag}$ \cellcolor{green!34} & \textbf{.0351}$^{\phantom{\ddag}}$ \cellcolor{green!35} \\
\texttt{Genbase} & .0011 \cellcolor{green!20} & \textbf{.0004} \cellcolor{green!35} & .0039$^{\phantom{\ddag}}$ \cellcolor{red!35} & .0005 \cellcolor{green!31} & .0011 \cellcolor{green!21} & \textbf{.0003}$^{\phantom{\ddag}}$ \cellcolor{green!35} & .0042 \cellcolor{red!35} & .0005 \cellcolor{green!31} & .0010 \cellcolor{green!22} & \textbf{.0003}$^{\phantom{\ddag}}$ \cellcolor{green!35} & .0041$^{\phantom{\ddag}}$ \cellcolor{red!35} & .0005$^{\phantom{\ddag}}$ \cellcolor{green!31} \\
\texttt{Medical} & .0233 \cellcolor{red!35} & .0133 \cellcolor{green!32} & .0190$^{\phantom{\ddag}}$ \cellcolor{red!6} & \textbf{.0129} \cellcolor{green!35} & .0211 \cellcolor{red!7} & .0135$^{\phantom{\ddag}}$ \cellcolor{green!33} & .0263 \cellcolor{red!35} & \textbf{.0131} \cellcolor{green!35} & .0189 \cellcolor{green!13} & .0133$^{\phantom{\ddag}}$ \cellcolor{green!34} & .0312$^{\phantom{\ddag}}$ \cellcolor{red!35} & \textbf{.0132}$^{\phantom{\ddag}}$ \cellcolor{green!35} \\
\texttt{tmc2007\_500} & .0384 \cellcolor{red!35} & .0248 \cellcolor{green!13} & .0202$^{\phantom{\ddag}}$ \cellcolor{green!29} & \textbf{.0187} \cellcolor{green!35} & .0526 \cellcolor{red!35} & .0407$^{\phantom{\ddag}}$ \cellcolor{red!6} & .0285 \cellcolor{green!22} & \textbf{.0230} \cellcolor{green!35} & .0546 \cellcolor{red!35} & .0432$^{\phantom{\ddag}}$ \cellcolor{red!10} & .0330$^{\phantom{\ddag}}$ \cellcolor{green!12} & \textbf{.0228}$^{\phantom{\ddag}}$ \cellcolor{green!35} \\
\texttt{Ohsumed} & .0294 \cellcolor{red!35} & .0186 \cellcolor{green!27} & \textbf{.0173}$^{\phantom{\ddag}}$ \cellcolor{green!35} & .0185 \cellcolor{green!27} & .0316 \cellcolor{red!35} & .0232$^{\phantom{\ddag}}$ \cellcolor{green!11} & \textbf{.0189} \cellcolor{green!35} & .0232 \cellcolor{green!11} & .0321 \cellcolor{red!35} & .0250$^{\phantom{\ddag}}$ \cellcolor{green!5} & \textbf{.0200}$^{\phantom{\ddag}}$ \cellcolor{green!35} & .0250$^{\phantom{\ddag}}$ \cellcolor{green!6} \\
\texttt{Enron} & .0918 \cellcolor{red!35} & .0208 \cellcolor{green!32} & .0183$^{\phantom{\ddag}}$ \cellcolor{green!34} & \textbf{.0182} \cellcolor{green!35} & .0915 \cellcolor{red!35} & .0253$^{\phantom{\ddag}}$ \cellcolor{green!33} & \textbf{.0238} \cellcolor{green!35} & .0243 \cellcolor{green!34} & .0838 \cellcolor{red!35} & .0261$^{\dag}$ \cellcolor{green!34} & \textbf{.0258}$^{\phantom{\ddag}}$ \cellcolor{green!35} & .0263$^{\dag}$ \cellcolor{green!34} \\
\texttt{Reuters-21578} & .0050 \cellcolor{red!35} & \textbf{.0039} \cellcolor{green!35} & .0048$^{\phantom{\ddag}}$ \cellcolor{red!22} & .0040 \cellcolor{green!30} & .0177 \cellcolor{red!35} & \textbf{.0055}$^{\phantom{\ddag}}$ \cellcolor{green!35} & .0079 \cellcolor{green!21} & .0056 \cellcolor{green!34} & .0956 \cellcolor{red!35} & .0088$^{\phantom{\ddag}}$ \cellcolor{green!34} & .0112$^{\phantom{\ddag}}$ \cellcolor{green!32} & \textbf{.0083}$^{\phantom{\ddag}}$ \cellcolor{green!35} \\
\texttt{RCV1-v2} & .0109 \cellcolor{red!35} & .0089 \cellcolor{green!30} & .0090$^{\dag}$ \cellcolor{green!26} & \textbf{.0088} \cellcolor{green!35} & .0185 \cellcolor{red!35} & \textbf{.0110}$^{\phantom{\ddag}}$ \cellcolor{green!35} & .0151 \cellcolor{red!3} & .0113 \cellcolor{green!32} & .0340 \cellcolor{red!35} & \textbf{.0173}$^{\phantom{\ddag}}$ \cellcolor{green!35} & .0261$^{\phantom{\ddag}}$ \cellcolor{red!2} & .0178$^{\phantom{\ddag}}$ \cellcolor{green!32} \\
\texttt{Mediamill} & .2040 \cellcolor{red!35} & .0237 \cellcolor{green!31} & .0151$^{\phantom{\ddag}}$ \cellcolor{green!34} & \textbf{.0145} \cellcolor{green!35} & .2204 \cellcolor{red!35} & .0444$^{\phantom{\ddag}}$ \cellcolor{green!27} & .0238 \cellcolor{green!34} & \textbf{.0223} \cellcolor{green!35} & .2481 \cellcolor{red!35} & .0695$^{\phantom{\ddag}}$ \cellcolor{green!21} & .0308$^{\phantom{\ddag}}$ \cellcolor{green!34} & \textbf{.0282}$^{\phantom{\ddag}}$ \cellcolor{green!35} \\
\texttt{Bibtex} & .0819 \cellcolor{red!35} & .0103 \cellcolor{green!34} & \textbf{.0100}$^{\phantom{\ddag}}$ \cellcolor{green!35} & .0101 \cellcolor{green!34} & .0919 \cellcolor{red!35} & .0116$^{\dag}$ \cellcolor{green!34} & .0137 \cellcolor{green!33} & \textbf{.0116} \cellcolor{green!35} & .1084 \cellcolor{red!35} & .0128$^{\ddag}$ \cellcolor{green!34} & .0183$^{\phantom{\ddag}}$ \cellcolor{green!30} & \textbf{.0127}$^{\phantom{\ddag}}$ \cellcolor{green!35} \\
\texttt{Corel5k} & .1043 \cellcolor{red!35} & \textbf{.0098} \cellcolor{green!35} & .0140$^{\phantom{\ddag}}$ \cellcolor{green!31} & .0178 \cellcolor{green!29} & .1041 \cellcolor{red!35} & \textbf{.0101}$^{\phantom{\ddag}}$ \cellcolor{green!35} & .0145 \cellcolor{green!31} & .0177 \cellcolor{green!29} & .1043 \cellcolor{red!35} & \textbf{.0099}$^{\phantom{\ddag}}$ \cellcolor{green!35} & .0155$^{\phantom{\ddag}}$ \cellcolor{green!30} & .0182$^{\phantom{\ddag}}$ \cellcolor{green!28} \\
\texttt{Delicious} & .1406 \cellcolor{red!35} & .0137 \cellcolor{green!32} & \textbf{.0095}$^{\phantom{\ddag}}$ \cellcolor{green!35} & .0100 \cellcolor{green!34} & .1511 \cellcolor{red!35} & .0155$^{\phantom{\ddag}}$ \cellcolor{green!32} & \textbf{.0110} \cellcolor{green!35} & .0114 \cellcolor{green!34} & .1345 \cellcolor{red!35} & .0155$^{\phantom{\ddag}}$ \cellcolor{green!32} & \textbf{.0106}$^{\phantom{\ddag}}$ \cellcolor{green!35} & .0108$^{\ddag}$ \cellcolor{green!34} \\
\hline
\multicolumn{1}{|c|}{Average} & .0842 \cellcolor{red!35} & .0219 \cellcolor{green!31} & .0189$^{\phantom{\ddag}}$ \cellcolor{green!34} & \textbf{.0182} \cellcolor{green!35} & .0862 \cellcolor{red!35} & .0346$^{\phantom{\ddag}}$ \cellcolor{green!27} & .0296 \cellcolor{green!33} & \textbf{.0285} \cellcolor{green!35} & .0957 \cellcolor{red!35} & .0562$^{\phantom{\ddag}}$ \cellcolor{green!20} & .0466$^{\phantom{\ddag}}$ \cellcolor{green!33} & \textbf{.0459}$^{\phantom{\ddag}}$ \cellcolor{green!35} \\
\multicolumn{1}{|c|}{Rank Average} &  3.8 \cellcolor{red!35} &  2.5 \cellcolor{green!8} &  2.1 \cellcolor{green!21} & \textbf{1.7} \cellcolor{green!35} &  3.7 \cellcolor{red!35} &  2.3 \cellcolor{green!17} &  2.3 \cellcolor{green!17} & \textbf{1.8} \cellcolor{green!35} &  3.7 \cellcolor{red!35} &  2.3 \cellcolor{green!11} &  2.3 \cellcolor{green!11} & \textbf{1.7} \cellcolor{green!35} \\

                \hline
                \end{tabular}%
                }
            
\end{table}

In the following sections, we turn to explore other instances of
methods of the four groups in Figure~\ref{fig:schema} beyond the ones
we choose in Table~\ref{tab:representative}.




\subsection{Testing Additional Instances of \bqmc}
\label{sec:exp:mlc}

\noindent In this section we explore other methods relying on
different multi-label classifiers, with the aim of studying the extent
to which the results we have obtained in the previous experiments
depend on the choice of the classifier being employed. To this aim, we
focus on the \bqmc\ group of methods, so that the aggregation stage
plays only a minimal role. As a quantification method we adopt PCC,
since this is the base quantifier that has yielded the best
performance overall in the experiments of
Section~\ref{sec:exp:main}. The methods we study here thus
consist of genuinely multi-label classifiers that generate
posterior probabilities, where the latter are then aggregated by computing the
expected value for each class.
The aim of this experiment is not to provide an exhaustive evaluation
of existing multi-label classifiers, but rather to study other \bqmc\
configurations in action, and hopefully pinpoint interesting
performance trends. 

With this in mind, we choose some representative
instances from the main families of multi-label classifiers discussed
in Section~\ref{sec:relwork}.
%
The multi-label classifiers we consider here include (a) multi-label
versions of KNN
(\textsc{ML-knn}),\footnote{\url{http://scikit.ml/api/skmultilearn.adapt.mlknn.html}} 
decision trees
(\textsc{DT}),\footnote{\url{https://scikit-learn.org/stable/modules/generated/sklearn.tree.DecisionTreeClassifier.html}} 
and random forests
(\textsc{RF})\footnote{\url{https://scikit-learn.org/stable/modules/generated/sklearn.ensemble.RandomForestClassifier.html\#sklearn.ensemble.RandomForestClassifier.fit}} 
as representatives of the family of ``algorithm adaptation'' methods
(Section~\ref{sec:relwork:algadapt}); (b) classifier chains
(\textsc{CChains})\footnote{\url{https://scikit-learn.org/stable/modules/generated/sklearn.multioutput.ClassifierChain.html\#sklearn.multioutput.ClassifierChain}}
as representative of the family of ``problem transformation'' methods
(Section~\ref{sec:relwork:probtrans}); 
and (c)
\textsc{CLEMS}\footnote{\url{http://scikit.ml/api/skmultilearn.embedding.clems.html}} 
and label space clustering (\textsc{LSC})\footnote{This was
implemented by combining different classes from
\textsc{scikit-multilearn}.} 
as representatives of the family of ``ensemble'' methods
(Section~\ref{sec:relwork:ensembles}). For the sake of comparison, we
also include stacked generalization (\textsc{SG} -- another ensemble
method and the multi-label classifier we choose for the experiments of
Section~\ref{sec:exp:main}), and \bqbc\ (with PCC as the binary
quantifier and LR as the binary classifier) acting as a lower bound
baseline. Values for SG and \bqbc\ are taken from
Table~\ref{tab:pcc_general}.
We carry out model selection by optimizing the hyperparameters listed
succinctly in Table~\ref{tab:modsel:mlc}. The results we have obtained
are presented in Table~\ref{tab:mlc}.

\begin{table}[ht]
  \centering
  \caption{Hyperparameters we explore during model selection for
  different multi-label classifiers.}
  \label{tab:modsel:mlc}
  \begin{tabular}{llll}
\toprule
\multicolumn{1}{c}{Classifier} & Hyperparameter & Description & Values \\ \midrule
\multirow{2}{*}{\textsc{ML-knn}\phantom{0}}  & $k$           & number of neighbours                  & \{1, 3, 5, 7, 9\}      \\
         & $s$           & smoothing factor                & \{0.5, 0.7, 1\}       \\\midrule
DT                       & \emph{Criterion}     & split function                  & \{Gini, Entropy\}  \\\midrule
RF                       & \emph{N\_estimators} & number of trees                       & \{10, 100, 200\}     \\
\midrule
\multirow{2}{*}{\textsc{CChains}} & $C$           & inverse regularization strength & $\{10^{-1}, \ldots, 10^{2}, 10^{3}\}$ \\
         & \emph{ClassWeight} & weights associated with classes & \{None, Balanced\} \\
\midrule
\multirow{3}{*}{\textsc{CLEMS}}   & \emph{N\_estimators} & number of trees                       & \{10, 20, 50\}     \\
         & $k$           & number of neighbours                  & \{1, 3, 5, 7, 9\}  \\
         & $s$           & smoothing factor                & \{0.5, 0.7, 1\}       \\\midrule
 \multirow{3}{*}{\textsc{LSC}}     & \emph{N\_clusters}   & number of clusters                    & \{2, 3, 5, 10, 50\}    \\
         & $k$           & number of neighbours                  & \{1, 3, 5, 7, 9\}  \\
         & $s$           & smoothing factor                & \{0.5, 0.7, 1\}       \\\midrule
 \multirow{3}{*}{\textsc{SG}}      & \emph{Norm} & center and scale features & \{True, False\} \\
         & $C$           & inverse regularization strength & $\{10^{-1}, \ldots, 10^{2}, 10^{3}\}$ \\
         & \emph{ClassWeight} & weights associated with classes & \{None, Balanced\} \\
\bottomrule
\end{tabular}
\end{table}

\begin{table}[tb]
  \centering
  \caption{Values of \abse\ obtained for different multi-label
  classifiers using PCC as the base quantifier in \bqmc. (All results
  are reported with only three digits after the decimal point, unlike
  in other tables, in order to maximize readability.)}
  \label{tab:mlc}
  
                \resizebox{\textwidth}{!}{%
            \setlength\tabcolsep{1.5pt}
            
                    \begin{tabular}{|l|c|ccccccc|c|ccccccc|c|ccccccc|} \cline{2-25}
\multicolumn{1}{c|}{} & \multicolumn{8}{c|}{low shift} & \multicolumn{8}{c|}{mid shift} & \multicolumn{8}{c|}{high shift}\\\cline{2-25}

\multicolumn{1}{c|}{} & \side{\bqbc} & \side{\textsc{ML-knn}\phantom{0}} & \side{\textsc{DT}} & \side{\textsc{RF}} & \side{\textsc{CChain}} & \side{\textsc{CLEMS}} & \side{\textsc{LSC}} & \side{\textsc{SG}} & \side{\bqbc} & \side{\textsc{ML-knn}\phantom{0}} & \side{\textsc{DT}} & \side{\textsc{RF}} & \side{\textsc{CChain}} & \side{\textsc{CLEMS}} & \side{\textsc{LSC}} & \side{\textsc{SG}} & \side{\bqbc} & \side{\textsc{ML-knn}\phantom{0}} & \side{\textsc{DT}} & \side{\textsc{RF}} & \side{\textsc{CChain}} & \side{\textsc{CLEMS}} & \side{\textsc{LSC}} & \side{\textsc{SG}}\\\hline
\texttt{Emotions} & .042 \cellcolor{green!34} & .223 \cellcolor{red!19} & .065 \cellcolor{green!27} & \textbf{.041}$^{\phantom{\ddag}}$ \cellcolor{green!35} & .050 \cellcolor{green!32} & .223 \cellcolor{red!19} & .274 \cellcolor{red!35} & .042 \cellcolor{green!34} & .069 \cellcolor{green!33} & .232 \cellcolor{red!19} & .088 \cellcolor{green!27} & .071 \cellcolor{green!33} & .071 \cellcolor{green!33} & .232 \cellcolor{red!19} & .279 \cellcolor{red!35} & \textbf{.065} \cellcolor{green!35} & .092 \cellcolor{green!32} & .250 \cellcolor{red!18} & .102$^{\phantom{\ddag}}$ \cellcolor{green!29} & .092 \cellcolor{green!32} & .096$^{\phantom{\ddag}}$ \cellcolor{green!31} & .250 \cellcolor{red!18} & .301 \cellcolor{red!35} & \textbf{.086} \cellcolor{green!35} \\
\texttt{Scene} & .084 \cellcolor{green!9} & .150 \cellcolor{red!25} & .042 \cellcolor{green!32} & .036$^{\ddag}$ \cellcolor{green!34} & .037 \cellcolor{green!34} & .150 \cellcolor{red!25} & .169 \cellcolor{red!35} & \textbf{.036} \cellcolor{green!35} & .114 \cellcolor{green!6} & .157 \cellcolor{red!22} & .084 \cellcolor{green!26} & .087 \cellcolor{green!24} & \textbf{.071} \cellcolor{green!35} & .157 \cellcolor{red!22} & .175 \cellcolor{red!35} & .083 \cellcolor{green!27} & .143 \cellcolor{red!5} & .160 \cellcolor{red!21} & .114$^{\phantom{\ddag}}$ \cellcolor{green!20} & .127 \cellcolor{green!9} & \textbf{.099}$^{\phantom{\ddag}}$ \cellcolor{green!35} & .160 \cellcolor{red!21} & .175 \cellcolor{red!35} & .122 \cellcolor{green!13} \\
\texttt{Yeast} & .176 \cellcolor{red!10} & .166 \cellcolor{red!7} & .055 \cellcolor{green!31} & .045$^{\phantom{\ddag}}$ \cellcolor{green!34} & .056 \cellcolor{green!30} & .161 \cellcolor{red!5} & .244 \cellcolor{red!35} & \textbf{.044} \cellcolor{green!35} & .190 \cellcolor{red!4} & .192 \cellcolor{red!5} & .106 \cellcolor{green!32} & .101 \cellcolor{green!34} & .103 \cellcolor{green!33} & .189 \cellcolor{red!4} & .259 \cellcolor{red!35} & \textbf{.099} \cellcolor{green!35} & .221 \cellcolor{red!1} & .222 \cellcolor{red!1} & .153$^{\phantom{\ddag}}$ \cellcolor{green!32} & .155 \cellcolor{green!30} & \textbf{.147}$^{\phantom{\ddag}}$ \cellcolor{green!35} & .220 \cellcolor{red!0} & .290 \cellcolor{red!35} & .155 \cellcolor{green!30} \\
\texttt{Birds} & .029 \cellcolor{green!7} & .048 \cellcolor{red!33} & .024 \cellcolor{green!18} & \textbf{.016}$^{\phantom{\ddag}}$ \cellcolor{green!35} & .019 \cellcolor{green!26} & .048 \cellcolor{red!33} & .049 \cellcolor{red!35} & .021 \cellcolor{green!24} & .033 \cellcolor{green!8} & .052 \cellcolor{red!33} & .026 \cellcolor{green!22} & \textbf{.021} \cellcolor{green!35} & .025 \cellcolor{green!26} & .052 \cellcolor{red!33} & .053 \cellcolor{red!35} & .025 \cellcolor{green!26} & .044 \cellcolor{green!5} & .063 \cellcolor{red!32} & .034$^{\phantom{\ddag}}$ \cellcolor{green!24} & \textbf{.029} \cellcolor{green!35} & .035$^{\phantom{\ddag}}$ \cellcolor{green!23} & .063 \cellcolor{red!32} & .065 \cellcolor{red!35} & .032 \cellcolor{green!28} \\
\texttt{Genbase} & .001 \cellcolor{green!34} & .057 \cellcolor{red!33} & .001 \cellcolor{green!34} & .003$^{\phantom{\ddag}}$ \cellcolor{green!31} & .001 \cellcolor{green!34} & .057 \cellcolor{red!33} & .058 \cellcolor{red!35} & \textbf{.001} \cellcolor{green!35} & .001 \cellcolor{green!34} & .057 \cellcolor{red!33} & .001 \cellcolor{green!34} & .004 \cellcolor{green!31} & .001 \cellcolor{green!34} & .057 \cellcolor{red!33} & .059 \cellcolor{red!35} & \textbf{.001} \cellcolor{green!35} & .001 \cellcolor{green!34} & .057 \cellcolor{red!33} & .001$^{\phantom{\ddag}}$ \cellcolor{green!34} & .003 \cellcolor{green!31} & .001$^{\phantom{\ddag}}$ \cellcolor{green!34} & .057 \cellcolor{red!33} & .059 \cellcolor{red!35} & \textbf{.001} \cellcolor{green!35} \\
\texttt{Medical} & \textbf{.013} \cellcolor{green!35} & .054 \cellcolor{red!27} & .014 \cellcolor{green!33} & .016$^{\phantom{\ddag}}$ \cellcolor{green!29} & .013 \cellcolor{green!34} & .054 \cellcolor{red!27} & .059 \cellcolor{red!35} & .014 \cellcolor{green!33} & \textbf{.015} \cellcolor{green!35} & .057 \cellcolor{red!28} & .015 \cellcolor{green!34} & .023 \cellcolor{green!22} & .015 \cellcolor{green!34} & .057 \cellcolor{red!27} & .062 \cellcolor{red!35} & .016 \cellcolor{green!33} & .017 \cellcolor{green!32} & .063 \cellcolor{red!27} & \textbf{.015}$^{\phantom{\ddag}}$ \cellcolor{green!35} & .031 \cellcolor{green!13} & .017$^{\phantom{\ddag}}$ \cellcolor{green!31} & .062 \cellcolor{red!26} & .068 \cellcolor{red!35} & .018 \cellcolor{green!30} \\
\texttt{tmc2007\_500} & .111 \cellcolor{red!35} & .078 \cellcolor{red!13} & \textbf{.001} \cellcolor{green!35} & .009$^{\phantom{\ddag}}$ \cellcolor{green!30} & .020 \cellcolor{green!23} & .078 \cellcolor{red!13} & .095 \cellcolor{red!24} & .019 \cellcolor{green!23} & .115 \cellcolor{red!35} & .095 \cellcolor{red!22} & \textbf{.002} \cellcolor{green!35} & .017 \cellcolor{green!25} & .033 \cellcolor{green!15} & .095 \cellcolor{red!22} & .112 \cellcolor{red!33} & .033 \cellcolor{green!15} & .101 \cellcolor{red!24} & .103 \cellcolor{red!25} & \textbf{.002}$^{\phantom{\ddag}}$ \cellcolor{green!35} & .023 \cellcolor{green!22} & .039$^{\phantom{\ddag}}$ \cellcolor{green!12} & .103 \cellcolor{red!25} & .119 \cellcolor{red!35} & .040 \cellcolor{green!12} \\
\texttt{Ohsumed} & .100 \cellcolor{red!35} & .067 \cellcolor{red!6} & .024 \cellcolor{green!29} & .021$^{\phantom{\ddag}}$ \cellcolor{green!32} & \textbf{.017} \cellcolor{green!35} & .067 \cellcolor{red!6} & .071 \cellcolor{red!10} & .018 \cellcolor{green!34} & .109 \cellcolor{red!35} & .077 \cellcolor{red!8} & .037 \cellcolor{green!25} & .036 \cellcolor{green!25} & \textbf{.025} \cellcolor{green!35} & .077 \cellcolor{red!8} & .082 \cellcolor{red!12} & .026 \cellcolor{green!33} & .118 \cellcolor{red!35} & .083 \cellcolor{red!7} & .046$^{\phantom{\ddag}}$ \cellcolor{green!22} & .049 \cellcolor{green!19} & \textbf{.029}$^{\phantom{\ddag}}$ \cellcolor{green!35} & .083 \cellcolor{red!7} & .087 \cellcolor{red!10} & .032 \cellcolor{green!32} \\
\texttt{Enron} & .035 \cellcolor{green!8} & .057 \cellcolor{red!22} & .021 \cellcolor{green!27} & .017$^{\phantom{\ddag}}$ \cellcolor{green!33} & .016 \cellcolor{green!34} & .058 \cellcolor{red!23} & .066 \cellcolor{red!35} & \textbf{.016} \cellcolor{green!35} & .040 \cellcolor{green!10} & .061 \cellcolor{red!22} & .027 \cellcolor{green!29} & .025 \cellcolor{green!33} & .024 \cellcolor{green!34} & .061 \cellcolor{red!22} & .069 \cellcolor{red!35} & \textbf{.023} \cellcolor{green!35} & .044 \cellcolor{green!8} & .060 \cellcolor{red!19} & .031$^{\phantom{\ddag}}$ \cellcolor{green!30} & .031 \cellcolor{green!29} & \textbf{.029}$^{\phantom{\ddag}}$ \cellcolor{green!35} & .060 \cellcolor{red!20} & .069 \cellcolor{red!35} & .029 \cellcolor{green!34} \\
\texttt{Reuters-21578} & .017 \cellcolor{red!35} & .015 \cellcolor{red!28} & .006 \cellcolor{green!24} & .007$^{\phantom{\ddag}}$ \cellcolor{green!18} & .004 \cellcolor{green!34} & .016 \cellcolor{red!28} & .016 \cellcolor{red!32} & \textbf{.004} \cellcolor{green!35} & .024 \cellcolor{red!35} & .024 \cellcolor{red!32} & .009 \cellcolor{green!22} & .013 \cellcolor{green!6} & \textbf{.006} \cellcolor{green!35} & .024 \cellcolor{red!32} & .024 \cellcolor{red!34} & .006 \cellcolor{green!34} & .037 \cellcolor{red!30} & .039 \cellcolor{red!33} & .012$^{\phantom{\ddag}}$ \cellcolor{green!23} & .019 \cellcolor{green!10} & .007$^{\ddag}$ \cellcolor{green!34} & .039 \cellcolor{red!33} & .039 \cellcolor{red!35} & \textbf{.007} \cellcolor{green!35} \\
\texttt{RCV1-v2} & .046 \cellcolor{red!35} & .030 \cellcolor{red!5} & .013 \cellcolor{green!25} & .014$^{\phantom{\ddag}}$ \cellcolor{green!25} & .009 \cellcolor{green!33} & .030 \cellcolor{red!5} & .033 \cellcolor{red!10} & \textbf{.008} \cellcolor{green!35} & .053 \cellcolor{red!35} & .036 \cellcolor{red!5} & .022 \cellcolor{green!18} & .027 \cellcolor{green!10} & .014 \cellcolor{green!33} & .036 \cellcolor{red!5} & .039 \cellcolor{red!9} & \textbf{.013} \cellcolor{green!35} & .065 \cellcolor{red!35} & .053 \cellcolor{red!16} & .036$^{\phantom{\ddag}}$ \cellcolor{green!10} & .046 \cellcolor{red!5} & .021$^{\phantom{\ddag}}$ \cellcolor{green!33} & .053 \cellcolor{red!16} & .056 \cellcolor{red!21} & \textbf{.020} \cellcolor{green!35} \\
\texttt{Mediamill} & .170 \cellcolor{red!35} & .032 \cellcolor{green!27} & .018 \cellcolor{green!33} & .016$^{\phantom{\ddag}}$ \cellcolor{green!34} & .019 \cellcolor{green!33} & .035 \cellcolor{green!26} & .039 \cellcolor{green!24} & \textbf{.015} \cellcolor{green!35} & .174 \cellcolor{red!35} & .044 \cellcolor{green!27} & .031 \cellcolor{green!33} & .030 \cellcolor{green!34} & .033 \cellcolor{green!32} & .046 \cellcolor{green!26} & .051 \cellcolor{green!24} & \textbf{.029} \cellcolor{green!35} & .181 \cellcolor{red!35} & .058 \cellcolor{green!26} & .043$^{\phantom{\ddag}}$ \cellcolor{green!33} & .043 \cellcolor{green!33} & .047$^{\phantom{\ddag}}$ \cellcolor{green!31} & .060 \cellcolor{green!25} & .065 \cellcolor{green!22} & \textbf{.040} \cellcolor{green!35} \\
\texttt{Bibtex} & .035 \cellcolor{red!35} & .016 \cellcolor{green!16} & .011 \cellcolor{green!30} & .009$^{\phantom{\ddag}}$ \cellcolor{green!34} & .010 \cellcolor{green!33} & .016 \cellcolor{green!16} & .016 \cellcolor{green!16} & \textbf{.009} \cellcolor{green!35} & .037 \cellcolor{red!35} & .021 \cellcolor{green!10} & .013 \cellcolor{green!29} & .013 \cellcolor{green!30} & \textbf{.011} \cellcolor{green!35} & .021 \cellcolor{green!10} & .021 \cellcolor{green!10} & .013 \cellcolor{green!31} & .042 \cellcolor{red!35} & .025 \cellcolor{green!5} & .014$^{\phantom{\ddag}}$ \cellcolor{green!30} & .015 \cellcolor{green!28} & \textbf{.012}$^{\phantom{\ddag}}$ \cellcolor{green!35} & .025 \cellcolor{green!5} & .025 \cellcolor{green!5} & .015 \cellcolor{green!28} \\
\texttt{Corel5k} & .058 \cellcolor{red!35} & .011 \cellcolor{green!30} & .013 \cellcolor{green!27} & .008$^{\phantom{\ddag}}$ \cellcolor{green!34} & .009 \cellcolor{green!33} & .011 \cellcolor{green!30} & .011 \cellcolor{green!30} & \textbf{.008} \cellcolor{green!35} & .059 \cellcolor{red!35} & .011 \cellcolor{green!31} & .013 \cellcolor{green!27} & .009 \cellcolor{green!34} & .009 \cellcolor{green!33} & .011 \cellcolor{green!30} & .011 \cellcolor{green!31} & \textbf{.008} \cellcolor{green!35} & .059 \cellcolor{red!35} & .011 \cellcolor{green!31} & .013$^{\phantom{\ddag}}$ \cellcolor{green!29} & .010 \cellcolor{green!33} & .010$^{\phantom{\ddag}}$ \cellcolor{green!33} & .012 \cellcolor{green!31} & .011 \cellcolor{green!31} & \textbf{.009} \cellcolor{green!35} \\
\texttt{Delicious} & .142 \cellcolor{red!35} & .016 \cellcolor{green!31} & .010 \cellcolor{green!34} & .009$^{\phantom{\ddag}}$ \cellcolor{green!34} & .011 \cellcolor{green!33} & .018 \cellcolor{green!30} & .017 \cellcolor{green!30} & \textbf{.009} \cellcolor{green!35} & .142 \cellcolor{red!35} & .017 \cellcolor{green!31} & .013 \cellcolor{green!34} & .012 \cellcolor{green!34} & .013 \cellcolor{green!34} & .019 \cellcolor{green!31} & .019 \cellcolor{green!31} & \textbf{.012} \cellcolor{green!35} & .124 \cellcolor{red!35} & .018 \cellcolor{green!31} & .012$^{\ddag}$ \cellcolor{green!34} & .013 \cellcolor{green!34} & .013$^{\phantom{\ddag}}$ \cellcolor{green!34} & .018 \cellcolor{green!31} & .018 \cellcolor{green!30} & \textbf{.012} \cellcolor{green!35} \\
\hline
\multicolumn{1}{|c|}{Average} & .068 \cellcolor{red!31} & .059 \cellcolor{red!20} & .019 \cellcolor{green!30} & .016 \cellcolor{green!34} & .017 \cellcolor{green!32} & .060 \cellcolor{red!21} & .070 \cellcolor{red!35} & \textbf{.016} \cellcolor{green!35} & .076 \cellcolor{red!15} & .081 \cellcolor{red!20} & .034 \cellcolor{green!31} & .034 \cellcolor{green!31} & .031 \cellcolor{green!34} & .081 \cellcolor{red!21} & .094 \cellcolor{red!35} & \textbf{.031} \cellcolor{green!35} & .101 \cellcolor{red!6} & .115 \cellcolor{red!19} & .059$^{\phantom{\ddag}}$ \cellcolor{green!32} & .065 \cellcolor{green!26} & \textbf{.056}$^{\phantom{\ddag}}$ \cellcolor{green!35} & .115 \cellcolor{red!19} & .132 \cellcolor{red!35} & .060 \cellcolor{green!31} \\
\multicolumn{1}{|c|}{Rank Average} &  6.2 \cellcolor{red!22} &  5.5 \cellcolor{red!14} &  3.7 \cellcolor{green!9} &  2.7 \cellcolor{green!21} &  2.8 \cellcolor{green!20} &  6.3 \cellcolor{red!24} &  7.2 \cellcolor{red!35} & \textbf{1.6} \cellcolor{green!35} &  6.1 \cellcolor{red!20} &  5.5 \cellcolor{red!13} &  3.7 \cellcolor{green!10} &  3.1 \cellcolor{green!17} &  2.3 \cellcolor{green!27} &  6.4 \cellcolor{red!24} &  7.2 \cellcolor{red!35} & \textbf{1.7} \cellcolor{green!35} &  5.8 \cellcolor{red!13} &  6.0 \cellcolor{red!16} &  2.9 \cellcolor{green!23} &  3.3 \cellcolor{green!19} &  2.4 \cellcolor{green!30} &  6.1 \cellcolor{red!18} &  7.4 \cellcolor{red!35} & \textbf{2.1} \cellcolor{green!35} \\

                \hline
                \end{tabular}%
                }
            
\end{table}

These results reveal that, despite the fact that \textsc{SG} is the
best-performing method, other multi-label classifiers work comparably
well and could be used to yield multi-label aggregative quantifiers
with similar performance levels. In particular, \textsc{CChains} tends
to fare very well in all cases, followed by \textsc{RF} and
\textsc{DT}; these results are, by and large, consistent with those
reported in~\cite{Madjarov:2012yl}. The methods \textsc{ML-knn},
\textsc{CLEMS}, and \textsc{LSC}, however, prove inferior, sometimes
performing even worse than the \bqbc\ baseline.

Although the results we report in Table~\ref{tab:mlc} are obtained on
the test set, we confirm that they are strongly correlated with the
performance levels we measured on the held-out validation set. Indeed,
we chose SG as our multi-label classifier for the experiments of
Section~\ref{sec:exp:main} since this was the model yielding the
lowest \abse\ during model selection.





\subsection{Testing Additional Instances of \mqbc}
\label{sec:exp:mlq}


\noindent In this section we compare the different multi-label
aggregation strategies proposed in Section~\ref{sec:quant:ml:reg} and
\ref{sec:quant:ml:lp}. In order to do so, we focus on the \mqbc\ group
of methods (i.e., those relying on binary classifiers for the label
predictions) so that all the label dependencies are modelled
exclusively at the aggregation stage.

For the label powerset -based strategy (LPQ)  we consider two different ways
for generating the clusters, after which \SLQ\ is applied to the
resulting label powersets of each cluster. In particular, we
investigate:
\begin{itemize}
\item \textsc{RakEL-LPQ}: inspired by
  \textsc{RakEL}~\cite{Tsoumakas:2011vp}; generates $k$ disjoint
  random clusters;
\item \textsc{kMeans-LPQ}: inspired by LSC~\cite{Szymanski:2016qj};
  generates clusters via $k$-means.
\end{itemize}
\noindent For the regression-based  (RQ) strategy we consider two alternative
regressors (other results exploring further regression algorithms can
be found in Appendix \ref{sec:otherregressors}):
\begin{itemize}
\item \textsc{Ridge-RQ}: using ridge
  regression;\footnote{\url{https://scikit-learn.org/stable/modules/generated/sklearn.linear_model.Ridge.html}}
\item \textsc{RF-RQ}: using random-forest
  regression.\footnote{\url{https://scikit-learn.org/stable/modules/generated/sklearn.ensemble.RandomForestRegressor.html}}
\end{itemize}
\noindent For the sake of comparison, we add the regression-based
strategy \textsc{SVR-RQ} (Section~\ref{sec:quant:ml:reg}), that
corresponds to our configuration of choice for \mqbc\ in
Section~\ref{sec:exp:main}, and the \bqbc\ system (PCC+LR) as a
lower-bound baseline. The results for \textsc{SVR-RQ} and \bqbc\ are
taken from Table~\ref{tab:pcc_general}.  Model selection is carried
out by exploring, via grid-search optimization, the hyperparameters
indicated in Table~\ref{tab:modsel:mlq}. The results we have obtained
are shown in Table~\ref{tab:mlq}.

\begin{table}[h!]
  \centering
  \caption{Hyperparameters explored during model selection for
  different multi-label quantifiers. All methods are deployed with an
  LR classifier; for the hyperparameters $C$ and
  \emph{ClassWeight}, we explore in the ranges
  $\{10^{-1}, \ldots, 10^{2},10^{3}\}$ and \{None, Balanced\},
  respectively.}
  \label{tab:modsel:mlq}
      \resizebox{\textwidth}{!}{%
\begin{tabular}{llll}
\toprule
Quantifier & Hyperparameter & Description & Values \\ \midrule
\multirow{1}{*}{\textsc{RakEL-LP}}  & $k$  & number of  clusters & \{2, 5, 10, 50, 100\}  \\
\multirow{1}{*}{\textsc{kMeans-LP}\phantom{0}} & $k$ & number of clusters                                             & \{5, 15, 50, 100\}    \\

\multirow{1}{*}{\textsc{Ridge-RQ}}                & \emph{Alpha}         & regularization strength               & $\{10^{-3}, \ldots, 10^{2},10^{3}\}$     \\
\multirow{1}{*}{\textsc{RF-RQ}}       & \emph{N\_estimators} & number of trees                             & \{10, 100, 200\} \\                  
\multirow{1}{*}{\textsc{SVR-RQ}}
                          & $C$             & inverse regularization strength in Linear SVR & $\{10^{-1}, \ldots, 10^{2},10^{3}\}$ \\

\bottomrule
\end{tabular}
}
\end{table}

\begin{table}[tb]
  \centering
  \caption{\abse\ for different multi-label aggregation methods in
  \mqbc.}
  \label{tab:mlq}
  
                \resizebox{\textwidth}{!}{%
            \setlength\tabcolsep{1.5pt}
            
                    \begin{tabular}{|l|c|ccccc|c|ccccc|c|ccccc|} \cline{2-19}
\multicolumn{1}{c|}{} & \multicolumn{6}{c|}{low shift} & \multicolumn{6}{c|}{mid shift} & \multicolumn{6}{c|}{high shift}\\\cline{2-19}

\multicolumn{1}{c|}{} & \side{\bqbc} & \side{\textsc{RakEL-LPQ}} & \side{\textsc{kMeans-LPQ}\phantom{0}} & \side{\textsc{Ridge-RQ}} & \side{\textsc{RF-RQ}} & \side{\textsc{SVR-RQ}} & \side{\bqbc} & \side{\textsc{RakEL-LPQ}} & \side{\textsc{kMeans-LPQ}\phantom{0}} & \side{\textsc{Ridge-RQ}} & \side{\textsc{RF-RQ}} & \side{\textsc{SVR-RQ}} & \side{\bqbc} & \side{\textsc{RakEL-LPQ}} & \side{\textsc{kMeans-LPQ}\phantom{0}} & \side{\textsc{Ridge-RQ}} & \side{\textsc{RF-RQ}} & \side{\textsc{SVR-RQ}}\\\hline
\texttt{Emotions} & .0418$^{\ddag}$ \cellcolor{green!34} & .0451$^{\phantom{\ddag}}$ \cellcolor{green!20} & \textbf{.0416} \cellcolor{green!35} & .0500 \cellcolor{green!0} & .0484 \cellcolor{green!7} & .0586 \cellcolor{red!35} & .0685 \cellcolor{red!8} & .0687 \cellcolor{red!10} & .0657 \cellcolor{green!12} & \textbf{.0626} \cellcolor{green!35} & .0707 \cellcolor{red!25} & .0720 \cellcolor{red!35} & .0923 \cellcolor{green!13} & .0940$^{\phantom{\ddag}}$ \cellcolor{green!10} & .0869 \cellcolor{green!21} & \textbf{.0782} \cellcolor{green!35} & .1235$^{\phantom{\ddag}}$ \cellcolor{red!35} & .0951 \cellcolor{green!8} \\
\texttt{Scene} & .0842$^{\phantom{\ddag}}$ \cellcolor{red!35} & .0333$^{\phantom{\ddag}}$ \cellcolor{green!31} & .0345 \cellcolor{green!29} & .0350 \cellcolor{green!28} & \textbf{.0303} \cellcolor{green!35} & .0384 \cellcolor{green!24} & .1140 \cellcolor{red!35} & .0680 \cellcolor{green!3} & .0735 \cellcolor{red!1} & .0454 \cellcolor{green!22} & \textbf{.0306} \cellcolor{green!35} & .0459 \cellcolor{green!22} & .1427 \cellcolor{red!35} & .0961$^{\phantom{\ddag}}$ \cellcolor{green!0} & .1063 \cellcolor{red!7} & .0581 \cellcolor{green!28} & .0507$^{\ddag}$ \cellcolor{green!34} & \textbf{.0499} \cellcolor{green!35} \\
\texttt{Yeast} & .1756$^{\phantom{\ddag}}$ \cellcolor{red!35} & .0443$^{\ddag}$ \cellcolor{green!34} & \textbf{.0443} \cellcolor{green!35} & .0477 \cellcolor{green!33} & .0561 \cellcolor{green!28} & .0480 \cellcolor{green!33} & .1900 \cellcolor{red!35} & .0972 \cellcolor{green!30} & .0970 \cellcolor{green!30} & .0920 \cellcolor{green!34} & .0956 \cellcolor{green!31} & \textbf{.0913} \cellcolor{green!35} & .2206 \cellcolor{red!35} & .1520$^{\phantom{\ddag}}$ \cellcolor{green!21} & .1521 \cellcolor{green!21} & .1374 \cellcolor{green!33} & .1517$^{\phantom{\ddag}}$ \cellcolor{green!22} & \textbf{.1361} \cellcolor{green!35} \\
\texttt{Birds} & .0286$^{\phantom{\ddag}}$ \cellcolor{red!35} & .0242$^{\phantom{\ddag}}$ \cellcolor{red!5} & .0194 \cellcolor{green!26} & .0213 \cellcolor{green!13} & .0185 \cellcolor{green!32} & \textbf{.0181} \cellcolor{green!35} & .0328 \cellcolor{red!35} & .0286 \cellcolor{red!2} & \textbf{.0237} \cellcolor{green!35} & .0289 \cellcolor{red!5} & .0251 \cellcolor{green!23} & .0243 \cellcolor{green!30} & .0440 \cellcolor{red!35} & .0405$^{\phantom{\ddag}}$ \cellcolor{red!15} & \textbf{.0317} \cellcolor{green!35} & .0405 \cellcolor{red!15} & .0382$^{\phantom{\ddag}}$ \cellcolor{red!2} & .0351 \cellcolor{green!15} \\
\texttt{Genbase} & .0011$^{\phantom{\ddag}}$ \cellcolor{green!30} & \textbf{.0005}$^{\phantom{\ddag}}$ \cellcolor{green!35} & .0014 \cellcolor{green!28} & .0015 \cellcolor{green!27} & .0103 \cellcolor{red!35} & .0022 \cellcolor{green!22} & .0011 \cellcolor{green!32} & \textbf{.0005} \cellcolor{green!35} & .0015 \cellcolor{green!30} & .0014 \cellcolor{green!30} & .0148 \cellcolor{red!35} & .0025 \cellcolor{green!25} & .0010 \cellcolor{green!32} & \textbf{.0004}$^{\phantom{\ddag}}$ \cellcolor{green!35} & .0013 \cellcolor{green!31} & .0013 \cellcolor{green!31} & .0172$^{\phantom{\ddag}}$ \cellcolor{red!35} & .0023 \cellcolor{green!27} \\
\texttt{Medical} & .0127$^{\phantom{\ddag}}$ \cellcolor{green!31} & \textbf{.0122}$^{\phantom{\ddag}}$ \cellcolor{green!35} & .0130 \cellcolor{green!29} & .0230 \cellcolor{red!35} & .0202 \cellcolor{red!17} & .0191 \cellcolor{red!9} & .0146 \cellcolor{green!33} & .0149 \cellcolor{green!33} & \textbf{.0142} \cellcolor{green!35} & .0306 \cellcolor{red!10} & .0393 \cellcolor{red!35} & .0279 \cellcolor{red!3} & .0169 \cellcolor{green!33} & .0181$^{\phantom{\ddag}}$ \cellcolor{green!31} & \textbf{.0158} \cellcolor{green!35} & .0365 \cellcolor{green!5} & .0651$^{\phantom{\ddag}}$ \cellcolor{red!35} & .0351 \cellcolor{green!7} \\
\texttt{tmc2007\_500} & .1108$^{\phantom{\ddag}}$ \cellcolor{red!35} & .0188$^{\phantom{\ddag}}$ \cellcolor{green!34} & \textbf{.0182} \cellcolor{green!35} & .0226 \cellcolor{green!31} & .0211 \cellcolor{green!32} & .0213 \cellcolor{green!32} & .1154 \cellcolor{red!35} & .0319 \cellcolor{green!27} & .0303 \cellcolor{green!29} & .0274 \cellcolor{green!31} & \textbf{.0224} \cellcolor{green!35} & .0292 \cellcolor{green!29} & .1008 \cellcolor{red!35} & .0385$^{\phantom{\ddag}}$ \cellcolor{green!21} & .0354 \cellcolor{green!24} & .0282 \cellcolor{green!30} & \textbf{.0237}$^{\phantom{\ddag}}$ \cellcolor{green!35} & .0321 \cellcolor{green!27} \\
\texttt{Ohsumed} & .1004$^{\phantom{\ddag}}$ \cellcolor{red!35} & \textbf{.0171}$^{\phantom{\ddag}}$ \cellcolor{green!35} & .0174 \cellcolor{green!34} & .0193 \cellcolor{green!33} & .0183 \cellcolor{green!33} & .0183 \cellcolor{green!34} & .1087 \cellcolor{red!35} & .0249 \cellcolor{green!29} & .0251 \cellcolor{green!29} & .0208 \cellcolor{green!33} & \textbf{.0184} \cellcolor{green!35} & .0209 \cellcolor{green!33} & .1177 \cellcolor{red!35} & .0300$^{\phantom{\ddag}}$ \cellcolor{green!27} & .0304 \cellcolor{green!27} & .0221 \cellcolor{green!33} & \textbf{.0201}$^{\phantom{\ddag}}$ \cellcolor{green!35} & .0215 \cellcolor{green!33} \\
\texttt{Enron} & .0347$^{\phantom{\ddag}}$ \cellcolor{red!35} & .0163$^{\phantom{\ddag}}$ \cellcolor{green!34} & \textbf{.0163} \cellcolor{green!35} & .0192 \cellcolor{green!23} & .0174 \cellcolor{green!30} & .0169 \cellcolor{green!32} & .0397 \cellcolor{red!35} & .0242 \cellcolor{green!26} & .0239 \cellcolor{green!26} & .0250 \cellcolor{green!22} & \textbf{.0219} \cellcolor{green!35} & .0227 \cellcolor{green!32} & .0439 \cellcolor{red!35} & .0301$^{\phantom{\ddag}}$ \cellcolor{green!7} & .0297 \cellcolor{green!8} & .0287 \cellcolor{green!11} & \textbf{.0212}$^{\phantom{\ddag}}$ \cellcolor{green!35} & .0253 \cellcolor{green!22} \\
\texttt{Reuters-21578} & .0167$^{\phantom{\ddag}}$ \cellcolor{red!35} & \textbf{.0037}$^{\phantom{\ddag}}$ \cellcolor{green!35} & .0037 \cellcolor{green!34} & .0048 \cellcolor{green!29} & .0073 \cellcolor{green!15} & .0049 \cellcolor{green!28} & .0243 \cellcolor{red!35} & .0062 \cellcolor{green!34} & \textbf{.0060} \cellcolor{green!35} & .0070 \cellcolor{green!31} & .0109 \cellcolor{green!16} & .0070 \cellcolor{green!31} & .0370 \cellcolor{red!35} & .0081$^{\phantom{\ddag}}$ \cellcolor{green!34} & \textbf{.0077} \cellcolor{green!35} & .0103 \cellcolor{green!28} & .0135$^{\phantom{\ddag}}$ \cellcolor{green!21} & .0088 \cellcolor{green!32} \\
\texttt{RCV1-v2} & .0456$^{\phantom{\ddag}}$ \cellcolor{red!35} & \textbf{.0085}$^{\phantom{\ddag}}$ \cellcolor{green!35} & .0089 \cellcolor{green!34} & .0095 \cellcolor{green!33} & .0125 \cellcolor{green!27} & .0093 \cellcolor{green!33} & .0533 \cellcolor{red!35} & \textbf{.0130} \cellcolor{green!35} & .0135 \cellcolor{green!34} & .0144 \cellcolor{green!32} & .0208 \cellcolor{green!21} & .0146 \cellcolor{green!32} & .0654 \cellcolor{red!35} & \textbf{.0196}$^{\phantom{\ddag}}$ \cellcolor{green!35} & .0205 \cellcolor{green!33} & .0221 \cellcolor{green!31} & .0370$^{\phantom{\ddag}}$ \cellcolor{green!8} & .0215 \cellcolor{green!32} \\
\texttt{Mediamill} & .1697$^{\phantom{\ddag}}$ \cellcolor{red!35} & .0159$^{\phantom{\ddag}}$ \cellcolor{green!34} & .0159 \cellcolor{green!34} & .0184 \cellcolor{green!33} & .0158 \cellcolor{green!34} & \textbf{.0157} \cellcolor{green!35} & .1736 \cellcolor{red!35} & .0297 \cellcolor{green!31} & .0298 \cellcolor{green!31} & .0259 \cellcolor{green!33} & \textbf{.0229} \cellcolor{green!35} & .0251 \cellcolor{green!33} & .1806 \cellcolor{red!35} & .0413$^{\phantom{\ddag}}$ \cellcolor{green!29} & .0414 \cellcolor{green!29} & .0316 \cellcolor{green!33} & \textbf{.0294}$^{\phantom{\ddag}}$ \cellcolor{green!35} & .0322 \cellcolor{green!33} \\
\texttt{Bibtex} & .0354$^{\phantom{\ddag}}$ \cellcolor{red!35} & .0092$^{\dag\phantom{\dag}}$ \cellcolor{green!34} & \textbf{.0092} \cellcolor{green!35} & .0097 \cellcolor{green!33} & .0104 \cellcolor{green!31} & .0092 \cellcolor{green!34} & .0374 \cellcolor{red!35} & .0126 \cellcolor{green!32} & .0127 \cellcolor{green!32} & .0125 \cellcolor{green!32} & .0147 \cellcolor{green!26} & \textbf{.0116} \cellcolor{green!35} & .0423 \cellcolor{red!35} & .0153$^{\phantom{\ddag}}$ \cellcolor{green!29} & .0154 \cellcolor{green!28} & .0140 \cellcolor{green!32} & .0199$^{\phantom{\ddag}}$ \cellcolor{green!18} & \textbf{.0129} \cellcolor{green!35} \\
\texttt{Corel5k} & .0582$^{\phantom{\ddag}}$ \cellcolor{red!35} & .0079$^{\phantom{\ddag}}$ \cellcolor{green!34} & .0080 \cellcolor{green!34} & \textbf{.0074} \cellcolor{green!35} & .0087 \cellcolor{green!33} & .0075 \cellcolor{green!34} & .0585 \cellcolor{red!35} & .0086 \cellcolor{green!34} & .0088 \cellcolor{green!34} & .0084 \cellcolor{green!34} & .0091 \cellcolor{green!33} & \textbf{.0082} \cellcolor{green!35} & .0594 \cellcolor{red!35} & .0088$^{\dag\phantom{\dag}}$ \cellcolor{green!34} & .0096 \cellcolor{green!33} & .0094 \cellcolor{green!34} & .0100$^{\phantom{\ddag}}$ \cellcolor{green!33} & \textbf{.0087} \cellcolor{green!35} \\
\texttt{Delicious} & .1420$^{\phantom{\ddag}}$ \cellcolor{red!35} & \textbf{.0089}$^{\phantom{\ddag}}$ \cellcolor{green!35} & .0091 \cellcolor{green!34} & .0093 \cellcolor{green!34} & .0093 \cellcolor{green!34} & .0093 \cellcolor{green!34} & .1417 \cellcolor{red!35} & .0121 \cellcolor{green!33} & .0124 \cellcolor{green!33} & .0108 \cellcolor{green!34} & \textbf{.0102} \cellcolor{green!35} & .0109 \cellcolor{green!34} & .1238 \cellcolor{red!35} & .0126$^{\phantom{\ddag}}$ \cellcolor{green!32} & .0131 \cellcolor{green!32} & .0102 \cellcolor{green!34} & \textbf{.0092}$^{\phantom{\ddag}}$ \cellcolor{green!35} & .0104 \cellcolor{green!34} \\
\hline
\multicolumn{1}{|c|}{Average} & .0677$^{\phantom{\ddag}}$ \cellcolor{red!35} & .0159$^{\phantom{\ddag}}$ \cellcolor{green!34} & \textbf{.0157} \cellcolor{green!35} & .0180 \cellcolor{green!31} & .0186 \cellcolor{green!31} & .0177 \cellcolor{green!32} & .0761 \cellcolor{red!35} & .0308 \cellcolor{green!32} & .0303 \cellcolor{green!32} & .0288 \cellcolor{green!34} & .0300 \cellcolor{green!33} & \textbf{.0288} \cellcolor{green!35} & .1012 \cellcolor{red!35} & .0560$^{\phantom{\ddag}}$ \cellcolor{green!21} & .0564 \cellcolor{green!20} & .0454 \cellcolor{green!34} & .0529$^{\phantom{\ddag}}$ \cellcolor{green!25} & \textbf{.0452} \cellcolor{green!35} \\
\multicolumn{1}{|c|}{Rank Average} &  5.2 \cellcolor{red!35} & \textbf{2.0} \cellcolor{green!35} &  2.2 \cellcolor{green!30} &  4.1 \cellcolor{red!10} &  4.1 \cellcolor{red!10} &  3.5 \cellcolor{green!2} &  5.3 \cellcolor{red!35} &  3.4 \cellcolor{green!17} &  3.3 \cellcolor{green!19} & \textbf{2.8} \cellcolor{green!35} &  3.3 \cellcolor{green!21} &  2.9 \cellcolor{green!31} &  5.3 \cellcolor{red!35} &  3.4 \cellcolor{green!14} &  3.5 \cellcolor{green!12} &  2.9 \cellcolor{green!28} &  3.4 \cellcolor{green!14} & \textbf{2.6} \cellcolor{green!35} \\

                \hline
                \end{tabular}%
                }
            
\end{table}


These results show that all the multi-label aggregation methods
perform comparably in the low-shift regime, although the LP-based
methods tend to perform slightly better. In the mid-shift and
high-shift regimes the regression-based strategies tend to fare
better. These results, obtained on the test set, are well correlated
with the results we obtain during model selection on the validation
set; our choice of \textsc{SVR-RQ} as a representative method for
\mqbc\ was indeed based on the performance of the different
multi-label aggregation methods obtained in the validation phase. 

The
most important observation we can draw from this table is that all
these methods tend to outperform not only the \bqbc\ system (as
expected) but also all the variants from the \bqmc\ group explored in
Section~\ref{sec:exp:mlc}, which may be an indication that in \MLQ,
bringing to bear the stochastic correlations among classes at the
aggregation phase is more effective
than doing so at the classification phase.

\subsection{Multi-Label Aggregation for Non-Aggregative Quantifiers}
\label{sec:MLAfornonaggregative}

\noindent Since the methods of type \mqmc\ that we have proposed in this paper have proven 
to be the most effective in all our experiments, we want to add an 
important observation about them.

Concerning our regression-based RQ method described in
Section~\ref{sec:quant:ml:reg}, although we have assumed, for ease of
exposition, that the quantifier $q$ is an aggregative one, this
assumption is not strictly necessary, since the regressor $r$ does not look at
predicted class labels for individual datapoints, but only at the
class prevalence estimates returned by the underlying quantifier $q$. 
A similar observation can be made for our label powerset -based LPQ method 
described in
Section~\ref{sec:quant:ml:lp}; this method leverages a single-label multiclass quantifier $q$ and uses its class prevalence estimates, and does not require any prediction at the level of the individual datapoint, which means that aggregative methods and non-aggregative methods are equally suitable for training $q$. In other words, both RQ and LPQ can use any type of quantification method, aggregative or non-aggregative.

The reasons why in this paper we have focused on aggregative quantifiers are (i) ease of explanation, and (b) the fact that, as a recent large-scale experimental study has confirmed~\cite{Castano:2021le}, non-aggregative quantification methods (such as the HDx method of~\cite{Gonzalez-Castro:2013fk}) are, from the point of view of sheer performance, not yet on a par with aggregative methods. However, the above observations indicate that, should high-performance non-aggregative quantification methods spring up in the future, RQ and LPQ can be used in connection with them straightaway.
%


\section{Conclusions}
\label{sec:conclusions}

\noindent In this paper we have investigated \MLQ, a quantification
task which had remained, since the origins of quantification research,
essentially unexplored. We have proposed the first protocol for the
evaluation of \MLQ\ systems that is able to confront these systems
with samples that exhibit from low to high levels of prior probability
shift. For ease of exposition we have particularly focused on multi-label quantifiers 
that work by aggregating predictions for individual datapoints issued by 
a classifier (``aggregative'' multi-label quantifiers), and have subdivided them in four groups, based on whether the correlations between classes are brought to bear in the classification stage (\bqmc), in the quantification stage (\mqbc), in both stages (\mqmc), or in neither of the two stages (\bqbc).

We have also described and experimentally compared a number of MLQ methods; 
some of them (specifically: those in the \bqbc\ and \bqmc\ groups) are 
trivial combinations of available classification and quantification methods, while others (specifically: those in the \mqbc\ and \mqmc\ groups) are non-obvious, and proposed here for the first time. The thorough experimentation that we have carried out on an extensive number of datasets has clearly shown that there is a substantial improvement in performance that
derives from simply replacing binary classifiers with truly
multi-label classifiers (i.e., from switching from BC to MLC), and that there is an equally substantial improvement when binary aggregation is replaced by truly multi-label aggregation (i.e., when  switching from BA to MLA). Consistently with these two intuitions, \mqmc\ methods unequivocally prove the best of the lot; of the two \mqmc\ methods we have proposed, RQ proves clearly superior to LPQ.  In the light of this superiority of MLA with respect to BA, it is also interesting that both RQ and LPQ can be straightforwardly used in association to non-aggregative quantifiers too.



\section*{Acknowledgments}
  \noindent The work of A.\ Moreo and F.\ Sebastiani has been
  supported by the \textsf{SoBigData++} project, funded by the
  European Commission (Grant 871042) under the H2020 Programme
  INFRAIA-2019-1, and by the \textsf{AI4Media} project, funded by the
  European Commission (Grant 951911) under the H2020 Programme
  ICT-48-2020; the authors' opinions do not necessarily reflect those
  of the European Commission. The work of M.\ Francisco has been
  supported by the FPI 2017 predoctoral programme, from the Spanish
  Ministry of Economy and Competitiveness (MINECO), grant
  BES-2017-081202.


\bibliographystyle{unsrt}

\clearpage
\newpage
\appendix


\section{Evaluation in Terms of Relative Absolute Error}
\label{sec:app:rae}

\noindent While the tables presented in the main body of the paper
report the results of our experiments in terms of the absolute error
(AE) measure, in this section we present, for the sake of
completeness, the results in terms of relative absolute error
(RAE). We do not comment on these results since the trends that emerge
from them are essentially the same as for the AE measure. Tables
\ref{tab:cc_general:mrae}, \ref{tab:pcc_general:mrae},
\ref{tab:acc_general:mrae}, \ref{tab:pacc_general:mrae},
\ref{tab:sld_general:mrae}, \ref{tab:mlc:mrae}, \ref{tab:mlq:mrae},
are the RAE equivalents of Tables \ref{tab:cc_general},
\ref{tab:pcc_general}, \ref{tab:acc_general}, \ref{tab:pacc_general},
\ref{tab:sld_general}, \ref{tab:mlc}, \ref{tab:mlq}, respectively

\begin{table}[h!]
  \centering
  \caption{Values of \rabse\ obtained in our experiments for different
  amounts of shift using CC as the base quantifier. Notational
  conventions are as in Table~\ref{tab:cc_general}.}
  \label{tab:cc_general:mrae}
  
                \resizebox{\textwidth}{!}{%
\begin{tabular}{|l|cccc|cccc|cccc|} \cline{2-13}
\multicolumn{1}{c|}{} & \multicolumn{4}{c|}{low shift} & \multicolumn{4}{c|}{mid shift} & \multicolumn{4}{c|}{high shift}\\\cline{2-13}

\multicolumn{1}{c|}{} & \side{\bqbc} & \side{\bqmc} & \side{\mqbc} & \side{\mqmc\phantom{0}} & \side{\bqbc} & \side{\bqmc} & \side{\mqbc} & \side{\mqmc\phantom{0}} & \side{\bqbc} & \side{\bqmc} & \side{\mqbc} & \side{\mqmc\phantom{0}}\\\hline
\texttt{Emotions} & 0.1728 \cellcolor{red!35} & 0.1463$^{\phantom{\ddag}}$ \cellcolor{red!2} & \textbf{0.1163} \cellcolor{green!35} & 0.1529 \cellcolor{red!10} & 0.3138 \cellcolor{green!6} & 0.3180 \cellcolor{green!1} & \textbf{0.2872} \cellcolor{green!35} & 0.3523$^{\phantom{\ddag}}$ \cellcolor{red!35} & 0.4024 \cellcolor{green!17} & 0.4225$^{\phantom{\ddag}}$ \cellcolor{green!9} & \textbf{0.3575} \cellcolor{green!35} & 0.5395$^{\phantom{\ddag}}$ \cellcolor{red!35} \\
\texttt{Scene} & 0.3509 \cellcolor{red!35} & 0.2647$^{\phantom{\ddag}}$ \cellcolor{red!5} & 0.1803 \cellcolor{green!23} & \textbf{0.1464} \cellcolor{green!35} & 0.5922 \cellcolor{red!35} & 0.4659 \cellcolor{red!9} & 0.3044 \cellcolor{green!23} & \textbf{0.2490}$^{\phantom{\ddag}}$ \cellcolor{green!35} &  2.7287 \cellcolor{red!35} &  2.2747$^{\phantom{\ddag}}$ \cellcolor{red!12} & \textbf{1.2898} \cellcolor{green!35} &  1.3553$^{\ddag}$ \cellcolor{green!31} \\
\texttt{Yeast} & 0.7124 \cellcolor{red!35} & 0.4604$^{\phantom{\ddag}}$ \cellcolor{green!0} & \textbf{0.2163} \cellcolor{green!35} & 0.2254 \cellcolor{green!33} & 0.7935 \cellcolor{red!35} & 0.5856 \cellcolor{red!1} & \textbf{0.3626} \cellcolor{green!35} & 0.4000$^{\phantom{\ddag}}$ \cellcolor{green!28} &  1.2682 \cellcolor{red!35} &  1.0984$^{\phantom{\ddag}}$ \cellcolor{green!7} & \textbf{0.9905} \cellcolor{green!35} &  1.0610$^{\phantom{\ddag}}$ \cellcolor{green!17} \\
\hline
                \end{tabular}%
                }
            
\end{table}

\begin{table}[h!]
  \centering
  \caption{Values of \rabse\ obtained in our experiments for different
  amounts of shift using PCC as the base quantifier. Notational
  conventions are as in Table~\ref{tab:cc_general}.}
  \label{tab:pcc_general:mrae}
  
                \resizebox{\textwidth}{!}{%
                    \begin{tabular}{|l|cccc|cccc|cccc|} \cline{2-13}
\multicolumn{1}{c|}{} & \multicolumn{4}{c|}{low shift} & \multicolumn{4}{c|}{mid shift} & \multicolumn{4}{c|}{high shift}\\\cline{2-13}

\multicolumn{1}{c|}{} & \side{\bqbc} & \side{\bqmc} & \side{\mqbc} & \side{\mqmc} & \side{\bqbc} & \side{\bqmc} & \side{\mqbc} & \side{\mqmc} & \side{\bqbc} & \side{\bqmc} & \side{\mqbc} & \side{\mqmc\phantom{0}}\\\hline
Emotions & 0.3105 \cellcolor{red!35} & 0.1720 \cellcolor{green!18} & \textbf{0.1297} \cellcolor{green!35} & 0.1318$^{\phantom{\ddag}}$ \cellcolor{green!34} & 0.4089 \cellcolor{red!35} & 0.2945 \cellcolor{green!27} & 0.3116 \cellcolor{green!18} & \textbf{0.2812}$^{\phantom{\ddag}}$ \cellcolor{green!35} & 0.4459 \cellcolor{green!25} & \textbf{0.4191}$^{\phantom{\ddag}}$ \cellcolor{green!35} & 0.6205 \cellcolor{red!35} & 0.4639$^{\phantom{\ddag}}$ \cellcolor{green!19} \\
Scene & 0.2023 \cellcolor{red!35} & 0.1416 \cellcolor{green!26} & 0.1530 \cellcolor{green!14} & \textbf{0.1332}$^{\phantom{\ddag}}$ \cellcolor{green!35} & 0.4292 \cellcolor{red!35} & 0.2893 \cellcolor{green!11} & 0.2266 \cellcolor{green!31} & \textbf{0.2171}$^{\phantom{\ddag}}$ \cellcolor{green!35} &  1.7304 \cellcolor{red!35} &  1.1136$^{\ddag}$ \cellcolor{green!34} & \textbf{1.1121} \cellcolor{green!35} &  1.3674$^{\phantom{\ddag}}$ \cellcolor{green!6} \\
Yeast &  1.4781 \cellcolor{red!35} & 0.8704 \cellcolor{red!1} & 0.2080 \cellcolor{green!34} & \textbf{0.2024}$^{\phantom{\ddag}}$ \cellcolor{green!35} &  2.0566 \cellcolor{red!35} &  1.0807 \cellcolor{green!4} & 0.3571 \cellcolor{green!33} & \textbf{0.3319}$^{\phantom{\ddag}}$ \cellcolor{green!35} &  2.9340 \cellcolor{red!35} &  1.4858$^{\phantom{\ddag}}$ \cellcolor{green!16} &  1.0029 \cellcolor{green!33} & \textbf{0.9498}$^{\phantom{\ddag}}$ \cellcolor{green!35} \\
\hline
                \end{tabular}%
                }
            
\end{table}

\begin{table}[h!]
  \centering
  \caption{Values of \rabse\ obtained in our experiments for different
  amounts of shift using ACC as the base quantifier. Notational
  conventions are as in Table~\ref{tab:cc_general}.}
  \label{tab:acc_general:mrae}
  
                \resizebox{\textwidth}{!}{%
                    \begin{tabular}{|l|cccc|cccc|cccc|} \cline{2-13}
\multicolumn{1}{c|}{} & \multicolumn{4}{c|}{low shift} & \multicolumn{4}{c|}{mid shift} & \multicolumn{4}{c|}{high shift}\\\cline{2-13}

\multicolumn{1}{c|}{} & \side{\bqbc} & \side{\bqmc} & \side{\mqbc} & \side{\mqmc\phantom{0}} & \side{\bqbc} & \side{\bqmc} & \side{\mqbc} & \side{\mqmc\phantom{0}} & \side{\bqbc} & \side{\bqmc} & \side{\mqbc} & \side{\mqmc\phantom{0}}\\\hline
\texttt{Emotions} & 0.2491 \cellcolor{red!16} & 0.2991 \cellcolor{red!35} & 0.1579$^{\phantom{\ddag}}$ \cellcolor{green!16} & \textbf{0.1069}$^{\phantom{\ddag}}$ \cellcolor{green!35} & 0.4297 \cellcolor{red!34} & 0.4301 \cellcolor{red!35} & 0.3294 \cellcolor{green!8} & \textbf{0.2672} \cellcolor{green!35} & 0.4519$^{\phantom{\ddag}}$ \cellcolor{green!26} & 0.4690 \cellcolor{green!20} & 0.6290$^{\phantom{\ddag}}$ \cellcolor{red!35} & \textbf{0.4285}$^{\phantom{\ddag}}$ \cellcolor{green!35} \\
\texttt{Scene} & 0.2035 \cellcolor{red!35} & 0.1676 \cellcolor{green!17} & 0.1792$^{\phantom{\ddag}}$ \cellcolor{green!0} & \textbf{0.1554}$^{\phantom{\ddag}}$ \cellcolor{green!35} & 0.3371 \cellcolor{red!8} & 0.3863 \cellcolor{red!35} & 0.2778 \cellcolor{green!24} & \textbf{0.2576} \cellcolor{green!35} &  1.3324$^{\ddag}$ \cellcolor{green!26} &  1.5409 \cellcolor{red!35} & \textbf{1.3054}$^{\phantom{\ddag}}$ \cellcolor{green!35} &  1.4589$^{\phantom{\ddag}}$ \cellcolor{red!10} \\
\texttt{Yeast} &  2.2331 \cellcolor{red!35} & 0.5355 \cellcolor{green!23} & \textbf{0.2116}$^{\phantom{\ddag}}$ \cellcolor{green!35} & 0.2121$^{\dag}$ \cellcolor{green!34} &  2.4014 \cellcolor{red!35} & 0.6702 \cellcolor{green!24} & \textbf{0.3733} \cellcolor{green!35} & 0.3834 \cellcolor{green!34} &  2.7847$^{\phantom{\ddag}}$ \cellcolor{red!35} &  1.4287 \cellcolor{green!17} & \textbf{0.9628}$^{\phantom{\ddag}}$ \cellcolor{green!35} &  1.1162$^{\phantom{\ddag}}$ \cellcolor{green!29} \\
\hline
                \end{tabular}%
                }
            
\end{table}

\begin{table}[h!]
  \centering
  \caption{Values of \rabse\ obtained in our experiments for different
  amounts of shift using PACC as the base quantifier. Notational
  conventions are as in Table~\ref{tab:cc_general}.}
  \label{tab:pacc_general:mrae}
  
                \resizebox{\textwidth}{!}{%
                    \begin{tabular}{|l|cccc|cccc|cccc|} \cline{2-13}
\multicolumn{1}{c|}{} & \multicolumn{4}{c|}{low shift} & \multicolumn{4}{c|}{mid shift} & \multicolumn{4}{c|}{high shift}\\\cline{2-13}

\multicolumn{1}{c|}{} & \side{\bqbc} & \side{\bqmc} & \side{\mqbc} & \side{\mqmc\phantom{0}} & \side{\bqbc} & \side{\bqmc} & \side{\mqbc} & \side{\mqmc\phantom{0}} & \side{\bqbc} & \side{\bqmc} & \side{\mqbc} & \side{\mqmc\phantom{0}}\\\hline
\texttt{Emotions} & 0.3105 \cellcolor{red!35} & 0.1720 \cellcolor{green!18} & \textbf{0.1297} \cellcolor{green!35} & 0.1318$^{\phantom{\ddag}}$ \cellcolor{green!34} & 0.4089 \cellcolor{red!35} & 0.2945 \cellcolor{green!27} & 0.3116 \cellcolor{green!18} & \textbf{0.2812}$^{\phantom{\ddag}}$ \cellcolor{green!35} & 0.4459 \cellcolor{green!25} & \textbf{0.4191}$^{\phantom{\ddag}}$ \cellcolor{green!35} & 0.6205 \cellcolor{red!35} & 0.4639$^{\phantom{\ddag}}$ \cellcolor{green!19} \\
\texttt{Scene} & 0.2023 \cellcolor{red!35} & 0.1416 \cellcolor{green!26} & 0.1530 \cellcolor{green!14} & \textbf{0.1332}$^{\phantom{\ddag}}$ \cellcolor{green!35} & 0.4292 \cellcolor{red!35} & 0.2893 \cellcolor{green!11} & 0.2266 \cellcolor{green!31} & \textbf{0.2171}$^{\phantom{\ddag}}$ \cellcolor{green!35} &  1.7304 \cellcolor{red!35} &  1.1136$^{\ddag}$ \cellcolor{green!34} & \textbf{1.1121} \cellcolor{green!35} &  1.3674$^{\phantom{\ddag}}$ \cellcolor{green!6} \\
\texttt{Yeast} &  1.4781 \cellcolor{red!35} & 0.8704 \cellcolor{red!1} & 0.2080 \cellcolor{green!34} & \textbf{0.2024}$^{\phantom{\ddag}}$ \cellcolor{green!35} &  2.0566 \cellcolor{red!35} &  1.0807 \cellcolor{green!4} & 0.3571 \cellcolor{green!33} & \textbf{0.3319}$^{\phantom{\ddag}}$ \cellcolor{green!35} &  2.9340 \cellcolor{red!35} &  1.4858$^{\phantom{\ddag}}$ \cellcolor{green!16} &  1.0029 \cellcolor{green!33} & \textbf{0.9498}$^{\phantom{\ddag}}$ \cellcolor{green!35} \\
\hline
                \end{tabular}%
                }
            
\end{table}

\begin{table}[h!]
  \centering
  \caption{Values of \rabse\ obtained in our experiments for different
  amounts of shift using SLD as the base quantifier. Notational
  conventions are as in Table~\ref{tab:cc_general}.}
  \label{tab:sld_general:mrae}
  
                \resizebox{\textwidth}{!}{%
                    \begin{tabular}{|l|cccc|cccc|cccc|} \cline{2-13}
\multicolumn{1}{c|}{} & \multicolumn{4}{c|}{low shift} & \multicolumn{4}{c|}{mid shift} & \multicolumn{4}{c|}{high shift}\\\cline{2-13}

\multicolumn{1}{c|}{} & \side{\bqbc} & \side{\bqmc} & \side{\mqbc} & \side{\mqmc\phantom{0}} & \side{\bqbc} & \side{\bqmc} & \side{\mqbc} & \side{\mqmc\phantom{0}} & \side{\bqbc} & \side{\bqmc} & \side{\mqbc} & \side{\mqmc\phantom{0}}\\\hline
\texttt{Emotions} & 0.5117 \cellcolor{red!35} & 0.1310 \cellcolor{green!33} & 0.1646 \cellcolor{green!27} & \textbf{0.1199} \cellcolor{green!35} & 0.6961 \cellcolor{red!35} & \textbf{0.2455} \cellcolor{green!35} & 0.3179$^{\phantom{\ddag}}$ \cellcolor{green!23} & 0.2907$^{\dag}$ \cellcolor{green!27} & 0.5526 \cellcolor{red!35} & \textbf{0.2769} \cellcolor{green!35} & 0.4625 \cellcolor{red!12} & 0.3885 \cellcolor{green!6} \\
\texttt{Scene} & 0.1519 \cellcolor{green!22} & 0.1914 \cellcolor{red!35} & \textbf{0.1432} \cellcolor{green!35} & 0.1847 \cellcolor{red!25} & \textbf{0.2518} \cellcolor{green!35} & 0.3744 \cellcolor{red!35} & 0.2587$^{\dag}$ \cellcolor{green!31} & 0.3693$^{\phantom{\ddag}}$ \cellcolor{red!32} & \textbf{1.0904} \cellcolor{green!35} &  2.1677 \cellcolor{red!35} &  1.3721 \cellcolor{green!16} &  1.9117 \cellcolor{red!18} \\
\texttt{Yeast} &  1.3664 \cellcolor{red!35} & 0.3688 \cellcolor{green!25} & \textbf{0.2111} \cellcolor{green!35} & 0.2210 \cellcolor{green!34} &  1.3507 \cellcolor{red!35} & 0.4597 \cellcolor{green!26} & 0.3471$^{\phantom{\ddag}}$ \cellcolor{green!34} & \textbf{0.3409}$^{\phantom{\ddag}}$ \cellcolor{green!35} &  1.8907 \cellcolor{red!35} & 0.9233 \cellcolor{green!32} &  1.0121 \cellcolor{green!26} & \textbf{0.8924} \cellcolor{green!35} \\
\hline
                \end{tabular}%
                }
            
\end{table}

\begin{table}[h!]
  \centering
  \caption{Values of \rabse\ obtained for different multi-label
  classifiers using PCC as the base quantifier in \bqmc.}
  \label{tab:mlc:mrae}
  
                \resizebox{\textwidth}{!}{%
            \setlength\tabcolsep{1.5pt}
            
                    \begin{tabular}{|l|c|ccccccc|c|ccccccc|c|ccccccc|} \cline{2-25}
\multicolumn{1}{c|}{} & \multicolumn{8}{c|}{low shift} & \multicolumn{8}{c|}{mid shift} & \multicolumn{8}{c|}{high shift}\\\cline{2-25}

\multicolumn{1}{c|}{} & \side{\bqbc} & \side{\textsc{ML-knn}\phantom{0}} & \side{\textsc{DT}} & \side{\textsc{RF}} & \side{\textsc{CChain}} & \side{\textsc{CLEMS}} & \side{\textsc{LSC}} & \side{\textsc{SG}} & \side{\bqbc} & \side{\textsc{ML-knn}\phantom{0}} & \side{\textsc{DT}} & \side{\textsc{RF}} & \side{\textsc{CChain}} & \side{\textsc{CLEMS}} & \side{\textsc{LSC}} & \side{\textsc{SG}} & \side{\bqbc} & \side{\textsc{ML-knn}\phantom{0}} & \side{\textsc{DT}} & \side{\textsc{RF}} & \side{\textsc{CChain}} & \side{\textsc{CLEMS}} & \side{\textsc{LSC}} & \side{\textsc{SG}}\\\hline
\texttt{Emotions} & 0.104$^{\dag}$ \cellcolor{green!34} & 0.511 \cellcolor{red!19} & 0.160 \cellcolor{green!27} & 0.106 \cellcolor{green!34} & 0.129 \cellcolor{green!31} & 0.512 \cellcolor{red!19} & 0.625 \cellcolor{red!35} & \textbf{0.104} \cellcolor{green!35} & 0.366 \cellcolor{green!29} & 0.589 \cellcolor{red!17} & 0.423 \cellcolor{green!17} & 0.388 \cellcolor{green!25} & 0.373 \cellcolor{green!28} & 0.590 \cellcolor{red!18} & 0.670 \cellcolor{red!35} & \textbf{0.342} \cellcolor{green!35} & 0.543 \cellcolor{green!28} & 0.790 \cellcolor{red!11} & 0.548 \cellcolor{green!27} & 0.556$^{\phantom{\ddag}}$ \cellcolor{green!26} & 0.593 \cellcolor{green!20} & 0.788 \cellcolor{red!11} & 0.932$^{\phantom{\ddag}}$ \cellcolor{red!35} & \textbf{0.504}$^{\phantom{\ddag}}$ \cellcolor{green!35} \\
\hline
                \end{tabular}%
                }
            
\end{table}

\begin{table}[h!]
  \centering
  \caption{Values of \rabse\ obtained for different multi-label
  aggregation methods in \mqbc.}
  \label{tab:mlq:mrae}
  
                \resizebox{\textwidth}{!}{%
            \setlength\tabcolsep{1.5pt}
            
                    \begin{tabular}{|l|c|ccccc|c|ccccc|c|ccccc|} \cline{2-19}
\multicolumn{1}{c|}{} & \multicolumn{6}{c|}{low shift} & \multicolumn{6}{c|}{mid shift} & \multicolumn{6}{c|}{high shift}\\\cline{2-19}

\multicolumn{1}{c|}{} & \side{\bqbc} & \side{\textsc{RakEL-LPQ}} & \side{\textsc{kMeans-LPQ}\phantom{0}} & \side{\textsc{Ridge-RQ}} & \side{\textsc{RF-RQ}} & \side{\textsc{SVR-RQ}} & \side{\bqbc} & \side{\textsc{RakEL-LPQ}} & \side{\textsc{kMeans-LPQ}\phantom{0}} & \side{\textsc{Ridge-RQ}} & \side{\textsc{RF-RQ}} & \side{\textsc{SVR-RQ}} & \side{\bqbc} & \side{\textsc{RakEL-LPQ}} & \side{\textsc{kMeans-LPQ}\phantom{0}} & \side{\textsc{Ridge-RQ}} & \side{\textsc{RF-RQ}} & \side{\textsc{SVR-RQ}}\\\hline
\texttt{Emotions} & 0.1045 \cellcolor{green!32} & 0.1119 \cellcolor{green!20} & \textbf{0.1028} \cellcolor{green!35} & 0.1202 \cellcolor{green!6} & 0.1179 \cellcolor{green!10} & 0.1454$^{\phantom{\ddag}}$ \cellcolor{red!35} & 0.3657 \cellcolor{red!35} & 0.3507$^{\phantom{\ddag}}$ \cellcolor{red!24} & 0.3490 \cellcolor{red!22} & 0.2843 \cellcolor{green!24} & \textbf{0.2700} \cellcolor{green!35} & 0.2935 \cellcolor{green!17} & 0.5430 \cellcolor{red!3} & 0.5707$^{\phantom{\ddag}}$ \cellcolor{red!12} & 0.5265$^{\phantom{\ddag}}$ \cellcolor{green!2} & \textbf{0.4284}$^{\phantom{\ddag}}$ \cellcolor{green!35} & 0.6379$^{\phantom{\ddag}}$ \cellcolor{red!35} & 0.4726$^{\phantom{\ddag}}$ \cellcolor{green!20} \\
\hline
                \end{tabular}%
                }
            
\end{table}


\clearpage
\newpage

\section{Exploring other Regressors in RQ}
\label{sec:otherregressors}

\noindent We here report additional experiments that extend the ones
presented in Section~\ref{sec:exp:mlq}; the present experiments
concern the use of regression algorithms other than ridge regression (Ridge-RQ), random forest regression (RF-RQ), and linear SVR (SVR-RQ), which were the 
only regression
algorithms we considered in
Section~\ref{sec:exp:mlq}. Tables~\ref{tab:mlqreg}
and~\ref{tab:mlqreg:mrae} report the results of these experiments in
terms of AE and RAE, respectively, obtained by optimizing the hyperparameters shown in Table~\ref{tab:modsel:reg}. These results show that there are no substantial differences in performance for the low-shift regime, while these differences are instead noticeable in the mid-shift and (especially) in the high-shift regimes. These results are well correlated with the results we obtained in the validation phase, on which we relied upon for choosing ridge regression, random forest regression, and linear SVR, as representative regression models.

\begin{table}[h!]
  \centering
  \caption{\abse\ for different regressors for multi-label
  quantifiers}
  \label{tab:mlqreg}
  
                \resizebox{\textwidth}{!}{%
            \setlength\tabcolsep{1.5pt}
            
                    \begin{tabular}{|l|c|cccccc|c|cccccc|c|cccccc|} \cline{2-22}
\multicolumn{1}{c|}{} & \multicolumn{7}{c|}{low shift} & \multicolumn{7}{c|}{mid shift} & \multicolumn{7}{c|}{high shift}\\\cline{2-22}

\multicolumn{1}{c|}{} & \side{\bqbc} & \side{Ridge (Ridge-RQ)} & \side{MultitaskLasso} & \side{RandomForest (RF-RQ) \mbox{ }} & \side{StackedLinearSVR} & \side{ChainedLinearSVR\phantom{0}} & \side{LinearSVR (SVR-RQ)\mbox{ }} & \side{\bqbc} & \side{Ridge (Ridge-RQ)} & \side{MultitaskLasso} & \side{RandomForest (RF-RQ) \mbox{ }} & \side{StackedLinearSVR} & \side{ChainedLinearSVR\phantom{0}} & \side{LinearSVR (SVR-RQ)\mbox{ }} & \side{\bqbc} & \side{Ridge (Ridge-RQ)} & \side{MultitaskLasso} & \side{RandomForest (RF-RQ) \mbox{ }} & \side{StackedLinearSVR} & \side{ChainedLinearSVR\phantom{0}} & \side{LinearSVR (SVR-RQ)\mbox{ }}\\\hline
\texttt{Emotions} & \textbf{.0418} \cellcolor{green!35} & .0500 \cellcolor{green!13} & .0633 \cellcolor{red!21} & .0484$^{\phantom{\ddag}}$ \cellcolor{green!17} & .0687 \cellcolor{red!35} & .0574 \cellcolor{red!5} & .0586$^{\phantom{\ddag}}$ \cellcolor{red!8} & .0685 \cellcolor{green!27} & \textbf{.0626} \cellcolor{green!35} & .0811 \cellcolor{green!10} & .0707 \cellcolor{green!24} & .1038 \cellcolor{red!19} & .1153 \cellcolor{red!35} & .0720$^{\phantom{\ddag}}$ \cellcolor{green!22} & .0923 \cellcolor{green!24} & \textbf{.0782}$^{\phantom{\ddag}}$ \cellcolor{green!35} & .1029 \cellcolor{green!17} & .1235$^{\phantom{\ddag}}$ \cellcolor{green!2} & .1295 \cellcolor{red!1} & .1767 \cellcolor{red!35} & .0951 \cellcolor{green!22} \\
\texttt{Scene} & .0842 \cellcolor{red!35} & .0350 \cellcolor{green!28} & .0440 \cellcolor{green!17} & \textbf{.0303}$^{\phantom{\ddag}}$ \cellcolor{green!35} & .0363 \cellcolor{green!27} & .0359 \cellcolor{green!27} & .0384$^{\phantom{\ddag}}$ \cellcolor{green!24} & .1140 \cellcolor{red!35} & .0454 \cellcolor{green!22} & .1110 \cellcolor{red!32} & \textbf{.0306} \cellcolor{green!35} & .0859 \cellcolor{red!11} & .0868 \cellcolor{red!12} & .0459$^{\phantom{\ddag}}$ \cellcolor{green!22} & .1427 \cellcolor{red!16} & .0581$^{\phantom{\ddag}}$ \cellcolor{green!30} & .1764 \cellcolor{red!35} & .0507$^{\ddag}$ \cellcolor{green!34} & .1251 \cellcolor{red!6} & .1260 \cellcolor{red!7} & \textbf{.0499} \cellcolor{green!35} \\
\texttt{Yeast} & .1756 \cellcolor{red!35} & .0477 \cellcolor{green!33} & .0491 \cellcolor{green!33} & .0561$^{\phantom{\ddag}}$ \cellcolor{green!29} & \textbf{.0457} \cellcolor{green!35} & .0464 \cellcolor{green!34} & .0480$^{\phantom{\ddag}}$ \cellcolor{green!33} & .1900 \cellcolor{red!35} & .0920 \cellcolor{green!34} & .0943 \cellcolor{green!32} & .0956 \cellcolor{green!31} & .0982 \cellcolor{green!30} & .1004 \cellcolor{green!28} & \textbf{.0913}$^{\phantom{\ddag}}$ \cellcolor{green!35} & .2206 \cellcolor{red!35} & .1374$^{\phantom{\ddag}}$ \cellcolor{green!33} & .1424 \cellcolor{green!29} & .1517$^{\phantom{\ddag}}$ \cellcolor{green!22} & .1517 \cellcolor{green!22} & .1542 \cellcolor{green!20} & \textbf{.1361} \cellcolor{green!35} \\
\texttt{Birds} & .0286 \cellcolor{red!35} & .0213 \cellcolor{green!13} & .0230 \cellcolor{green!1} & .0185$^{\phantom{\ddag}}$ \cellcolor{green!32} & .0217 \cellcolor{green!10} & .0213 \cellcolor{green!13} & \textbf{.0181}$^{\phantom{\ddag}}$ \cellcolor{green!35} & .0328 \cellcolor{red!35} & .0289 \cellcolor{red!3} & .0299 \cellcolor{red!11} & .0251 \cellcolor{green!27} & .0297 \cellcolor{red!10} & .0280 \cellcolor{green!4} & \textbf{.0243}$^{\phantom{\ddag}}$ \cellcolor{green!35} & .0440 \cellcolor{red!35} & .0405$^{\phantom{\ddag}}$ \cellcolor{red!7} & .0424 \cellcolor{red!22} & .0382$^{\phantom{\ddag}}$ \cellcolor{green!10} & .0436 \cellcolor{red!32} & .0427 \cellcolor{red!25} & \textbf{.0351} \cellcolor{green!35} \\
\texttt{Genbase} & \textbf{.0011} \cellcolor{green!35} & .0015 \cellcolor{green!33} & .0137 \cellcolor{red!17} & .0103$^{\phantom{\ddag}}$ \cellcolor{red!3} & .0168 \cellcolor{red!30} & .0177 \cellcolor{red!35} & .0022$^{\phantom{\ddag}}$ \cellcolor{green!30} & \textbf{.0011} \cellcolor{green!35} & .0014 \cellcolor{green!33} & .0175 \cellcolor{red!13} & .0148 \cellcolor{red!5} & .0245 \cellcolor{red!34} & .0246 \cellcolor{red!35} & .0025$^{\phantom{\ddag}}$ \cellcolor{green!30} & \textbf{.0010} \cellcolor{green!35} & .0013$^{\phantom{\ddag}}$ \cellcolor{green!34} & .0173 \cellcolor{green!2} & .0172$^{\phantom{\ddag}}$ \cellcolor{green!3} & .0364 \cellcolor{red!35} & .0347 \cellcolor{red!31} & .0023 \cellcolor{green!32} \\
\texttt{Medical} & \textbf{.0127} \cellcolor{green!35} & .0230 \cellcolor{red!35} & .0186 \cellcolor{red!5} & .0202$^{\phantom{\ddag}}$ \cellcolor{red!16} & .0207 \cellcolor{red!19} & .0194 \cellcolor{red!10} & .0191$^{\phantom{\ddag}}$ \cellcolor{red!8} & \textbf{.0146} \cellcolor{green!35} & .0306 \cellcolor{red!10} & .0263 \cellcolor{green!1} & .0393 \cellcolor{red!35} & .0315 \cellcolor{red!12} & .0265 \cellcolor{green!1} & .0279$^{\phantom{\ddag}}$ \cellcolor{red!2} & \textbf{.0169} \cellcolor{green!35} & .0365$^{\phantom{\ddag}}$ \cellcolor{green!6} & .0314 \cellcolor{green!13} & .0651$^{\phantom{\ddag}}$ \cellcolor{red!35} & .0439 \cellcolor{red!4} & .0380 \cellcolor{green!4} & .0351 \cellcolor{green!8} \\
\texttt{tmc2007\_500} & .1108 \cellcolor{red!35} & .0226 \cellcolor{green!33} & .0260 \cellcolor{green!31} & \textbf{.0211}$^{\phantom{\ddag}}$ \cellcolor{green!35} & .0312 \cellcolor{green!27} & .0272 \cellcolor{green!30} & .0213$^{\ddag}$ \cellcolor{green!34} & .1154 \cellcolor{red!35} & .0274 \cellcolor{green!31} & .0436 \cellcolor{green!19} & \textbf{.0224} \cellcolor{green!35} & .0528 \cellcolor{green!12} & .0508 \cellcolor{green!13} & .0292$^{\phantom{\ddag}}$ \cellcolor{green!29} & .1008 \cellcolor{red!35} & .0282$^{\phantom{\ddag}}$ \cellcolor{green!30} & .0571 \cellcolor{green!4} & \textbf{.0237}$^{\phantom{\ddag}}$ \cellcolor{green!35} & .0726 \cellcolor{red!9} & .0714 \cellcolor{red!8} & .0321 \cellcolor{green!27} \\
\texttt{Ohsumed} & .1004 \cellcolor{red!35} & .0193 \cellcolor{green!34} & .0220 \cellcolor{green!31} & .0183$^{\ddag}$ \cellcolor{green!34} & .0231 \cellcolor{green!30} & .0230 \cellcolor{green!30} & \textbf{.0183}$^{\phantom{\ddag}}$ \cellcolor{green!35} & .1087 \cellcolor{red!35} & .0208 \cellcolor{green!33} & .0363 \cellcolor{green!21} & \textbf{.0184} \cellcolor{green!35} & .0433 \cellcolor{green!15} & .0428 \cellcolor{green!16} & .0209$^{\phantom{\ddag}}$ \cellcolor{green!33} & .1177 \cellcolor{red!35} & .0221$^{\phantom{\ddag}}$ \cellcolor{green!33} & .0472 \cellcolor{green!15} & \textbf{.0201}$^{\phantom{\ddag}}$ \cellcolor{green!35} & .0601 \cellcolor{green!6} & .0591 \cellcolor{green!7} & .0215 \cellcolor{green!33} \\
\texttt{Enron} & .0347 \cellcolor{red!35} & .0192 \cellcolor{green!26} & .0187 \cellcolor{green!27} & .0174$^{\phantom{\ddag}}$ \cellcolor{green!33} & .0180 \cellcolor{green!30} & .0189 \cellcolor{green!27} & \textbf{.0169}$^{\phantom{\ddag}}$ \cellcolor{green!35} & .0397 \cellcolor{red!35} & .0250 \cellcolor{green!22} & .0256 \cellcolor{green!20} & \textbf{.0219} \cellcolor{green!35} & .0267 \cellcolor{green!16} & .0269 \cellcolor{green!15} & .0227$^{\phantom{\ddag}}$ \cellcolor{green!32} & .0439 \cellcolor{red!35} & .0287$^{\phantom{\ddag}}$ \cellcolor{green!11} & .0300 \cellcolor{green!7} & \textbf{.0212}$^{\phantom{\ddag}}$ \cellcolor{green!35} & .0332 \cellcolor{red!1} & .0325 \cellcolor{green!0} & .0253 \cellcolor{green!22} \\
\texttt{Reuters-21578} & .0167 \cellcolor{red!35} & \textbf{.0048} \cellcolor{green!35} & .0072 \cellcolor{green!20} & .0073$^{\phantom{\ddag}}$ \cellcolor{green!20} & .0082 \cellcolor{green!14} & .0082 \cellcolor{green!14} & .0049$^{\phantom{\ddag}}$ \cellcolor{green!34} & .0243 \cellcolor{red!35} & \textbf{.0070} \cellcolor{green!35} & .0114 \cellcolor{green!17} & .0109 \cellcolor{green!19} & .0181 \cellcolor{red!9} & .0182 \cellcolor{red!10} & .0070$^{\ddag}$ \cellcolor{green!34} & .0370 \cellcolor{red!35} & .0103$^{\phantom{\ddag}}$ \cellcolor{green!31} & .0143 \cellcolor{green!21} & .0135$^{\phantom{\ddag}}$ \cellcolor{green!23} & .0267 \cellcolor{red!9} & .0279 \cellcolor{red!12} & \textbf{.0088} \cellcolor{green!35} \\
\texttt{RCV1-v2} & .0456 \cellcolor{red!35} & .0095 \cellcolor{green!34} & .0129 \cellcolor{green!28} & .0125$^{\phantom{\ddag}}$ \cellcolor{green!28} & .0147 \cellcolor{green!24} & .0147 \cellcolor{green!24} & \textbf{.0093}$^{\phantom{\ddag}}$ \cellcolor{green!35} & .0533 \cellcolor{red!35} & \textbf{.0144} \cellcolor{green!35} & .0245 \cellcolor{green!16} & .0208 \cellcolor{green!23} & .0295 \cellcolor{green!7} & .0295 \cellcolor{green!7} & .0146$^{\phantom{\ddag}}$ \cellcolor{green!34} & .0654 \cellcolor{red!35} & .0221$^{\phantom{\ddag}}$ \cellcolor{green!33} & .0366 \cellcolor{green!10} & .0370$^{\phantom{\ddag}}$ \cellcolor{green!10} & .0472 \cellcolor{red!6} & .0472 \cellcolor{red!5} & \textbf{.0215} \cellcolor{green!35} \\
\texttt{Mediamill} & .1697 \cellcolor{red!35} & .0184 \cellcolor{green!33} & .0169 \cellcolor{green!34} & .0158$^{\phantom{\ddag}}$ \cellcolor{green!34} & .0167 \cellcolor{green!34} & .0176 \cellcolor{green!34} & \textbf{.0157}$^{\phantom{\ddag}}$ \cellcolor{green!35} & .1736 \cellcolor{red!35} & .0259 \cellcolor{green!33} & .0266 \cellcolor{green!33} & \textbf{.0229} \cellcolor{green!35} & .0315 \cellcolor{green!30} & .0321 \cellcolor{green!30} & .0251$^{\phantom{\ddag}}$ \cellcolor{green!33} & .1806 \cellcolor{red!35} & .0316$^{\phantom{\ddag}}$ \cellcolor{green!33} & .0321 \cellcolor{green!33} & \textbf{.0294}$^{\phantom{\ddag}}$ \cellcolor{green!35} & .0440 \cellcolor{green!28} & .0442 \cellcolor{green!28} & .0322 \cellcolor{green!33} \\
\texttt{Bibtex} & .0354 \cellcolor{red!35} & .0097 \cellcolor{green!33} & .0105 \cellcolor{green!31} & .0104$^{\phantom{\ddag}}$ \cellcolor{green!31} & .0106 \cellcolor{green!31} & .0111 \cellcolor{green!29} & \textbf{.0092}$^{\phantom{\ddag}}$ \cellcolor{green!35} & .0374 \cellcolor{red!35} & .0125 \cellcolor{green!32} & .0154 \cellcolor{green!24} & .0147 \cellcolor{green!26} & .0168 \cellcolor{green!20} & .0157 \cellcolor{green!23} & \textbf{.0116}$^{\phantom{\ddag}}$ \cellcolor{green!35} & .0423 \cellcolor{red!35} & .0140$^{\phantom{\ddag}}$ \cellcolor{green!32} & .0198 \cellcolor{green!18} & .0199$^{\phantom{\ddag}}$ \cellcolor{green!18} & .0224 \cellcolor{green!12} & .0197 \cellcolor{green!18} & \textbf{.0129} \cellcolor{green!35} \\
Corel5k & .0582 \cellcolor{red!35} & .0074 \cellcolor{green!34} & .0080 \cellcolor{green!33} & .0087$^{\phantom{\ddag}}$ \cellcolor{green!33} & \textbf{.0073} \cellcolor{green!35} & .0091 \cellcolor{green!32} & .0075$^{\phantom{\ddag}}$ \cellcolor{green!34} & .0585 \cellcolor{red!35} & .0084 \cellcolor{green!34} & .0088 \cellcolor{green!34} & .0091 \cellcolor{green!33} & .0088 \cellcolor{green!34} & .0102 \cellcolor{green!32} & \textbf{.0082}$^{\phantom{\ddag}}$ \cellcolor{green!35} & .0594 \cellcolor{red!35} & .0094$^{\phantom{\ddag}}$ \cellcolor{green!34} & .0098 \cellcolor{green!33} & .0100$^{\phantom{\ddag}}$ \cellcolor{green!33} & .0106 \cellcolor{green!32} & .0117 \cellcolor{green!30} & \textbf{.0087} \cellcolor{green!35} \\
\texttt{Delicious} & .1420 \cellcolor{red!35} & \textbf{.0093} \cellcolor{green!35} & .0096 \cellcolor{green!34} & .0093$^{\phantom{\ddag}}$ \cellcolor{green!34} & .0100 \cellcolor{green!34} & .0109 \cellcolor{green!34} & .0093$^{\phantom{\ddag}}$ \cellcolor{green!34} & .1417 \cellcolor{red!35} & .0108 \cellcolor{green!34} & .0134 \cellcolor{green!33} & \textbf{.0102} \cellcolor{green!35} & .0139 \cellcolor{green!33} & .0147 \cellcolor{green!32} & .0109$^{\phantom{\ddag}}$ \cellcolor{green!34} & .1238 \cellcolor{red!35} & .0102$^{\phantom{\ddag}}$ \cellcolor{green!34} & .0151 \cellcolor{green!31} & \textbf{.0092}$^{\phantom{\ddag}}$ \cellcolor{green!35} & .0163 \cellcolor{green!30} & .0172 \cellcolor{green!30} & .0104 \cellcolor{green!34} \\
\hline
\multicolumn{1}{|c|}{Average} & .0677 \cellcolor{red!35} & .0180 \cellcolor{green!34} & .0206 \cellcolor{green!30} & .0186$^{\phantom{\ddag}}$ \cellcolor{green!33} & .0211 \cellcolor{green!30} & .0207 \cellcolor{green!30} & \textbf{.0177}$^{\phantom{\ddag}}$ \cellcolor{green!35} & .0761 \cellcolor{red!35} & .0288 \cellcolor{green!34} & .0401 \cellcolor{green!18} & .0300 \cellcolor{green!33} & .0440 \cellcolor{green!12} & .0444 \cellcolor{green!11} & \textbf{.0288}$^{\phantom{\ddag}}$ \cellcolor{green!35} & .1012 \cellcolor{red!35} & .0454$^{\ddag}$ \cellcolor{green!34} & .0781 \cellcolor{red!6} & .0529$^{\phantom{\ddag}}$ \cellcolor{green!25} & .0788 \cellcolor{red!7} & .0831 \cellcolor{red!12} & \textbf{.0452} \cellcolor{green!35} \\
\multicolumn{1}{|c|}{Rank Average} &  5.8 \cellcolor{red!35} &  3.1 \cellcolor{green!18} &  4.3 \cellcolor{red!5} &  3.0 \cellcolor{green!21} &  4.6 \cellcolor{red!10} &  4.8 \cellcolor{red!14} & \textbf{2.3} \cellcolor{green!35} &  5.9 \cellcolor{red!35} & \textbf{2.3} \cellcolor{green!35} &  4.2 \cellcolor{red!2} &  2.7 \cellcolor{green!27} &  5.2 \cellcolor{red!22} &  5.5 \cellcolor{red!27} &  2.3 \cellcolor{green!33} &  5.8 \cellcolor{red!35} &  2.3 \cellcolor{green!28} &  3.9 \cellcolor{green!0} &  3.1 \cellcolor{green!15} &  5.5 \cellcolor{red!30} &  5.4 \cellcolor{red!27} & \textbf{2.0} \cellcolor{green!35} \\

                \hline
                \end{tabular}%
                }
            
\end{table}

\begin{table}[h!]
  \centering
  \caption{\rabse\ for different regressors for multi-label
  quantifiers.}
  \label{tab:mlqreg:mrae}
  
                \resizebox{\textwidth}{!}{%
            \setlength\tabcolsep{1.5pt}
            
                    \begin{tabular}{|l|c|cccccc|c|cccccc|c|cccccc|} \cline{2-22}
\multicolumn{1}{c|}{} & \multicolumn{7}{c|}{low shift} & \multicolumn{7}{c|}{mid shift} & \multicolumn{7}{c|}{high shift}\\\cline{2-22}

\multicolumn{1}{c|}{} & \side{\bqbc} & \side{Ridge (Ridge-RQ)} & \side{MultitaskLasso} & \side{RandomForest (RF-RQ) \mbox{ }} & \side{StackedLinearSVR} & \side{ChainedLinearSVR\phantom{0}} & \side{LinearSVR (SVR-RQ)\mbox{ }} & \side{\bqbc} & \side{Ridge (Ridge-RQ)} & \side{MultitaskLasso} & \side{RandomForest (RF-RQ) \mbox{ }} & \side{StackedLinearSVR} & \side{ChainedLinearSVR\phantom{0}} & \side{LinearSVR (SVR-RQ)\mbox{ }} & \side{\bqbc} & \side{Ridge (Ridge-RQ)} & \side{MultitaskLasso} & \side{RandomForest (RF-RQ) \mbox{ }} & \side{StackedLinearSVR} & \side{ChainedLinearSVR\phantom{0}} & \side{LinearSVR (SVR-RQ)\mbox{ }}\\\hline
\texttt{Emotions} & \textbf{0.104} \cellcolor{green!35} & 0.120 \cellcolor{green!16} & 0.149 \cellcolor{red!16} & 0.118 \cellcolor{green!19} & 0.165 \cellcolor{red!35} & 0.149$^{\phantom{\ddag}}$ \cellcolor{red!17} & 0.145$^{\phantom{\ddag}}$ \cellcolor{red!12} & 0.366 \cellcolor{green!11} & 0.284$^{\phantom{\ddag}}$ \cellcolor{green!31} & 0.360$^{\phantom{\ddag}}$ \cellcolor{green!13} & \textbf{0.270} \cellcolor{green!35} & 0.454 \cellcolor{red!9} & 0.559 \cellcolor{red!35} & 0.293 \cellcolor{green!29} & 0.543 \cellcolor{green!21} & \textbf{0.428}$^{\phantom{\ddag}}$ \cellcolor{green!35} & 0.543$^{\phantom{\ddag}}$ \cellcolor{green!21} & 0.638$^{\phantom{\ddag}}$ \cellcolor{green!10} & 0.716 \cellcolor{green!1} &  1.026 \cellcolor{red!35} & 0.473$^{\phantom{\ddag}}$ \cellcolor{green!29} \\
\texttt{Scene} & 0.407 \cellcolor{red!35} & 0.143 \cellcolor{green!33} & 0.220 \cellcolor{green!13} & \textbf{0.138} \cellcolor{green!35} & 0.173 \cellcolor{green!26} & 0.170$^{\phantom{\ddag}}$ \cellcolor{green!26} & 0.152$^{\phantom{\ddag}}$ \cellcolor{green!31} & 0.742 \cellcolor{red!35} & 0.234$^{\phantom{\ddag}}$ \cellcolor{green!27} & 0.578$^{\phantom{\ddag}}$ \cellcolor{red!14} & \textbf{0.169} \cellcolor{green!35} & 0.455 \cellcolor{green!0} & 0.443 \cellcolor{green!1} & 0.252 \cellcolor{green!24} &  3.491 \cellcolor{red!22} &  1.192$^{\phantom{\ddag}}$ \cellcolor{green!31} &  4.040$^{\phantom{\ddag}}$ \cellcolor{red!35} &  1.295$^{\dag}$ \cellcolor{green!29} &  2.857 \cellcolor{red!7} &  2.743 \cellcolor{red!4} & \textbf{1.055}$^{\phantom{\ddag}}$ \cellcolor{green!35} \\
\texttt{Yeast} & 0.929 \cellcolor{red!35} & 0.226 \cellcolor{green!33} & 0.217 \cellcolor{green!34} & 0.247 \cellcolor{green!31} & 0.220 \cellcolor{green!33} & 0.228$^{\phantom{\ddag}}$ \cellcolor{green!33} & \textbf{0.208}$^{\phantom{\ddag}}$ \cellcolor{green!35} &  1.038 \cellcolor{red!35} & 0.407$^{\phantom{\ddag}}$ \cellcolor{green!32} & 0.403$^{\ddag}$ \cellcolor{green!32} & \textbf{0.380} \cellcolor{green!35} & 0.438 \cellcolor{green!28} & 0.455 \cellcolor{green!27} & 0.382 \cellcolor{green!34} &  1.607 \cellcolor{red!35} &  1.049$^{\phantom{\ddag}}$ \cellcolor{green!25} &  1.081$^{\dag}$ \cellcolor{green!21} & \textbf{0.956}$^{\phantom{\ddag}}$ \cellcolor{green!35} &  1.199 \cellcolor{green!8} &  1.238 \cellcolor{green!4} &  1.059$^{\phantom{\ddag}}$ \cellcolor{green!23} \\
\texttt{Birds} & 0.550 \cellcolor{red!35} & 0.335 \cellcolor{green!17} & 0.346 \cellcolor{green!14} & \textbf{0.264} \cellcolor{green!35} & 0.353 \cellcolor{green!13} & 0.305$^{\phantom{\ddag}}$ \cellcolor{green!25} & 0.265$^{\dag}$ \cellcolor{green!34} & 0.684 \cellcolor{red!35} & 0.426$^{\phantom{\ddag}}$ \cellcolor{green!16} & 0.435$^{\phantom{\ddag}}$ \cellcolor{green!14} & \textbf{0.335} \cellcolor{green!35} & 0.462 \cellcolor{green!9} & 0.405 \cellcolor{green!20} & 0.337 \cellcolor{green!34} &  1.007 \cellcolor{red!35} & 0.484$^{\phantom{\ddag}}$ \cellcolor{green!22} & 0.510$^{\phantom{\ddag}}$ \cellcolor{green!19} & 0.412$^{\phantom{\ddag}}$ \cellcolor{green!29} & 0.595 \cellcolor{green!9} & 0.609 \cellcolor{green!8} & \textbf{0.364}$^{\phantom{\ddag}}$ \cellcolor{green!35} \\
\texttt{Genbase} & 0.047 \cellcolor{green!29} & \textbf{0.021} \cellcolor{green!35} & 0.266 \cellcolor{red!19} & 0.226 \cellcolor{red!10} & 0.283 \cellcolor{red!23} & 0.336$^{\phantom{\ddag}}$ \cellcolor{red!35} & 0.087$^{\phantom{\ddag}}$ \cellcolor{green!20} & 0.051 \cellcolor{green!30} & \textbf{0.022}$^{\phantom{\ddag}}$ \cellcolor{green!35} & 0.355$^{\phantom{\ddag}}$ \cellcolor{red!20} & 0.307 \cellcolor{red!12} & 0.411 \cellcolor{red!29} & 0.445 \cellcolor{red!35} & 0.095 \cellcolor{green!22} & 0.053 \cellcolor{green!31} & \textbf{0.022}$^{\phantom{\ddag}}$ \cellcolor{green!35} & 0.369$^{\phantom{\ddag}}$ \cellcolor{red!2} & 0.345$^{\phantom{\ddag}}$ \cellcolor{green!0} & 0.652 \cellcolor{red!32} & 0.675 \cellcolor{red!35} & 0.086$^{\phantom{\ddag}}$ \cellcolor{green!28} \\
\texttt{Medical} & \textbf{0.173} \cellcolor{green!35} & 0.336 \cellcolor{red!30} & 0.305 \cellcolor{red!18} & 0.319 \cellcolor{red!23} & 0.297 \cellcolor{red!14} & 0.226$^{\phantom{\ddag}}$ \cellcolor{green!13} & 0.346$^{\phantom{\ddag}}$ \cellcolor{red!35} & \textbf{0.218} \cellcolor{green!35} & 0.471$^{\phantom{\ddag}}$ \cellcolor{red!10} & 0.445$^{\phantom{\ddag}}$ \cellcolor{red!5} & 0.612 \cellcolor{red!35} & 0.449 \cellcolor{red!6} & 0.320 \cellcolor{green!16} & 0.523 \cellcolor{red!19} & \textbf{0.397} \cellcolor{green!35} & 0.795$^{\phantom{\ddag}}$ \cellcolor{green!11} & 0.693$^{\phantom{\ddag}}$ \cellcolor{green!17} &  1.566$^{\phantom{\ddag}}$ \cellcolor{red!35} & 0.919 \cellcolor{green!3} & 0.711 \cellcolor{green!16} & 0.896$^{\phantom{\ddag}}$ \cellcolor{green!5} \\
\texttt{tmc2007\_500} &  1.939 \cellcolor{red!35} & \textbf{0.246} \cellcolor{green!35} & 0.377 \cellcolor{green!29} & 0.291 \cellcolor{green!33} & 0.331 \cellcolor{green!31} & 0.279$^{\phantom{\ddag}}$ \cellcolor{green!33} & 0.263$^{\phantom{\ddag}}$ \cellcolor{green!34} &  2.192 \cellcolor{red!35} & \textbf{0.270}$^{\phantom{\ddag}}$ \cellcolor{green!35} & 0.495$^{\phantom{\ddag}}$ \cellcolor{green!26} & 0.307 \cellcolor{green!33} & 0.526 \cellcolor{green!25} & 0.463 \cellcolor{green!27} & 0.323 \cellcolor{green!33} &  2.402 \cellcolor{red!35} & \textbf{0.336}$^{\phantom{\ddag}}$ \cellcolor{green!35} & 0.900$^{\phantom{\ddag}}$ \cellcolor{green!15} & 0.385$^{\phantom{\ddag}}$ \cellcolor{green!33} &  1.092 \cellcolor{green!9} & 0.975 \cellcolor{green!13} & 0.466$^{\phantom{\ddag}}$ \cellcolor{green!30} \\
\texttt{Ohsumed} &  1.912 \cellcolor{red!35} & \textbf{0.219} \cellcolor{green!35} & 0.332 \cellcolor{green!30} & 0.255 \cellcolor{green!33} & 0.254 \cellcolor{green!33} & 0.250$^{\phantom{\ddag}}$ \cellcolor{green!33} & 0.228$^{\phantom{\ddag}}$ \cellcolor{green!34} &  2.326 \cellcolor{red!35} & \textbf{0.256}$^{\phantom{\ddag}}$ \cellcolor{green!35} & 0.456$^{\phantom{\ddag}}$ \cellcolor{green!28} & 0.286 \cellcolor{green!34} & 0.444 \cellcolor{green!28} & 0.435 \cellcolor{green!28} & 0.274 \cellcolor{green!34} &  2.961 \cellcolor{red!35} & \textbf{0.339}$^{\phantom{\ddag}}$ \cellcolor{green!35} & 0.953$^{\phantom{\ddag}}$ \cellcolor{green!18} & 0.400$^{\phantom{\ddag}}$ \cellcolor{green!33} &  1.186 \cellcolor{green!12} &  1.168 \cellcolor{green!12} & 0.387$^{\phantom{\ddag}}$ \cellcolor{green!33} \\
\texttt{Enron} & 0.621 \cellcolor{red!35} & 0.343 \cellcolor{green!31} & 0.346 \cellcolor{green!30} & 0.339 \cellcolor{green!32} & 0.338 \cellcolor{green!32} & 0.328$^{\ddag}$ \cellcolor{green!34} & \textbf{0.327}$^{\phantom{\ddag}}$ \cellcolor{green!35} & 0.741 \cellcolor{red!35} & 0.420$^{\phantom{\ddag}}$ \cellcolor{green!33} & 0.443$^{\phantom{\ddag}}$ \cellcolor{green!28} & 0.416 \cellcolor{green!33} & 0.455 \cellcolor{green!25} & 0.428 \cellcolor{green!31} & \textbf{0.412} \cellcolor{green!35} &  1.146 \cellcolor{red!35} & 0.628$^{\phantom{\ddag}}$ \cellcolor{green!20} & 0.691$^{\phantom{\ddag}}$ \cellcolor{green!14} & \textbf{0.498}$^{\phantom{\ddag}}$ \cellcolor{green!35} & 0.741 \cellcolor{green!8} & 0.651 \cellcolor{green!18} & 0.569$^{\phantom{\ddag}}$ \cellcolor{green!27} \\
\texttt{Reuters-21578} & 0.910 \cellcolor{red!35} & 0.147 \cellcolor{green!34} & 0.261 \cellcolor{green!23} & 0.254 \cellcolor{green!24} & 0.227 \cellcolor{green!26} & 0.232$^{\phantom{\ddag}}$ \cellcolor{green!26} & \textbf{0.139}$^{\phantom{\ddag}}$ \cellcolor{green!35} &  1.184 \cellcolor{red!35} & 0.175$^{\dag}$ \cellcolor{green!34} & 0.291$^{\phantom{\ddag}}$ \cellcolor{green!26} & 0.273 \cellcolor{green!28} & 0.389 \cellcolor{green!20} & 0.382 \cellcolor{green!20} & \textbf{0.174} \cellcolor{green!35} &  1.040 \cellcolor{red!35} & 0.238$^{\phantom{\ddag}}$ \cellcolor{green!28} & 0.246$^{\phantom{\ddag}}$ \cellcolor{green!28} & 0.246$^{\phantom{\ddag}}$ \cellcolor{green!28} & 0.393 \cellcolor{green!16} & 0.437 \cellcolor{green!13} & \textbf{0.162}$^{\phantom{\ddag}}$ \cellcolor{green!35} \\
\texttt{RCV1-v2} &  2.028 \cellcolor{red!35} & 0.204 \cellcolor{green!34} & 0.329 \cellcolor{green!29} & 0.338 \cellcolor{green!29} & 0.324 \cellcolor{green!29} & 0.324$^{\phantom{\ddag}}$ \cellcolor{green!29} & \textbf{0.190}$^{\phantom{\ddag}}$ \cellcolor{green!35} &  2.525 \cellcolor{red!35} & 0.315$^{\phantom{\ddag}}$ \cellcolor{green!34} & 0.585$^{\phantom{\ddag}}$ \cellcolor{green!25} & 0.521 \cellcolor{green!27} & 0.638 \cellcolor{green!23} & 0.643 \cellcolor{green!23} & \textbf{0.284} \cellcolor{green!35} &  2.970 \cellcolor{red!35} & 0.566$^{\phantom{\ddag}}$ \cellcolor{green!33} & 0.889$^{\phantom{\ddag}}$ \cellcolor{green!24} & 0.911$^{\phantom{\ddag}}$ \cellcolor{green!23} &  1.060 \cellcolor{green!19} &  1.068 \cellcolor{green!19} & \textbf{0.506}$^{\phantom{\ddag}}$ \cellcolor{green!35} \\
\texttt{Mediamill} &  7.677 \cellcolor{red!35} & 0.417 \cellcolor{green!33} & 0.407 \cellcolor{green!33} & 0.377 \cellcolor{green!33} & 0.331 \cellcolor{green!34} & 0.361$^{\phantom{\ddag}}$ \cellcolor{green!34} & \textbf{0.271}$^{\phantom{\ddag}}$ \cellcolor{green!35} &  8.616 \cellcolor{red!35} & 0.539$^{\phantom{\ddag}}$ \cellcolor{green!33} & 0.524$^{\phantom{\ddag}}$ \cellcolor{green!33} & 0.464 \cellcolor{green!34} & 0.523 \cellcolor{green!33} & 0.561 \cellcolor{green!33} & \textbf{0.372} \cellcolor{green!35} &  8.941 \cellcolor{red!35} & 0.675$^{\phantom{\ddag}}$ \cellcolor{green!33} & 0.649$^{\phantom{\ddag}}$ \cellcolor{green!34} & 0.580$^{\phantom{\ddag}}$ \cellcolor{green!34} & 0.810 \cellcolor{green!32} & 0.874 \cellcolor{green!32} & \textbf{0.532}$^{\phantom{\ddag}}$ \cellcolor{green!35} \\
\texttt{Bibtex} &  1.669 \cellcolor{red!35} & 0.396 \cellcolor{green!32} & 0.434 \cellcolor{green!30} & 0.433 \cellcolor{green!30} & 0.425 \cellcolor{green!31} & 0.427$^{\phantom{\ddag}}$ \cellcolor{green!31} & \textbf{0.353}$^{\phantom{\ddag}}$ \cellcolor{green!35} &  1.643 \cellcolor{red!35} & 0.390$^{\phantom{\ddag}}$ \cellcolor{green!31} & 0.490$^{\phantom{\ddag}}$ \cellcolor{green!26} & 0.488 \cellcolor{green!26} & 0.523 \cellcolor{green!24} & 0.477 \cellcolor{green!27} & \textbf{0.330} \cellcolor{green!35} &  1.513 \cellcolor{red!35} & 0.380$^{\phantom{\ddag}}$ \cellcolor{green!30} & 0.567$^{\phantom{\ddag}}$ \cellcolor{green!19} & 0.623$^{\phantom{\ddag}}$ \cellcolor{green!16} & 0.661 \cellcolor{green!14} & 0.544 \cellcolor{green!21} & \textbf{0.305}$^{\phantom{\ddag}}$ \cellcolor{green!35} \\
\texttt{Corel5k} &  2.284 \cellcolor{red!35} & 0.300 \cellcolor{green!32} & 0.296 \cellcolor{green!32} & 0.290 \cellcolor{green!32} & 0.320 \cellcolor{green!31} & 0.423$^{\phantom{\ddag}}$ \cellcolor{green!28} & \textbf{0.230}$^{\phantom{\ddag}}$ \cellcolor{green!35} &  2.511 \cellcolor{red!35} & 0.327$^{\phantom{\ddag}}$ \cellcolor{green!32} & 0.316$^{\phantom{\ddag}}$ \cellcolor{green!32} & 0.309 \cellcolor{green!33} & 0.362 \cellcolor{green!31} & 0.453 \cellcolor{green!28} & \textbf{0.246} \cellcolor{green!35} &  2.973 \cellcolor{red!35} & 0.396$^{\phantom{\ddag}}$ \cellcolor{green!32} & 0.374$^{\phantom{\ddag}}$ \cellcolor{green!32} & 0.369$^{\phantom{\ddag}}$ \cellcolor{green!32} & 0.460 \cellcolor{green!30} & 0.549 \cellcolor{green!28} & \textbf{0.290}$^{\phantom{\ddag}}$ \cellcolor{green!35} \\
\texttt{Delicious} &  6.631 \cellcolor{red!35} & 0.343 \cellcolor{green!34} & 0.369 \cellcolor{green!34} & 0.363 \cellcolor{green!34} & 0.371 \cellcolor{green!34} & 0.415$^{\phantom{\ddag}}$ \cellcolor{green!33} & \textbf{0.301}$^{\phantom{\ddag}}$ \cellcolor{green!35} &  7.474 \cellcolor{red!35} & 0.363$^{\phantom{\ddag}}$ \cellcolor{green!34} & 0.462$^{\phantom{\ddag}}$ \cellcolor{green!33} & 0.364 \cellcolor{green!34} & 0.478 \cellcolor{green!33} & 0.532 \cellcolor{green!33} & \textbf{0.336} \cellcolor{green!35} &  7.327 \cellcolor{red!35} & 0.357$^{\ddag}$ \cellcolor{green!34} & 0.567$^{\phantom{\ddag}}$ \cellcolor{green!32} & \textbf{0.346}$^{\phantom{\ddag}}$ \cellcolor{green!35} & 0.648 \cellcolor{green!31} & 0.720 \cellcolor{green!31} & 0.360$^{\dag}$ \cellcolor{green!34} \\
\hline
\multicolumn{1}{|c|}{Average} &  1.990 \cellcolor{red!35} & 0.260 \cellcolor{green!34} & 0.319 \cellcolor{green!31} & 0.296 \cellcolor{green!32} & 0.302 \cellcolor{green!32} & 0.307 \cellcolor{green!32} & \textbf{0.237}$^{\phantom{\ddag}}$ \cellcolor{green!35} &  1.931 \cellcolor{red!35} & 0.321$^{\phantom{\ddag}}$ \cellcolor{green!34} & 0.458$^{\phantom{\ddag}}$ \cellcolor{green!28} & 0.370 \cellcolor{green!32} & 0.483 \cellcolor{green!27} & 0.478 \cellcolor{green!27} & \textbf{0.304} \cellcolor{green!35} &  2.421 \cellcolor{red!35} & 0.642$^{\ddag}$ \cellcolor{green!34} &  1.358$^{\phantom{\ddag}}$ \cellcolor{green!6} & 0.782$^{\phantom{\ddag}}$ \cellcolor{green!28} &  1.276 \cellcolor{green!9} &  1.260 \cellcolor{green!10} & \textbf{0.627}$^{\phantom{\ddag}}$ \cellcolor{green!35} \\
\multicolumn{1}{|c|}{Rank Average} &  5.9 \cellcolor{red!35} &  3.0 \cellcolor{green!17} &  4.9 \cellcolor{red!17} &  3.8 \cellcolor{green!3} &  4.3 \cellcolor{red!5} &  4.1 \cellcolor{red!1} & \textbf{2.1} \cellcolor{green!35} &  6.1 \cellcolor{red!35} &  2.7 \cellcolor{green!24} &  4.4 \cellcolor{red!5} &  2.6 \cellcolor{green!25} &  5.3 \cellcolor{red!20} &  4.9 \cellcolor{red!13} & \textbf{2.1} \cellcolor{green!35} &  6.0 \cellcolor{red!35} &  2.3 \cellcolor{green!30} &  3.9 \cellcolor{green!1} &  3.0 \cellcolor{green!17} &  5.5 \cellcolor{red!25} &  5.3 \cellcolor{red!23} & \textbf{2.0} \cellcolor{green!35} \\

                \hline
                \end{tabular}%
                }
            
\end{table}

\begin{table}[h!]
  \centering
  \caption{Hyperparameters explored during model selection for
  different regressors. Only the hyperparameters which are specific to
  the regressor are listed. All methods are deployed with an LR
  classifier; for hyperparameters $C$ and \emph{ClassWeight},
  we explore in the ranges $\{10^{-1}, \ldots, 10^{2},10^{3}\}$ and
  \{Balanced, None\}, respectively.}
  \label{tab:modsel:reg}
                  \resizebox{\textwidth}{!}{%
\begin{tabular}{llll}
\toprule
Regressor 				& Hyperparameter & Description & Values \\ \midrule
Ridge    & \emph{Alpha}         & regularization strength               & \{$10^{-3}, \ldots, 10^{2},10^{3}$\}     \\
MultitaskLasso       & \emph{Alpha}         & regularization strength               & \{$10^{-3}, \ldots, 10^{2},10^{3}$\}     \\
RandomForest      & \emph{N\_estimators} & number of trees                             & \{10, 100, 200\} \\
StackedLinearSVR & $C$           & inverse of regularization strength in Linear SVR & \{$10^{-1}, \ldots, 10^{2},10^{3}$\}          \\
ChainedLinearSVR\phantom{0}      & $C$           & inverse of regularization strength in Linear SVR & \{$10^{-1}, \ldots, 10^{2},10^{3}$\}          \\
LinearSVR      & $C$           & inverse of regularization strength in Linear SVR & \{$10^{-1}, \ldots, 10^{2},10^{3}$\}          \\
\bottomrule
\end{tabular}
}
\end{table}


\end{document}